\pdfoutput=1

\documentclass[11pt]{article}
\usepackage[table]{xcolor}
\definecolor{verylightgray}{gray}{0.82}
\usepackage[]{acl}

\usepackage{times}
\usepackage{latexsym}
\usepackage{multirow}
\usepackage{adjustbox}
\usepackage{booktabs} 
\usepackage{subcaption}
\usepackage[T1]{fontenc}
\usepackage[utf8]{inputenc}
\usepackage{microtype}
\usepackage{inconsolata}
\usepackage{graphicx}
\usepackage{ccicons}
\usepackage[most]{tcolorbox}
\usepackage{tabularx}

\usepackage{booktabs}
\usepackage{graphicx}
\usepackage{rotating}

\newcommand{\featuremiss}{\textcolor{black!30}{--}}

\usepackage{xspace}
\newcommand{\noask}{No-ask\xspace}
\newcommand{\frugal}{Frugal prompt\xspace}
\newcommand{\voigate}{VoI gate\xspace}
\newcommand{\budgeted}{Budgeted\xspace}
\newcommand{\budgetedreview}{Budgeted~+ Review\xspace}
\newcommand{\budgetedsummary}{Budgeted~+ Review~+ Summary\xspace}
\newcommand{\budgetedmgr}{Budgeted~+ MGR\xspace}
\newcommand{\mgr}{MGR\xspace}
\newcommand{\mgragent}{MGR agent\xspace}

\usepackage{tikz} 
\DeclareRobustCommand{\stepnum}[1]{\tikz[baseline=(c.base)]\node[circle,fill=black!85,
  text=white,inner sep=0pt,minimum size=1.05em,
  font=\sffamily\bfseries\scriptsize](c){#1};}

\usepackage{tcolorbox}
\tcbuselibrary{skins,breakable}
\tcbset{promptbox/.style={
    enhanced,breakable,
    colback=black!2,colframe=black!45,
    colbacktitle=black!8,coltitle=black,
    boxrule=0.4pt,arc=1mm,
    left=6pt,right=6pt,top=5pt,bottom=5pt,
    before skip=8pt,after skip=10pt,
    fontupper=\small\ttfamily,
    before upper={\raggedright\sloppy\setlength{\parindent}{0pt}\setlength{\parskip}{0pt}}
}}

\usepackage{xcolor}
\definecolor{posgreen}{RGB}{0,128,0}

\tcbset{
    promptbox/.style={
        colback=gray!20, 
        colframe=gray!50, 
        boxrule=0.5pt,
        arc=3pt,
        left=5pt,
        right=5pt,
        top=5pt,
        bottom=5pt,
        sharp corners,
        fontupper=\ttfamily\tiny, 
    }
}

\title{Ask Less, Ask Better: Allocating User Attention in Proactive LLM Agents with Interruption Budgets}

\author{
 \textbf{Xuanming Zhang\textsuperscript{1}\thanks{Work done during an internship at Oracle.}}, 
 \textbf{Amir H. Rezaeian\textsuperscript{2}}, 
 \textbf{Miguel Ballesteros\textsuperscript{2}}, 
 \textbf{Weiyi Sun\textsuperscript{2}}, 
 \textbf{Werner Geyer\textsuperscript{2}} \\
 \textbf{Lydia B. Chilton\textsuperscript{1}}, 
 \textbf{Zhou Yu\textsuperscript{1}}, 
 \textbf{Yassine Benajiba\textsuperscript{2}}, 
 \textbf{Dan Roth\textsuperscript{2}}
\\
\\
 \textsuperscript{1}Columbia University,
 \textsuperscript{2}Oracle AI
\\
\small{
 \texttt{
   xz2995@columbia.edu, \{amir.rezaeian,dan.roth\}@oracle.com
 }}
}

\begin{document}
\maketitle
\begin{abstract}
LLM agents can ask clarifying questions to complete tasks, but answering each question demands users' time and effort. Existing evaluations typically measure this burden by counting questions, treating necessary and avoidable questions equally. We propose an evaluation protocol that distinguishes how often agents ask from how well they ask. An interruption budget limits the number of questions, while a user-attention cost charges extra for questions that are hard to read or not worth asking, as rated by a human-validated LLM judge. Beyond question counts, this cost captures perceived effort: on conversation pairs with the same number of questions, human annotators agreed with the cost on which conversation required more effort 75\% of the time. Across eight LLMs and four customer-service domains, limiting agents to one to three questions reduces user-attention cost by 36--73\% while keeping 86--94\% of the task reward they earn with unlimited questions. At the same budget, reviewing each question against the memory of users' earlier answers cuts attention cost by another 13\% with little reward loss and yields more valuable questions. Models also differ sharply in how well they ask: at the same user-attention cost, the most efficient models gain more than twice as much reward from their questions as the others. These results show that asking less and asking better are distinct goals, and that evaluations of proactive agents should measure both.

\end{abstract}

\section{Introduction}

\begin{figure}[!t]
\centering
\includegraphics[width=0.97\columnwidth]{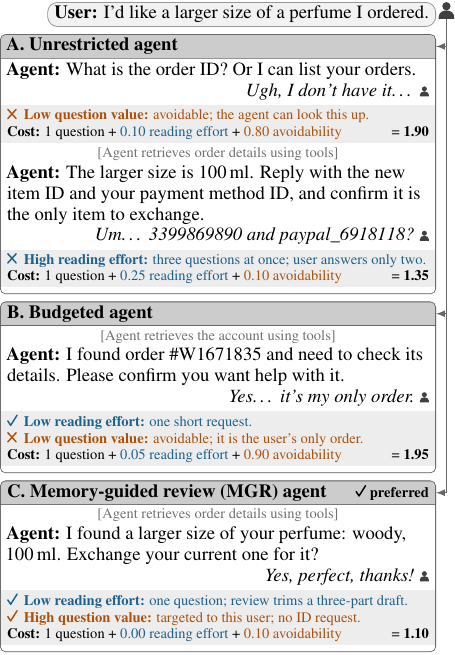}
\caption{\textbf{Question count alone misses what each question costs the user.}
Three agents complete the same request (adapted from real runs); B and C
each ask once under a two-question interruption budget, but C incurs the
lowest user-attention cost and is preferred. We define: Avoidability $=1-{}$(judged
question value)$/10$. Reading effort and question value are scored by a
human-validated LLM judge and defined in
Section~\ref{sec:task-formulation}.}
\label{fig:intro-conversations}
\end{figure}

LLM agents increasingly act as proactive collaborators that use tools
and ask users clarifying questions while completing tasks
\citep{lu2025proactive,wu2025collabllm}. Questions help agents learn
what users want, but each one demands the user's time and effort.
Because attention is scarce \citep{simon1973applying}, the benefit of
asking must be weighed against the cost of interrupting
\citep{horvitz1999principles,mark2008cost}. Evaluations should therefore
measure what questions cost users, not only whether tasks succeed.

Existing evaluations mostly score task success
\citep{yao2024tau,barres2025tau2}, and those that consider the user's
burden typically cap or count questions
\citep{dong2026value,zhao2026and,hamblin2026specbench}. Counts capture
how often an agent asks but not how well, treating necessary and
avoidable questions equally. In Figure~\ref{fig:intro-conversations},
agents B and C each ask once, so a count cannot tell them apart. Yet B
asks the user to confirm their only order, while C proposes a
replacement that only the user can approve.

We propose an evaluation protocol for how agents allocate user
attention, distinguishing how often they ask from how well they ask. An \emph{interruption budget}
caps questions per task. A \emph{user-attention cost} charges
for every question, and more for questions that are hard to read
(\emph{reading effort}) or not worth asking (low \emph{question value}),
both scored by a human-validated LLM judge
(\S\ref{sec:task-formulation}). Agents are compared at equal
question count or user-attention cost. Beyond question counts, this cost reflects
perceived effort: at equal counts, annotators rated the higher-cost
conversation more effortful in 75\% of non-tie judgments
(\S\ref{sec:results-rq1}).

We evaluate eight LLMs on four customer-service domains, where simulated
users reveal preferences only when asked, with interactive research as
an out-of-domain test. 
In customer service, budgets of one to three questions cut user-attention cost by 36--73\% and keep 86--94\% of the task reward earned with unlimited questions. They also raise asking efficiency, the reward gained from asking per unit of cost (\S\ref{sec:results-rq2}).
Models differ sharply: GPT-6-Astra and Grok-4.6 are
the most efficient askers, gaining more than twice as much reward per
unit of cost as any other model, and budgets make both significantly
more efficient. 
Asking more does not mean gaining more: how often a
model asks does not predict how much it gains from asking
(\S\ref{sec:results-rq4}).

Budgets limit how many questions an agent asks, but not which ones.
A question is avoidable when the user has already answered it or the
agent can look it up. Our case study therefore evaluates a
\emph{memory-guided review} (\mgr) agent that checks each proposed
question against persistent memory of the user's answers and task
evidence (\S\ref{sec:methods}). Under the same budget, it cuts attention
cost by a further 13\% with a minimal reward loss, and its questions are
judged more valuable even at matched question counts. Without memory, review
lowers both reward and judged value at equal cost
(\S\ref{sec:results-rq3}). Asking less and asking better are thus distinct objectives, and
evaluations of proactive agents should measure both. 

Overall, our contributions are:
\begin{itemize}\setlength{\itemsep}{0pt}
    \item \textbf{An automatic user-attention cost metric} that tracks human-perceived effort beyond question count.
    \item \textbf{An evaluation protocol} that caps how often agents ask
    and compares how well they ask at equal question count or attention cost.
    \item \textbf{An empirical study across eight models and two task settings}, showing how budgets trade off task reward against user attention and when memory-guided review helps.
\end{itemize}

\section{Related Work}
\label{sec:related-work}

\paragraph{Evaluating user interaction.}
Interruption management treats human attention as scarce, with timing
and context shaping disruption
\citep{simon1973applying,horvitz1999principles,mcfarlane2002scope,horvitz2003learning,mark2008cost};
interactive systems reduce user effort by reusing context or adding
user review \citep{zhang2024jumpstarter,long2025doubleagents}. Agent
benchmarks with simulated users evaluate tool use, hidden intents, and
specification gathering
\citep{yao2024tau,barres2025tau2,zhang2026pi,wang2026turning},
and some bound interaction through turn limits \citep{zhao2026and}
or query budgets \citep{hamblin2026specbench}. $\pi$-Bench distinguishes turn
count from user burden but does not compare burden across transcripts
\citep{zhang2026pi}. Our protocol combines interruption budgets, a
cost beyond count, equal-count and equal-cost comparisons, and human
burden judgments (Table~\ref{tab:related-evaluation}).

\DeclareRobustCommand{\featurecheck}{\tikz[baseline=-.3ex]{\draw[line width=.6pt,line cap=round,line join=round] (0,.10) -- (.065,0) -- (.21,.23);}}

\begin{table}[t]
\centering\footnotesize
\setlength{\tabcolsep}{1.2pt}
\newcommand{\hd}[1]{{\scriptsize #1}}
\begin{tabular}{@{}l ccc cc@{}}
\toprule
& \multicolumn{3}{c}{Protocol} & \multicolumn{2}{c}{Human evidence} \\
\cmidrule(lr){2-4}\cmidrule(l){5-6}
& Limit & Content & Matched & Judge & Burden \\
\midrule
$\tau$-bench \citeyearpar{yao2024tau}   & \featurecheck & \featuremiss & \featuremiss & \featuremiss & \featuremiss \\
SpecBench \citeyearpar{wang2026turning} & \featurecheck & \featuremiss & \featuremiss & \featuremiss & \featuremiss \\
VoI \citeyearpar{dong2026value}         & \featurecheck & \featuremiss & \featuremiss & \featuremiss & \featuremiss \\
AskBench \citeyearpar{zhao2026and}      & \featurecheck & \featurecheck & \featuremiss & \featuremiss & \featuremiss \\
$\pi$-Bench \citeyearpar{zhang2026pi}   & \featuremiss & \featuremiss & \featuremiss & \featurecheck & \featuremiss \\
\midrule
\textbf{This work} & \featurecheck & \featurecheck & \featurecheck & \featurecheck & \featurecheck \\
\bottomrule
\end{tabular}
\caption{\textbf{Evaluation features addressed by related work.}
\emph{Limit}: caps actions, turns, or queries; \emph{Content}: cost
depends on question content; \emph{Matched}: policies compared at equal
cost or question count; \emph{Judge}: evaluator validated against humans;
\emph{Burden}: humans compare conversation burden. Details:
Appendix~\ref{app:related-evaluation}.}
\label{tab:related-evaluation}
\end{table}

\paragraph{Deciding when to ask.}
Prior work studies when and what to ask
\citep{rao2018learning,aliannejadi2019asking,kuhn2022clam,toles2025learning}, using
uncertainty, learned help-seeking, turn-limited planning, information
sufficiency, or training for clarification
\citep{ren2023robots,feng2024large,suri2026structured,hu2024uncertainty,li2024mediq,chen2025learning,berant2025learning,wu2025collabllm}.
Value of information (VoI) weighs a question's expected benefit against
its communication cost \citep{dong2026value}, and anytime
reasoning studies quality under compute budgets
\citep{zhang2026budget}. We instead evaluate how agents allocate
a task-level budget of user attention.
\section{Problem Formulation}
\label{sec:task-formulation}

We use interruption budgets to study whether agents can ask fewer, more valuable questions while preserving task reward. A budget controls how often an agent asks, and user-attention cost measures how demanding and justified each request is. Table~\ref{tab:judge-rubrics} lists what the LLM judge scores.


\paragraph{Setting.}
A user begins a session with a request. The proactive agent uses tools and decides when to ask for further input. Task-relevant information may already be in the conversation, be retrievable through tools, or be held only by the user, such as an unstated preference. At each step $t$, the agent's policy $\pi$ chooses whether to ask the user, call a tool, or respond, given prior messages and tool results. We denote the complete session by $\tau$ and score its outcome with task reward $R(\tau)\in[0,1]$.

\paragraph{Interruption budget.}
An interruption, or question, is a discretionary request for user input after the initial request: a clarification, a confirmation, or a request to respond to a suggestion. Exchanges required by the environment's policy, such as authentication or mandatory confirmations, are exempt; a message that adds any discretionary request is charged. A classifier shared across all methods labels each proposed request (91.0\% agreement with human labels, versus 90.0\% between annotators; Appendix~\ref{app:classifier-validation}). Let $Q(\tau)$ be the discretionary request messages delivered in $\tau$, and each $q\in Q(\tau)$ counts once. A budget $N$ imposes $|Q(\tau)|\le N$. The cap may go unspent. Once it is reached, further discretionary requests are refused and the agent proceeds without them (Appendix~\ref{app:enforcement}).
Appendix~\ref{app:accounting-sensitivity} reports the savings when we also charge for exchanges required by the environment’s policy.

\begin{table}[t]
\centering\footnotesize
\setlength{\tabcolsep}{3pt}
\renewcommand{\arraystretch}{1.05}
\begin{tabularx}{\columnwidth}{@{}>{\raggedright\arraybackslash}p{0.27\columnwidth}>{\raggedright\arraybackslash}X@{}}
\toprule
\multicolumn{2}{@{}l}{\textbf{Reading effort}: five dimensions, each rated 1--5} \\
\midrule
1. Goal clarity & Is it clear what the agent needs and why? \\
2. Single ask & One answerable thing, not several bundled? \\
3. Answer effort & Answerable without decoding jargon? \\
4. Actionability & If it asks for an action, is it concrete? \\
5. Conciseness & As short as possible while still clear? \\
\midrule
\multicolumn{2}{@{}l}{\textbf{Question value}: high only if all four criteria hold} \\
\midrule
1. Missing & The answer is not in the conversation log. \\
2. Unretrievable & The agent cannot find it with its tools. \\
3. Needed & The task requires the answer. \\
4. User-held & The user can supply the answer. \\
\bottomrule
\end{tabularx}
\caption{\textbf{What the LLM judge scores.} Reading effort uses readability \citep{sauberli-etal-2026-controlling}, with an overall score $r(q)\in[0,20]$; question value follows \citet{dong2026value}, with $v(q)\in[0,10]$. See full prompts in Appendix~\ref{app:prompts}, and notation in Table~\ref{tab:notation}.}
\label{tab:judge-rubrics}
\end{table}

\paragraph{User-attention cost beyond count.}
A simple question count treats all requests alike, so we add two surcharges, scored by a human-validated LLM judge using domain-specific rubrics (Appendices~\ref{app:human-validation} and~\ref{app:prompts}).

\emph{Reading effort} is the human's work needed to understand a request and decide how to respond. We use readability as a proxy \citep{sauberli-etal-2026-controlling}. The judge rates the complete message on five dimensions (Table~\ref{tab:judge-rubrics}), then gives an overall score $r(q)\in[0,20]$ (higher is easier). This overall score, not a sum of the dimensions, sets the effort cost $e(q)=1-r(q)/20$.


\emph{Prospective question value} is how useful the answer is expected to be when the agent asks \citep{dong2026value}. Seeing the question and the preceding interaction, but not the reply or the task outcome, the judge assigns $v(q)\in[0,10]$  by four criteria (Table~\ref{tab:judge-rubrics}): the answer is missing from the conversation, cannot be retrieved with tools, is needed for the task, and can be supplied by the user. For example, asking a shopper's unstated size preference can be valuable, whereas asking for a price available on the product page is avoidable. Lower value adds a larger avoidability cost $w(q)=1-v(q)/10$.

Because even a useful question can be hard to read, the cost charges both on top of the question count. With surcharge weights $\kappa_{\mathrm{eff}},\kappa_{\mathrm{val}}\ge 0$, we define the user-attention cost of a session to be
\begin{equation}
A_{\mathrm{full}}(\tau)=\sum_{q\in Q(\tau)}\bigl[1+\kappa_{\mathrm{eff}}\,e(q)+\kappa_{\mathrm{val}}\,w(q)\bigr].
\label{eq:episode-cost}
\end{equation}
We set both weights to 1, so each surcharge counts at most as much as the
question itself and each request costs 1--3 units. Because the weights were
fixed before any human judgments were collected, the human study
(\S\ref{sec:results-rq1}) tests the cost rather than fitting it. The weights
also matter little for our main user-attention cost savings (Appendix~\ref{app:accounting-sensitivity}).

\paragraph{Utility and objective.}
To study the trade-off between reward and user-attention cost, we define utility as $U_\lambda(\tau)=R(\tau)-\lambda A_{\mathrm{full}}(\tau)$, where $\lambda\ge 0$ is the reward a user would give up to avoid one unit of cost.
The objective is to maximize expected utility subject to $|Q(\tau)|\leq N$. Because $\lambda$ differs across users and is unknown in advance, we report reward and cost separately, along with the break-even valuation $\lambda^*$ at which two policies have equal utility, and users who value their attention more than $\lambda^*$ prefer the policy that asks less. The observed tradeoff can thus guide budget choices for different models, tasks, and user preferences.

\section{Memory-Guided Review Agent}
\label{sec:methods}

\begin{figure}[t]
    \centering
    \includegraphics[width=\columnwidth]{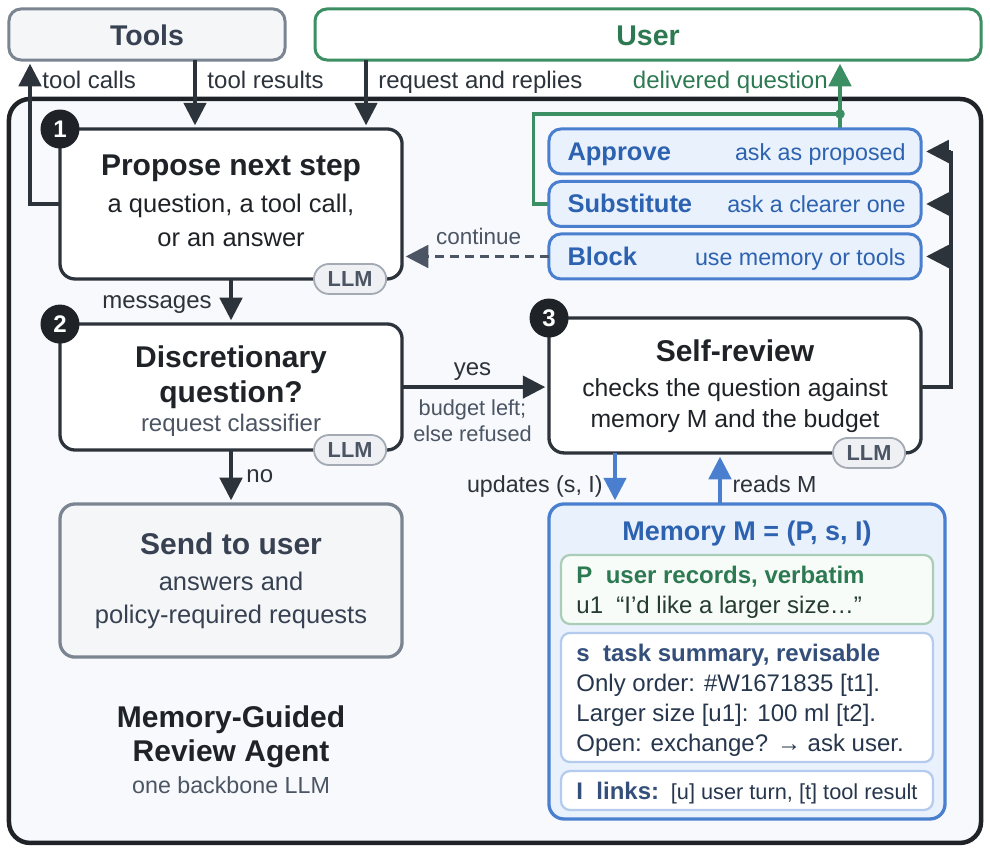}
    \caption{\textbf{The \mgragent.} The backbone LLM proposes a step
    \stepnum{1}. Tool calls go straight to the tools. An LLM classifier
    \stepnum{2} sends answers and policy-required requests to the user
    directly; these are not interruptions and spend no budget. While budget
    remains, a discretionary question goes to self-review \stepnum{3}, a
    second backbone LLM call that checks it against memory $M$: approved or
    substituted questions are delivered and spend one budget; blocked ones
    are not asked.}
    \label{fig:gap-aware-controller}
\end{figure}

An interruption budget limits how often an agent asks, but not which
questions it spends the budget on. We propose the \emph{memory-guided
review} (\mgr) agent (\budgetedmgr in our tables), which uses
\emph{self-review} and persistent memory to choose which questions to
ask. \mgr builds on agent memory
\citep{packer2023memgpt,park2023generative,xu2026mem} and self-feedback
\citep{madaan2023self,shinn2023reflexion}, but applies self-review to
proposed questions rather than task outputs. We expect memory to matter most when an earlier answer or tool finding
is no longer in the recent context, so review without memory cannot see it.
Figure~\ref{fig:gap-aware-controller} shows one step of the agent: the
backbone LLM proposes a step \stepnum{1}, a classifier routes it
\stepnum{2}, and self-review \stepnum{3} decides whether a discretionary
question is worth asking.

\paragraph{Proposal and routing.}
\label{sec:mgr-routing}
At each step $t$, the backbone LLM proposes a question, a tool call, or an
answer \stepnum{1}. A fixed LLM classifier shared by all policies
(Section~\ref{sec:task-formulation}) labels each message to the user
\stepnum{2}. Answers and exchanges required by policy, such as
authentication, are sent to the user directly and spend no budget, and all
other messages are discretionary questions. The prompt states the budget
$N$, and a discretionary question goes to self-review only while budget
remains, $b_t=N-|Q_{<t}|>0$, where $Q_{<t}$ contains the questions
already delivered. Once the budget is spent, further discretionary
questions are refused and the agent continues without them.

\paragraph{Self-review.}
\label{sec:budget-aware-review}
Self-review \stepnum{3} is a separate call to the same backbone LLM, with
the same reasoning settings. It reads persistent memory, selected recent
evidence, and the remaining budget $b_t$, and returns one of three
decisions: \textsc{Approve} a useful question that only the user can
answer; \textsc{Substitute} a clearer, better-targeted question for one
that is needed but poorly targeted; or \textsc{Block} a question whose
answer is already in memory or retrievable with tools, so the agent
continues without asking. Substitutes are classified again before
delivery, and each delivered question spends one unit of budget.
Self-review makes qualitative decisions and does not numerically
optimize the evaluation cost or utility. 
To check that its gains do
not depend on the judge, we also report savings in question count
alone (Section~\ref{sec:results-review}) and compare judges from two
model families (Appendix~\ref{app:second-judge}).


\paragraph{Persistent memory.}
\label{sec:grounded-persistent-memory}
Memory $M=(P,s,I)$ persists across review calls within a session. User records $P$
store the initial request and every user reply verbatim; the model can
append to them but not rewrite them. A revisable summary $s$ records
task constraints, relevant tool findings, and progress, and evidence
links $I$ point to the user turn or tool result that supports each
entry. Each review reads
$M$ and selected recent evidence, then returns its question decision
together with updated $(s,I)$, so review and memory maintenance share
one call. 
Keeping user records separate from tool findings is intended to stop the
agent from treating retrieved facts as user preferences: finding a
larger size in stock does not establish that the user wants to exchange
for it (Figure~\ref{fig:gap-aware-controller}).
The agent can
then avoid repeated questions while still asking about unresolved
preferences, and Section~\ref{sec:results-rq3} measures how often it
obtains them.
Ablations that remove user records (\budgetedsummary) or all persistent
memory (\budgetedreview) test these components (Section~\ref{sec:results-review}).
Self-review and memory maintenance add an average of 1.8 model calls and about 15k tokens per session (Appendix~\ref{app:compute-overhead}).

\section{Experimental Setup}
\label{sec:experimental-setup}



Our experiments address four questions: whether user-attention cost
tracks human burden beyond question count (RQ1, \S\ref{sec:results-rq1}),
how budgets trade task reward for user attention (RQ2, \S\ref{sec:results-rq2}), whether
memory-guided review improves how an agent spends its budget (RQ3, \S\ref{sec:results-rq3}), and
how these effects vary across models and tasks (RQ4, \S\ref{sec:results-rq4}).


\subsection{Datasets}
\label{sec:datasets}
\label{sec:experimental-datasets}

\paragraph{Customer service.}
Our main testbed combines customer support and shopping assistance. 
For $\tau$-bench, we use the \textit{retail} and \textit{airline} domains of
\citet{yao2024tau} and the \textit{telecom} domain of
\citet{barres2025tau2}, with the official $\tau^2$-bench harness,
tools, and reward criteria. Because user requests often leave context
and preferences implicit \citep{malaviya2025contextualized}, we
change the simulated user's instructions: it starts from a high-level
goal and gives further details only when asked. As a robustness check, we also evaluate agents under the
benchmark's original user instructions, where budgets still cut user-attention cost and earn higher utility than unrestricted asking (Appendix~\ref{app:simulator-disclosure}).



For \emph{Ambiguous WebShop}, we adapt synthetic shopping goals
\citep{yao2022webshop} by
withholding preferences \citep{dong2026value}. Each initial request
becomes \textit{``Find me \{query\}.''}, using the product's short query;
attributes, options, and price limits are given privately to the
simulated shopper. For example, \textit{``Find me milks \& creams.''}
omits dietary requirements, package size, and a price cap. The shopper reveals
only requested requirements, potentially several per message. Purchases
are scored against the unchanged complete goal using WebShop's
programmatic reward.

\paragraph{Interactive research.}
\textit{Ambiguous BrowseComp-Plus} provides a separate out-of-domain test.
We adapt BrowseComp-Plus \citep{chen2025browsecomp} by
withholding identifying constraints from the
initial query. The simulated user holds these constraints and supplies
them when asked, but does not know the answer to the research question
itself, which the agent must still find by searching. We report its
five-model results separately from the four-domain main evaluation
(Appendix~\ref{app:budget-across-tasks}).

\paragraph{Data splits.}
We split each task population into about 20\% development tasks for
prompt and self-review design and 80\% test tasks for reported
evaluation. Table~\ref{tab:dataset-splits-preview} lists the size of each dataset and
split, and the test splits total 382 customer-service and 559 research tasks.

\subsection{Baselines and Ablations}
\label{sec:baselines}

\textbf{Unrestricted} allows uncapped asking;
\textbf{\frugal} instructs the model to ask sparingly and concisely.
\textbf{\voigate} adapts \citet{dong2026value}, comparing a question's
estimated information gain about user needs with an asking-cost
threshold. \textbf{\budgeted} enforces an interruption budget $N$ without self-review or added persistent memory. The budget is stated in the prompt for each domain.
\textbf{\budgetedmgr} adds memory-guided review at the same $N$
(Section~\ref{sec:methods}). Two ablations remove its components:
\textbf{\budgetedreview} reviews questions without persistent memory,
and \textbf{\budgetedsummary} keeps the task summary and evidence
links but drops verbatim user records.
\textbf{\noask} sets $N=0$ without separately optimizing the backbone for this setting. See more baseline details in Appendix~\ref{app:baselines}. 
For customer service, where agents ask 2.84 questions per session without
a budget (Appendix~\ref{app:budget-across-tasks}), we test $N=1$--3. For interactive research, where the user holds 2--20 hidden clues per query and
reveals one per question (Appendix~\ref{app:datasets-interactive-research}),
we test $N=1$--8 given the compute limits.

\subsection{Models}
\label{sec:experimental-models}
We evaluate GPT-6-Astra, GPT-5.4, GPT-4.1, Grok-4.6, Grok-4.3, GPT-OSS-120B, Llama-4-Maverick, and Llama-3.3-70B, with three trials per model in the main evaluation on customer service. Interactive research covers GPT-6-Astra, GPT-5.4, GPT-4.1, Llama-4-Maverick, and Grok-4.3, with three trials each. Within comparisons,
methods share the backbone model, provider route, tools, simulated user, and task IDs. Self-review and memory updates use the same model, route, and reasoning effort (default to medium) as action generation.
Additionally, GPT-5.5 simulates the user, GPT-5.6-Terra classifies discretionary requests, and GPT-5.6-Sol judges readability and prospective question value (Appendix~\ref{app:implementation}); because the judge shares a vendor with four backbones, Appendix~\ref{app:second-judge}
compares it with a judge from another family and tests for vendor preference.



\subsection{Metrics and Reporting}
\label{sec:evaluation-metrics}
\label{sec:reporting}
We report four metrics:
(1) \textbf{Task reward} ($R$) is the benchmark's official task score.
(2) \textbf{User-attention cost} ($A_{\mathrm{full}}$) combines
interruption count with reading-effort and avoidability costs, using
human-validated readability and question-value judgments
(Section~\ref{sec:task-formulation}).
(3) \textbf{Utility} ($U_\lambda=R-\lambda A_{\mathrm{full}}$) balances
reward against user-attention cost; we report $\lambda=0.03$, valuing
each cost unit at three reward points on the 0--100 scale, as an illustration, together with the break-even valuation $\lambda^*$, which does not depend on this choice.
(4) \textbf{Question value} ($V$) averages prospective-value scores
over delivered discretionary questions, measuring their justification
given the information available when asking.
Rescaling, alternative valuations, and validation details appear in
Appendix~\ref{app:evaluation-metrics}; judge rubrics are in
Appendix~\ref{app:prompts}.

\section{Results and Analysis}
\label{sec:results}

Table~\ref{tab:main-results} summarizes reward, user-attention cost,
and judged question value for each method.

\begin{table}[t]
\centering\footnotesize
\setlength{\tabcolsep}{3.4pt}
\begin{tabular}{@{}l ccc@{\hspace{8pt}}ccc@{}}
\toprule
& \multicolumn{3}{c}{Customer service}
& \multicolumn{3}{c}{Interactive research} \\
\cmidrule(lr){2-4}\cmidrule(l){5-7}
Method & $R\uparrow$ & $\tilde A\downarrow$ & $\tilde V\uparrow$
       & $R\uparrow$ & $\tilde A\downarrow$ & $\tilde V\uparrow$ \\
\midrule
Unrestricted & \textbf{49.4} & 29.5 & 63.5 & \textbf{32.7} & 19.0 & 64.3 \\
\frugal       & 47.0 & 24.9 & 65.7 & 31.7 & 15.8 & 67.0 \\
\voigate      & 47.4 & 30.7 & 58.6 & 32.3 & 18.4 & 67.3 \\
\budgeted     & 44.7 & 13.7 & 69.4 & 29.6 & 16.0 & 68.0 \\
\budgetedmgr  & 43.9 & \textbf{12.0} & \textbf{71.7} & 28.7 & \textbf{15.1} & \textbf{69.8} \\
\bottomrule
\end{tabular}
\caption{\textbf{Reward, user-attention cost, and judged question value
by method.} Customer service: four domains, eight models; interactive
research: Ambiguous BrowseComp-Plus, five models; three trials each.
Budgeted rows average $N{=}1$--3 and $N{=}1$--8, respectively. $\tilde A$, $\tilde V$:
0--100 scales (Appendix~\ref{app:scoring}). Per-domain and utility results: Tables~\ref{tab:external-domain-preview}
and~\ref{tab:main-domain-summary-full}.}
\label{tab:main-results}
\label{tab:main}
\end{table}

\subsection{RQ1: User-Attention Cost Tracks Burden Beyond Question Count}
\label{sec:results-rq1}
\label{sec:results-burden}

\begin{table}[t]
\centering
\footnotesize
\setlength{\tabcolsep}{2.5pt}
\begin{tabular}{@{}lcc@{}}
\toprule
& \multicolumn{2}{c}{Annotator judgment} \\
\cmidrule(l){2-3}
Pairs (60 each) & Question burden & Overall burden \\
\midrule
Different counts & $90.8\ [84.9,95.8]$ & $62.5\ [52.5,72.5]$ \\
Equal counts & ${71.7}\ [59.1,82.9]$ & ${75.0}\ [64.5,84.9]$ \\
\quad value alone & $71.1\ [58.5,82.8]$ & $73.6\ [62.7,83.7]$ \\
\quad effort alone & $62.7\ [48.1,76.4]$ & $57.3\ [44.0,70.0]$ \\
\quad question words & $49.6\ [35.7,63.4]$ & $56.7\ [44.2,69.6]$ \\
\bottomrule
\end{tabular}
\caption{\textbf{Human burden study: agreement with the user-attention
cost ranking.} Share (\%) of non-tie answers choosing the conversation
ranked higher by the full cost or, in indented rows, by one score alone
(question words: word count of the agent's discretionary questions);
brackets: 95\% pair-bootstrap intervals. Annotators rarely answered
``same'' (at most 7\% per setting). Details: Appendix~\ref{app:burden-study}.}
\label{tab:human-alignment-summary-preview}
\end{table}

User-attention cost captures burden that counts miss. At equal question
counts, conversations whose questions were not worth asking feel more
tiring. Two professional annotators compared 120 transcript pairs, blind
to model, condition, cost, and outcome: 60 differ in question count, and
60 have equal counts, similar length, and a cost gap of at least 0.5
units (Appendix~\ref{app:burden-study}). The judge behind the cost
agrees with annotators on 90\% of held-out comparisons, about as often
as annotators agree with each other
(Appendix~\ref{app:human-validation}).

\paragraph{At equal counts, the cost agrees with human judgments.}
Annotators judged the higher-cost conversation more tiring overall in
75.0\% of non-tie answers, and its questions more burdensome in 71.7\%
(Table~\ref{tab:human-alignment-summary-preview}). When counts differ, count alone already ranks the pair: annotators judge the
conversation with more questions to have more burdensome questions in
90.8\% of answers but to be more tiring overall in only 62.5\%, so
question count alone is a weak guide to overall burden.

\paragraph{Unworthy questions, not long ones, make conversations tiring.}
Scoring equal-count pairs by one component at a time locates the cost's
advantage in judged question value: value alone reaches 73.6\% overall, while the word count of the agent's questions, a
simple length proxy, stays near chance (56.7\%). With both surcharges
weighted equally ($\kappa_{\mathrm{eff}}=\kappa_{\mathrm{val}}=1$ in
Eq.~\ref{eq:episode-cost}), the full cost matches value alone ($+0.6$
and $+1.4$ points on question and overall burden) and beats effort alone
on overall burden by 17.7 points, so adding reading effort does not
dilute alignment. We keep equal weights as a simple default rather than
fitting them to the data, and our savings persist without the surcharges
(Appendix~\ref{app:accounting-sensitivity}).


\subsection{RQ2: Small Budgets Cut Attention Cost and Spend It More Efficiently}
\label{sec:results-rq2}
\label{sec:results-budget}

\begin{figure}[t]
    \centering
    \includegraphics[width=\columnwidth]{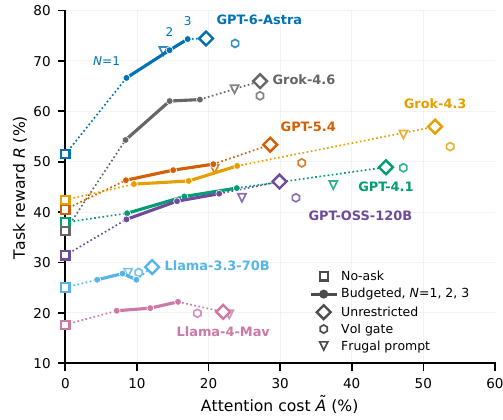}
    \caption{\textbf{Budgets trade reward for user attention, at rates that differ by model.}
Each path is one model, averaged over four domains, from No-ask (square,
zero cost) through Budgeted at $N=1$--3 (dots) to Unrestricted
(diamond); triangles and circles mark the Frugal prompt and VoI gate.
The slope from the square is asking efficiency, the reward gained per
unit of attention cost; upper-left is better. Per-domain panels:
Figure~\ref{fig:rq1-reward-attention-all}.}
    \label{fig:rq1-reward-attention}
\end{figure}

Budgets of one, two, and three questions cut user-attention cost by
73\%, 52\%, and 36\% relative to Unrestricted while lowering reward by
only 7.1, 4.0, and 2.8 points, keeping 86--94\% of total reward
(Figure~\ref{fig:rq1-reward-attention}). However, total reward flatters budgets: \noask already earns much of it, and budgets keep only 49\%,
71\%, and 80\% of the reward that asking adds
(Table~\ref{tab:domain-tradeoff}).

\paragraph{Questions beyond the budget earn little.}
We measure asking efficiency as \emph{yield}, the reward gained over
\noask per unit of user-attention cost, shown by the slopes in
Figure~\ref{fig:rq1-reward-attention}.
The
questions Unrestricted asks beyond a three-question budget earn only 1.8
reward points per unit, about half the average model's yield of 3.5, so
the budget raises yield to 3.9 ($+0.47$ [$+0.14$, $+0.77$]); counting
questions without the judge gives the same result
(Table~\ref{tab:yield-budget}). This marginal rate is also the
break-even valuation, $\lambda^*\approx0.02$
(Table~\ref{tab:paired-effects-full}): users who would give up more than
about two reward points to avoid one unit of cost are better served by
a budget, and others by unrestricted asking. 

\paragraph{Only a hard cap reliably controls frequency.}
Telling agents to ask sparingly (\frugal) saves only 16\% of cost and
leaves yield unchanged; our \voigate does not lower cost and lowers
yield (Tables~\ref{tab:main-results} and~\ref{tab:yield-policies}). A
hard budget is thus what makes frequency a controlled variable. Once
frequency is controlled, differences in quality become visible: in a
follow-up on three GPT models, the VoI gate under the same budget asks as
often as \budgeted but chooses questions judged 8.8 points less valuable
and earns 4.6 fewer reward points at equal cost
(Appendix~\ref{app:voi}).

\subsection{RQ3: Memory Helps Agents Ask Better, Not Only Less}
\label{sec:results-rq3}
\label{sec:results-review}

Under the same budget, the \mgragent asks fewer questions and drops
mostly those not worth asking; without persistent memory, review lowers
both reward and judged value at equal cost.

\paragraph{\mgr asks less without losing reward at equal cost.}
At the same $N$ as \budgeted, \mgr lowers user-attention cost by 12.7\% with minimal reward loss, mostly by asking less often (10.9\% fewer
requests) rather than by bundling questions
(Appendix~\ref{app:accounting-sensitivity}). The loss reflects asking
less, not asking worse: against a \budgeted policy that spends the same
expected attention, reward does not differ detectably ($+0.2$ [$-0.4$,
$+0.8$]).

\paragraph{\mgr drops the questions not worth asking.}
Judged question value rises by 2.3 points at the same budget and stays
1.8 points higher ([1.3, 2.3]) when the two policies ask equally often on
average (Appendix~\ref{app:agent-preview-robustness}), so the gain
reflects which questions are asked, not how many. \mgr removes 20\% of
questions scored 4 or below but only 5\% of those scored 8 or above
(Table~\ref{tab:value-levels}). The gain holds under the human-validated
judge; a judge from another model family, which agrees less with
annotators, finds a smaller gain (Appendix~\ref{app:second-judge}).

\begin{table}[t]
\centering
\small
\setlength{\tabcolsep}{2.5pt}
\renewcommand{\arraystretch}{1.15}
\begin{tabular}{@{}l rr@{\hspace{8pt}}rr@{}}
\toprule
 & \multicolumn{2}{c}{Observed} & \multicolumn{2}{c}{Equal cost} \\
\cmidrule(lr){2-3}\cmidrule(l){4-5}
Policy & $A\downarrow$ & $V\uparrow$ & $\Delta R\uparrow$ & $\Delta\tilde V\uparrow$ \\
\midrule
\budgeted & 13.7 & 69.2 & -- & -- \\
\budgetedreview & 12.8 & 68.6 & $-0.7^{*}$ & $-0.9^{*}$ \\
\budgetedsummary & 12.3 & 70.6 & $+0.1\phantom{^{*}}$ & $+0.9^{*}$ \\
\textbf{\budgetedmgr} & \textbf{12.0} & \textbf{71.5} & $+0.2\phantom{^{*}}$ & $+1.7^{*}$ \\
\bottomrule
\end{tabular}
\caption{\textbf{Memory ablation for self-review.} Means over $N=1$--3,
four domains, and eight models, on the scales of Table~\ref{tab:main-results}.
$V$ averages domains within model, so it differs slightly from that
table. Deltas: differences from \budgeted at equal expected cost;
$^{*}$: unadjusted paired 95\% interval excludes zero. 
Holm-adjusted tests: Table~\ref{tab:headline-family}.}
\label{tab:equal-cost}
\end{table}

\paragraph{Memory is what makes review selective.}
Without memory, review (\budgetedreview) lowers reward ($-0.7$) and
judged value ($-0.9$) relative to \budgeted at equal cost
(Table~\ref{tab:equal-cost}) and removes low- and high-value questions
alike (Table~\ref{tab:value-levels}). With memory, review costs 6.2\%
less and gains 2.9 value points at the same $N$. Keeping the task
summary but dropping the verbatim user records (\budgetedsummary)
recovers only part of this gain. Averaged over the four domains,
self-review needs to know what the user has already said.

\paragraph{Question value must be read alongside recall.}
Value scores only delivered questions, so it cannot see questions an
agent should have asked. On WebShop, where each goal withholds 7.1
requirements on average, we also measure elicitation recall, the share
of withheld requirements the shopper discloses
(Appendix~\ref{app:elicitation-recall}). Even Unrestricted obtains only
31.1\%, so missed requirements, not excess questions, are the larger problem there. 
Recall also exposes a failure that value misses: \mgr's
recall is 2.9 points below \budgeted's at matched count, because its
review rule mistakes options listed on product pages for known
preferences. Requiring a grounded reason for each block restores recall
to within 1.4 points of \budgeted and raises reward by 5.2 points on
three models. It did not transfer to Retail in development, so we apply
it to WebShop only, apart from the main results
(Appendix~\ref{app:review-revision}). Evaluations that report question
value should therefore report recall alongside it.


\subsection{RQ4: Asking Efficiency Varies Across Models, Tasks, and Users}
\label{sec:results-rq4}
\label{sec:results-models}

The protocol separates models by how efficiently they ask, a skill that
neither question frequency nor base task ability reveals, and the value
of budgets and question review depends on the task and the simulated user.

\paragraph{Some models ask far more efficiently than others.}
GPT-6-Astra and Grok-4.6 gain 7.3--7.8 reward points per unit of
attention cost under unrestricted asking, more than twice any other
model (at most 3.3), and a three-question budget raises both
significantly, to 8.9--9.2 (Figure~\ref{fig:rq1-reward-attention};
Table~\ref{tab:yield-budget}). Frequency does not predict efficiency: on
Telecom under Unrestricted, GPT-4.1 asks 5.7 questions for a 5.5-point
gain over \noask, whereas GPT-6-Astra asks 1.6 for a 31.9-point gain
(Figure~\ref{fig:model-domain-asking-preview}). Nor does base ability:
Grok-4.3 earns the second-highest reward without asking but ranks sixth
in efficiency.

\paragraph{The best budget follows asking efficiency.}
At $\lambda=0.03$, the three most efficient askers (GPT-6-Astra,
Grok-4.6, and GPT-OSS-120B) do best with two or three questions, and the
other five with one (Table~\ref{tab:best-budget}), and per-model budgets add
0.9 utility points over the best common budget. The better a model asks,
the more questions it should be allowed.

\paragraph{Review saves attention for most models; reward gains are
model-specific.} \mgr reduces cost for seven of eight models and raises
judged value for six, but its reward gains at equal cost are supported
only for Grok-4.6 and Llama-3.3 (Figure~\ref{fig:rq2-budget-gains}).

\paragraph{Tasks and simulated users change the value of asking.}
On interactive research (five models; $N=1$--8), budgets save only
15.6\% of cost at a 3.1-point reward loss, and \mgr a further 5.4\% at a
further 0.9-point loss. Unlike in customer service, neither improves
utility over Unrestricted at $\lambda=0.03$
(Table~\ref{tab:external-domain-preview}). In a two-model, three-domain
study, \mgr with GPT-6-Astra at $N=1$ raises reward by 11.7 points when
simulated users disclose only what is asked but lowers it by 2.5 points
under the benchmarks' original instructions, while budgets save cost
under both (Appendix~\ref{app:simulator-disclosure}). Because neither
regime is known to be more realistic, evaluations should vary how
forthcoming simulated users are.

\section{Conclusion}
We proposed an evaluation protocol that separates how often proactive
agents ask, capped by an interruption budget, from how well they ask,
measured by user-attention cost. At equal question counts, annotators
judged the higher-cost conversation more tiring in 75\% of non-tie
answers. Budgets of one to three questions cut attention cost by
36--73\% while keeping 86--94\% of reward, though only 49--80\% of the
gain from asking. At the same budget, memory-guided review cuts
attention cost by a further 13\% with little reward loss and asks more
valuable questions; without memory, review hurts. Models also differ
sharply in how well they ask: at equal cost, the most efficient gain
over twice as much reward from their questions as the others. Asking
less and asking better are thus distinct: evaluations should report
reward and attention cost together, compare agents at equal count or
cost, pair question value with recall, and vary how forthcoming
simulated users are.

\section{Limitations}
\label{sec:limitations}

\paragraph{Extending to live collaboration.}
Our controlled customer-service and interactive-research tasks provide a
starting point for studying budgeted interaction with real users,
naturally underspecified requests, and additional languages and modalities.
The current evaluation uses English interactions with simulated users;
disclosure instructions affect the tradeoff
(Appendix~\ref{app:simulator-disclosure}). Future evaluations can test how budgets transfer to users with different knowledge and communication styles.


\paragraph{Measuring experienced burden.}
The conversation study supports evaluation beyond count and motivates
larger, counterbalanced studies in which users interact with different
agents and report effort, fatigue, and willingness to continue. Such
studies would extend the current two-annotator judgments on selected
conversation pairs to experienced burden. Independent participants and
broader pair sampling would strengthen this evidence.

\paragraph{Learning adaptive budgets.}
The observed curves guide choices among tested question limits. A natural extension is to learn budgets from task demands and elicited user preferences. This also invites limits on individual questions within bundled messages and broader comparisons with optimized no-question and VoI policies. 

\section{Ethics Statement}

\paragraph{Human annotation.}
Six professional annotators from an in-house pool took part, two in each
of three tasks, with no annotator in more than one: validating the
judge's readability and question-value scores, validating the request
classifier's labels, and rating pairs of conversations for burden. All
are fluent English speakers with extensive prior NLP annotation
experience; they performed this work as part of their paid employment
and were paid \$30 per hour. Annotators were informed of the purpose of
the study. The material consists of simulated conversations and contains
no personal data.

\paragraph{Data and models.}
We use publicly released benchmarks ($\tau$-bench, $\tau^2$-bench,
WebShop, and BrowseComp-Plus) under their licenses, and commercial and
open-weight language models. Task conversations use simulated users.

\bibliography{custom}

@inproceedings{sauberli-etal-2026-controlling,
    title = "Controlling Reading Ease with Gaze-Guided Text Generation",
    author = {S{\"a}uberli, Andreas  and
      Jepifanova, Darja  and
      Frassinelli, Diego  and
      Plank, Barbara},
    editor = "Demberg, Vera  and
      Inui, Kentaro  and
      Marquez, Llu{\'i}s",
    booktitle = "Proceedings of the 19th Conference of the {E}uropean Chapter of the {A}ssociation for {C}omputational {L}inguistics (Volume 1: Long Papers)",
    month = mar,
    year = "2026",
    address = "Rabat, Morocco",
    publisher = "Association for Computational Linguistics",
    url = "https://aclanthology.org/2026.eacl-long.107/",
    doi = "10.18653/v1/2026.eacl-long.107",
    pages = "2383--2397",
    ISBN = "979-8-89176-380-7"
}

@inproceedings{dong2026value,
  title={Value of Information: A Framework for Human--Agent Communication},
  author={Dong, Yijiang River and Hu, Tiancheng and Hui, Zheng and Zhang, Caiqi and Vuli{\'c}, Ivan and Bobu, Andreea and Collier, Nigel},
  booktitle={Proceedings of the 64th Annual Meeting of the Association for Computational Linguistics (Volume 1: Long Papers)},
  pages={42879--42896},
  year={2026}
}

@article{zheng2023judging,
  title={Judging llm-as-a-judge with mt-bench and chatbot arena},
  author={Zheng, Lianmin and Chiang, Wei-Lin and Sheng, Ying and Zhuang, Siyuan and Wu, Zhanghao and Zhuang, Yonghao and Lin, Zi and Li, Zhuohan and Li, Dacheng and Xing, Eric and others},
  journal={Advances in neural information processing systems},
  volume={36},
  pages={46595--46623},
  year={2023}
}

@inproceedings{liu2023g,
  title={G-eval: NLG evaluation using gpt-4 with better human alignment},
  author={Liu, Yang and Iter, Dan and Xu, Yichong and Wang, Shuohang and Xu, Ruochen and Zhu, Chenguang},
  booktitle={Proceedings of the 2023 conference on empirical methods in natural language processing},
  pages={2511--2522},
  year={2023}
}

@misc{yao2024tau,
      title={$\tau$-bench: A Benchmark for Tool-Agent-User Interaction in Real-World Domains}, 
      author={Shunyu Yao and Noah Shinn and Pedram Razavi and Karthik Narasimhan},
      year={2024},
      eprint={2406.12045},
      archivePrefix={arXiv},
      primaryClass={cs.AI},
      url={https://arxiv.org/abs/2406.12045}, 
}

@misc{barres2025tau2,
      title={$\tau^2$-Bench: Evaluating Conversational Agents in a Dual-Control Environment}, 
      author={Victor Barres and Honghua Dong and Soham Ray and Xujie Si and Karthik Narasimhan},
      year={2025},
      eprint={2506.07982},
      archivePrefix={arXiv},
      primaryClass={cs.AI},
      url={https://arxiv.org/abs/2506.07982}, 
}

@article{malaviya2025contextualized,
  title={Contextualized evaluations: Judging language model responses to underspecified queries},
  author={Malaviya, Chaitanya and Chang, Joseph Chee and Roth, Dan and Iyyer, Mohit and Yatskar, Mark and Lo, Kyle},
  journal={Transactions of the Association for Computational Linguistics},
  volume={13},
  pages={878--900},
  year={2025}
}

@article{yao2022webshop,
  title={Webshop: Towards scalable real-world web interaction with grounded language agents},
  author={Yao, Shunyu and Chen, Howard and Yang, John and Narasimhan, Karthik},
  journal={Advances in Neural Information Processing Systems},
  volume={35},
  pages={20744--20757},
  year={2022}
}

@article{chen2025browsecomp,
  title={Browsecomp-plus: A more fair and transparent evaluation benchmark of deep-research agent},
  author={Chen, Zijian and Ma, Xueguang and Zhuang, Shengyao and Nie, Ping and Zou, Kai and Liu, Andrew and Green, Joshua and Patel, Kshama and Meng, Ruoxi and Su, Mingyi and others},
  journal={arXiv preprint arXiv:2508.06600},
  year={2025}
}

@article{ouedraogo2025human,
  title={Human-Aligned Code Readability Assessment with Large Language Models},
  author={Ou{\'e}draogo, Wendk{\^u}uni C and Li, Yinghua and Dang, Xueqi and Borsukiewicz, Pawel and Zhou, Xin and Koyuncu, Anil and Klein, Jacques and Lo, David and Bissyand{\'e}, Tegawend{\'e} F},
  journal={arXiv preprint arXiv:2510.16579},
  year={2025}
}

@inproceedings{horvitz1999principles,
  title={Principles of mixed-initiative user interfaces},
  author={Horvitz, Eric},
  booktitle={Proceedings of the SIGCHI conference on Human Factors in Computing Systems},
  pages={159--166},
  year={1999}
}

@inproceedings{lu2025proactive,
  title={Proactive agent: Shifting llm agents from reactive responses to active assistance},
  author={Lu, Yaxi and Yang, Shenzhi and Qian, Cheng and Chen, Guirong and Luo, Qinyu and Wu, Yesai and Wang, Huadong and Cong, Xin and Zhang, Zhong and Lin, Yankai and others},
  booktitle={International Conference on Learning Representations},
  volume={2025},
  pages={47431--47457},
  year={2025}
}

@article{wu2025collabllm,
  title={Collabllm: From passive responders to active collaborators},
  author={Wu, Shirley and Galley, Michel and Peng, Baolin and Cheng, Hao and Li, Gavin and Dou, Yao and Cai, Weixin and Zou, James and Leskovec, Jure and Gao, Jianfeng},
  journal={arXiv preprint arXiv:2502.00640},
  year={2025}
}

@article{simon1973applying,
  title={Applying information technology to organization design},
  author={Simon, Herbert A},
  journal={Public administration review},
  volume={33},
  number={3},
  pages={268--278},
  year={1973},
  publisher={JSTOR}
}

@article{mcfarlane2002scope,
  title={The scope and importance of human interruption in human-computer interaction design},
  author={McFarlane, Daniel C and Latorella, Kara A},
  journal={Human-Computer Interaction},
  volume={17},
  number={1},
  pages={1--61},
  year={2002},
  publisher={Taylor \& Francis}
}

@inproceedings{horvitz2003learning,
  title={Learning and reasoning about interruption},
  author={Horvitz, Eric and Apacible, Johnson},
  booktitle={Proceedings of the 5th international conference on Multimodal interfaces},
  pages={20--27},
  year={2003}
}

@article{zhang2026pi,
  title={pi-Bench: Evaluating Proactive Personal Assistant Agents in Long-Horizon Workflows},
  author={Zhang, Haoran and Xu, Luxin and Wang, Zhilin and Gui, Runquan and Zhang, Shunkai and Lei, Haodi and He, Zihao and He, Bingsu and Qin, Chicheng and Zhu, Tong and others},
  journal={arXiv preprint arXiv:2605.14678},
  year={2026}
}

@article{wang2026turning,
  title={Turning Intent into Specifications: A Benchmark and an Interactive User-Assistant Agent},
  author={Wang, Hao and Han, Ligong and Xu, Kai and Srivastava, Akash},
  journal={arXiv preprint arXiv:2606.20585},
  year={2026}
}

@inproceedings{rao2018learning,
  title={Learning to ask good questions: Ranking clarification questions using neural expected value of perfect information},
  author={Rao, Sudha and Daum{\'e} III, Hal},
  booktitle={Proceedings of the 56th Annual Meeting of the Association for Computational Linguistics (Volume 1: Long Papers)},
  pages={2737--2746},
  year={2018}
}

@inproceedings{aliannejadi2019asking,
  title={Asking clarifying questions in open-domain information-seeking conversations},
  author={Aliannejadi, Mohammad and Zamani, Hamed and Crestani, Fabio and Croft, W Bruce},
  booktitle={Proceedings of the 42nd international acm sigir conference on research and development in information retrieval},
  pages={475--484},
  year={2019}
}

@article{kuhn2022clam,
  title={Clam: Selective clarification for ambiguous questions with generative language models},
  author={Kuhn, Lorenz and Gal, Yarin and Farquhar, Sebastian},
  journal={arXiv preprint arXiv:2212.07769},
  year={2022}
}

@inproceedings{zhao2026and,
  title={When and what to ask: Askbench and rubric-guided rlvr for llm clarification},
  author={Zhao, Jiale and Fang, Ke and Cheng, Lu},
  booktitle={Findings of the Association for Computational Linguistics: ACL 2026},
  pages={17120--17140},
  year={2026}
}

@inproceedings{suri2026structured,
  title={Structured uncertainty guided clarification for LLM agents},
  author={Suri, Manan and Mathur, Puneet and Lipka, Nedim and Dernoncourt, Franck and Rossi, Ryan A and Manocha, Dinesh},
  booktitle={Findings of the Association for Computational Linguistics: ACL 2026},
  pages={40811--40838},
  year={2026}
}

@article{packer2023memgpt,
  title={Memgpt: Towards llms as operating systems},
  author={Packer, Charles and Wooders, Sarah and Lin, Kevin and Fang, Vivian and Patil, Shishir G and Stoica, Ion and Gonzalez, Joseph E},
  journal={arXiv preprint arXiv:2310.08560},
  year={2023}
}

@inproceedings{chen2025learning,
  title={Learning to clarify: Multi-turn conversations with action-based contrastive self-training},
  author={Chen, Maximillian and Sun, Ruoxi and Pfister, Tomas and Arik, Sercan},
  booktitle={International Conference on Learning Representations},
  volume={2025},
  pages={32244--32279},
  year={2025}
}

@article{berant2025learning,
  title={Learning steerable clarification policies with collaborative self-play},
  author={Berant, Jonathan and Chen, Maximillian and Fisch, Adam and Aghajani, Reza and Huot, Fantine and Lapata, Mirella and Eisenstein, Jacob},
  journal={arXiv preprint arXiv:2512.04068},
  year={2025}
}

@inproceedings{mark2008cost,
  title={The cost of interrupted work: more speed and stress},
  author={Mark, Gloria and Gudith, Daniela and Klocke, Ulrich},
  booktitle={Proceedings of the SIGCHI conference on Human Factors in Computing Systems},
  pages={107--110},
  year={2008}
}

@article{zhang2024jumpstarter,
  title={Jumpstarter: Human-ai planning with task-structured context curation},
  author={Zhang, Xuanming and Wang, Sitong and Ma, Jenny and Hwang, Alyssa and Yu, Zhou and Chilton, Lydia B},
  journal={arXiv preprint arXiv:2410.03882},
  year={2024}
}

@article{long2025doubleagents,
  title={DoubleAgents: Human-Agent Alignment in a Socially Embedded Workflow},
  author={Long, Tao and Zhang, Xuanming and Wang, Sitong and Yu, Zhou and Chilton, Lydia B},
  journal={arXiv preprint arXiv:2509.12626},
  year={2025}
}

@article{ren2023robots,
  title={Robots that ask for help: Uncertainty alignment for large language model planners},
  author={Ren, Allen Z and Dixit, Anushri and Bodrova, Alexandra and Singh, Sumeet and Tu, Stephen and Brown, Noah and Xu, Peng and Takayama, Leila and Xia, Fei and Varley, Jake and others},
  journal={arXiv preprint arXiv:2307.01928},
  year={2023}
}

@inproceedings{feng2024large,
  title={Large language model-based human-agent collaboration for complex task solving},
  author={Feng, Xueyang and Chen, Zhi-Yuan and Qin, Yujia and Lin, Yankai and Chen, Xu and Liu, Zhiyuan and Wen, Ji-Rong},
  booktitle={Findings of the Association for Computational Linguistics: EMNLP 2024},
  pages={1336--1357},
  year={2024}
}

@article{hu2024uncertainty,
  title={Uncertainty of thoughts: Uncertainty-aware planning enhances information seeking in llms},
  author={Hu, Zhiyuan and Liu, Chumin and Feng, Xidong and Zhao, Yilun and Ng, See-Kiong and Luu, Anh T and He, Junxian and Koh, Pang W and Hooi, Bryan},
  journal={Advances in Neural Information Processing Systems},
  volume={37},
  pages={24181--24215},
  year={2024}
}

@article{li2024mediq,
  title={Mediq: Question-asking llms and a benchmark for reliable interactive clinical reasoning},
  author={Li, Shuyue S and Balachandran, Vidhisha and Feng, Shangbin and Ilgen, Jonathan S and Pierson, Emma and Koh, Pang W and Tsvetkov, Yulia},
  journal={Advances in Neural Information Processing Systems},
  volume={37},
  pages={28858--28888},
  year={2024}
}

@inproceedings{zhang2026budget,
  title={Budget-Aware Anytime Reasoning with LLM-Synthesized Preference Data},
  author={Zhang, Xuanming and Ashrafi, Shwan and Mirsaidova, Aziza and Rezaeian, Amir H and Ballesteros, Miguel and Chilton, Lydia and Yu, Zhou and Roth, Dan},
  booktitle={Findings of the Association for Computational Linguistics: ACL 2026},
  pages={8587--8599},
  year={2026}
}

@inproceedings{park2023generative,
  title={Generative agents: Interactive simulacra of human behavior},
  author={Park, Joon Sung and O'Brien, Joseph and Cai, Carrie Jun and Morris, Meredith Ringel and Liang, Percy and Bernstein, Michael S},
  booktitle={Proceedings of the 36th annual acm symposium on user interface software and technology},
  pages={1--22},
  year={2023}
}

@article{xu2026mem,
  title={A-mem: Agentic memory for llm agents},
  author={Xu, Wujiang and Liang, Zujie and Mei, Kai and Gao, Hang and Tan, Juntao and Zhang, Yongfeng},
  journal={Advances in Neural Information Processing Systems},
  volume={38},
  pages={17577--17604},
  year={2025}
}

@article{madaan2023self,
  title={Self-refine: Iterative refinement with self-feedback},
  author={Madaan, Aman and Tandon, Niket and Gupta, Prakhar and Hallinan, Skyler and Gao, Luyu and Wiegreffe, Sarah and Alon, Uri and Dziri, Nouha and Prabhumoye, Shrimai and Yang, Yiming and others},
  journal={Advances in neural information processing systems},
  volume={36},
  pages={46534--46594},
  year={2023}
}

@article{shinn2023reflexion,
  title={Reflexion: Language agents with verbal reinforcement learning},
  author={Shinn, Noah and Cassano, Federico and Gopinath, Ashwin and Narasimhan, Karthik and Yao, Shunyu},
  journal={Advances in neural information processing systems},
  volume={36},
  pages={8634--8652},
  year={2023}
}

@article{fieller1954some,
  title={Some problems in interval estimation},
  author={Fieller, Edgar C},
  journal={Journal of the Royal Statistical Society Series B: Statistical Methodology},
  volume={16},
  number={2},
  pages={175--185},
  year={1954},
  publisher={Oxford University Press}
}

@article{hamblin2026specbench,
  title={SpecBench: Evaluating Specification-Level Reasoning for Software Engineering LLM Agents},
  author={Hamblin, Grant and Song, Kevin and Zhu, Zhanda and Jayarajan, Anand and Liu, Sihang and Vijaykumar, Nandita and Pekhimenko, Gennady},
  journal={arXiv preprint arXiv:2605.30314},
  year={2026}
}

@inproceedings{toles2025learning,
  title={Learning and Evaluating Factual Clarification Question Generation Without Examples},
  author={Toles, Matthew and Huang, Yukun and Yu, Zhou},
  booktitle={Proceedings of the Fourth Workshop on Generation, Evaluation and Metrics (GEM$^2$)},
  pages={200--211},
  year={2025}
}

\appendix

\clearpage

\section{Use of AI Assistants}
\label{app:ai-assistance}
We used an AI assistant for writing support. It gave feedback on drafts, suggested clearer and shorter wording for sections and captions, helped condense the paper to the page limit, and proposed the layout of the method figure
(Figure~\ref{fig:gap-aware-controller}).
The authors designed the study, ran all experiments and analyses,
checked every suggested edit against the results, and take full
responsibility for the content. LLMs that act as agents, simulated
users, request classifiers, or judges are research objects, described
in Section~\ref{sec:experimental-setup} and
Appendix~\ref{app:implementation}.

\section{Dataset Construction and Environment Details}
\label{app:datasets}
\subsection{Customer Service}
\label{app:datasets-customer-service}
\paragraph{Retail, Airline, and Telecom.}
We execute Retail and Airline from $\tau$-bench \citep{yao2024tau}
and base Telecom from $\tau^2$-bench \citep{barres2025tau2} in the
$\tau^2$ harness, retaining policies, tools, initial states, task specifications,
and success checks. Telecom also supports simulated-user tool actions.
The main \emph{Guarded} simulator begins with a high-level request and
supplies scenario details when asked. Required-only exchanges are exempt;
mixed messages are charged (Appendix~\ref{app:request-control}).
The upstream \emph{Original} instructions also prescribe progressive
disclosure. A separate prompt-regime comparison, including respective
Telecom user-tool guidelines, appears in Appendix~\ref{app:simulator-disclosure}.

\paragraph{Ambiguous WebShop.}
Following withheld-preference clarification \citep{dong2026value},
we deterministically adapt WebShop synthetic goals \citep{yao2022webshop}.
The opening becomes \textit{``Find me \{query\}.''}; desired attributes,
options, and any finite price bound are private to the shopper. The complete
goal and programmatic purchase reward remain unchanged. Product loading,
goal generation, and shuffling use seed 42; we retain the first 200 eligible
goals. Checks require a nonempty query, withheld options or a finite price
bound, and exact reconstruction of the original instruction. Openings
containing private attribute or option values are rejected (19 candidates);
there are no reconstruction failures. Evaluation uses identical seeded prices.

GPT-5.5 is instructed to answer only requested requirements, disclose
several facts when asked together, and state no preference when a requested
requirement is absent. These instructions do not guarantee compliance.
Agents search, inspect, choose options, and purchase; reward evaluates
attributes, options, price, and product type independently of our question
judge. A compound request consumes one interruption.

\subsection{Interactive Research}
\label{app:datasets-interactive-research}
\paragraph{Ambiguous BrowseComp-Plus.}
GPT-5.6-Sol decomposes 830 BrowseComp-Plus queries \citep{chen2025browsecomp}
into targets and identifying clues, ranks clues weakest to strongest, and
writes openings with two weak clues plus \textit{``I remember a few more
details if you need them.''} Remaining clues are private; precise queries
and reference answers remain unchanged for evaluation. The same model audits
consistency, searchability, naturalness, hidden-clue omission, and answer-name
leakage. It must retrieve another entity compatible with every opening clue,
excluding reference-answer documents, then recheck up to 700 words of its
documents. At least one alternative must survive; string checks also screen
answer names. No actor-success results enter construction. Of 830 items,
699 pass, 130 fail, and one has a construction error; accepted items have
2--20 hidden clues. Same-model auditing and treating unmentioned facts as
compatible limit this check: it does not prove retrieval cannot identify
the answer.

Agents search a fixed corpus, open documents, use native ASK, and submit
answers. Broad requests elicit one unseen clue, strongest first; targeted
requests elicit matching clues. The simulator additionally sees up to three
reference-document excerpts of 50 words each. It does not receive the answer
field directly and is instructed not to solve the question, but excerpts
may contain answer-related evidence; the user channel exceeds intent clues
alone. Every ASK is charged, without an exemption classifier. The environment
permits 100 action turns and ten answered requests; Unrestricted means no
experimental cap within these limits. Five models have three trials,
$N=1$--8, and binary correctness under the upstream answer rubric. This
custom interactive variant is not an official leaderboard evaluation.

\subsection{Splits and Coverage}
\label{app:datasets-splits}

\begin{table}[t]
\centering
\small
\setlength{\tabcolsep}{4pt}
\begin{tabular}{@{}lrrr@{}}
\toprule
Dataset & Total & Dev & Test \\
\midrule
\multicolumn{4}{@{}l}{\textit{Customer service}} \\
Retail & 114 & 23 & 91 \\
Airline & 50 & 10 & 40 \\
Telecom & 114 & 23 & 91 \\
Ambiguous WebShop & 200 & 40 & 160 \\
\midrule
\multicolumn{4}{@{}l}{\textit{Interactive research}} \\
Ambiguous BrowseComp-Plus & 699 & 140 & 559 \\
\bottomrule
\end{tabular}
\caption{Dataset sizes and development/test splits. WebShop and BrowseComp-Plus use the ambiguous variants.}
\label{tab:dataset-splits-preview}
\end{table}

We assign tasks once with seed 42 to approximately 20\% development and
80\% test partitions (Table~\ref{tab:dataset-splits-preview}).
Development supports design and debugging; comparisons share test IDs,
initial specifications, and trial coverage within a model. Variants and
repetitions inherit their parent task's partition. Historical exposure
means these are not newly untouched test sets. 
We will release the construction code and experiment settings for each dataset we run.

\section{Baseline Policies and Agent Configuration}
\label{app:baselines}
\paragraph{Reference policies.}
Methods share tools, users, and request accounting. \textbf{Unrestricted}
has no experimental cap; \textbf{\frugal} prompts sparing, concise asking;
\textbf{\budgeted} caps delivered discretionary messages at $N$;
\textbf{\noask} sets $N=0$ without separate optimization. Following
Section~\ref{sec:baselines}, budgeted methods state $N$ in the agent's
prompt in every domain, and the harness enforces the cap. The remaining
budget is also supplied to the \mgragent's self-review.
Customer service uses $N=1$--3; research uses $N=1$--8. Prompts appear
in Appendix~\ref{app:prompts}.

\subsection{Value-of-Information Comparisons}
\label{app:voi}
Following \citet{dong2026value}, the backbone enumerates up to five
interpretations of user needs, assigns priors, proposes up to four questions,
predicts replies, and updates probabilities. It asks the best question if
its expected increase in the maximum interpretation probability exceeds $c$.
This measures uncertainty, not predicted reward or our judge's value score.

The original adaptation uses $c=0.2686$, the mean valuation-grid value,
without announcing it in the action prompt. This historical numerical
choice does not equate the probability-gain threshold $c$ with the
reward-per-cost-unit valuation $\lambda$. Clarifications are gated;
confirmations and suggestions bypass review. Limits are two drafts per
$\tau$ turn and six per WebShop session. A $\tau$ draft still rejected
after two revisions is delivered; parsing failures approve. BrowseComp
uses a prompted approximation. In 3,445 first-trial reviews from three
GPT models, decisions are 45\% approve, 37\% substitute, and 18\% reject.
Best-question estimates have median 0.45 (10th--90th percentiles 0.19--0.66);
82\% exceed $c$. Thresholds 0.40/0.50/0.60/0.75 pass 61\%/41\%/21\%/1.5\%.
Estimated and judged value have pooled rank correlation 0.00 among approved
questions. These diagnostics concern our implementation, not the original
method in its original setting.

\paragraph{Enforced-gate follow-up.}
We score every discretionary request, review every draft, and refuse drafts
still rejected after revisions. Thresholds $c=0.50,0.65$ were chosen from
the diagnostic before running GPT-4.1, GPT-5.4, and GPT-6-Astra on four
domains, one trial, 1,146 sessions per threshold. Counts fall from 3.40
to 1.73/0.44. Because the gate implementation also changes, these
points and the original $c=0.2686$ result are not a threshold-only
sweep. \budgeted earns 4.2/3.0 more reward points at equal
expected cost (Table~\ref{tab:strict-voi}; unadjusted paired intervals
exclude zero). At $c=0.50$, value is 65.3 versus 73.1 for \budgeted at
$N=2$. Reviews average 6.4/8.1 per session, vetoing 71\%/94\%.
\begin{table*}[t]
\centering
\footnotesize
\setlength{\tabcolsep}{4pt}
\begin{tabular}{@{}l rrr r rr r@{}}
\toprule
 & Requests & User-attention & & $\Delta R$ at & Reviews & Vetoed & Fixed message \\
Policy & per session & cost & Reward & equal cost & per session & (\%) & (\%, $\tau$-bench) \\
\midrule
Unrestricted & 3.12 & 31.0 & 58.9 & -- & -- & -- & 0.0 \\
\voigate as run ($c=0.27$) & 3.40 & 34.6 & 57.3 & -- & -- & -- & 0.0 \\
\midrule
\voigate, enforced veto ($c=0.50$) & 1.73 & 17.2 & 51.0 & $-4.2$$^{*}$ & 6.4 & 71 & 6.9 \\
\voigate, enforced veto ($c=0.65$) & 0.44 & 4.1 & 44.8 & $-3.0$$^{*}$ & 8.1 & 94 & 12.0 \\
\midrule
\budgeted ($N=1$) & 0.94 & 8.5 & 50.9 & 0 & -- & -- & 6.2 \\
\budgeted ($N=2$) & 1.63 & 15.4 & 54.5 & 0 & -- & -- & 3.8 \\
\budgeted ($N=3$) & 2.14 & 20.6 & 56.2 & 0 & -- & -- & 2.6 \\
\bottomrule
\end{tabular}
\caption{\textbf{Enforced VoI gate.} Three GPT models, four domains, one trial, 1,146 sessions per threshold. Every discretionary draft is reviewed and vetoes enforced. Unrestricted, \voigate as run, and \budgeted rows: main evaluation, three trials, same tasks. The original row uses a different gate; it is not another point in the same threshold sweep. $R,A$: main display scales; $\Delta R$: \voigate minus cost-matched \budgeted. Fixed message: $\tau$ fallback rate. $^{*}$: unadjusted paired 95\% interval excludes zero. Maximum reviews: 52/86 at $c=0.50/0.65$.}
\label{tab:strict-voi}
\end{table*}

\paragraph{VoI gate under the same budget.}
\label{app:voi-cap}
Neither run above gives the \voigate the hard budget of the budgeted
agents. We therefore add the enforced gate to \budgeted: while allowance
remains, every discretionary request is scored at the original
$c=0.2686$; a preferred question replaces the draft and is charged; a
rejected request is refused without charge. The cap, charging, refusal
path, and action prompt are those of \budgeted, so this arm
(\budgeted~+ \voigate) differs from \budgeted by the gate alone, as
\budgetedmgr differs from it by memory-guided review alone.
GPT-4.1, GPT-5.4, and GPT-6-Astra run four domains at $N=1$--3 with
three trials, 3,438 sessions per budget. The follow-up was added after
the results above were known, with interpretation rules recorded before
its first trial; the second and third trials were added after the first
was read. The gate reviews 1.2/2.1/2.8 drafts per session, approving
43\%, substituting 36\%, and rejecting 21\%. Request counts match
\budgeted (0.91/1.61/2.14 versus 0.94/1.63/2.14), but pooled over
budgets reward is 4.5 points lower, judged value 8.8 points lower, and
reward at equal expected cost 4.6 points lower
(Table~\ref{tab:voi-cap}). Against \budgetedmgr, the arm incurs 18.6\%
more user-attention cost, with value 11.2 points and equal-cost reward
3.8 points lower. These unadjusted paired intervals exclude zero. The
first trial alone gave 5.3, 8.8, and 5.5 points against \budgeted and
18.1\%, 11.2, and 4.7 points against \budgetedmgr. The equal-cost
deficit against \budgetedmgr is 2.2/3.2/6.1 points for
GPT-4.1/GPT-5.4/GPT-6-Astra and 5.6/2.0/5.6/2.1 points on
Retail/Airline/Telecom/WebShop; the Airline interval includes zero, and
on WebShop the deficit is confined to $N=1$. Under the same budget, our
VoI adaptation thus selects lower-valued questions than either budgeted
agent. This covers three GPT models and one threshold; it does not
evaluate VoI in its original setting. Our scoring checks originally
rejected any WebShop session that contained a VoI veto under the
\budgeted configuration, as a guard against mixing arms; we changed this
check once to admit the new arm and recorded the change. Every
first-trial session was kept and none was rerun.
\begin{table*}[t]
\centering
\footnotesize
\setlength{\tabcolsep}{3.5pt}
\begin{tabular}{@{}ll rrrr r rr r@{}}
\toprule
 & & Requests & User-attention & & Question & $\Delta R$ at & Reviews & Vetoed & Fixed message \\
$N$ & Policy & per session & cost & Reward & value & equal cost & per session & (\%) & (\%, $\tau$-bench) \\
\midrule
1 & \budgeted & 0.94 & 8.5 & 50.9 & 78.1 & 0 & -- & -- & 6.2 \\
 & \budgeted~+ \voigate & 0.91 & 8.8 & 46.5 & 67.8 & $-4.6$$^{*}$ & 1.2 & 20 & 5.4 \\
 & \budgetedmgr & 0.83 & 7.4 & 49.3 & 79.3 & $-0.3$\phantom{$^{*}$} & -- & -- & 6.9 \\
\midrule
2 & \budgeted & 1.63 & 15.4 & 54.5 & 73.1 & 0 & -- & -- & 3.8 \\
 & \budgeted~+ \voigate & 1.61 & 16.1 & 50.3 & 64.6 & $-4.6$$^{*}$ & 2.1 & 21 & 4.5 \\
 & \budgetedmgr & 1.47 & 13.5 & 51.5 & 76.1 & $-1.5$\phantom{$^{*}$} & -- & -- & 4.2 \\
\midrule
3 & \budgeted & 2.14 & 20.6 & 56.2 & 71.4 & 0 & -- & -- & 2.6 \\
 & \budgeted~+ \voigate & 2.14 & 21.4 & 51.4 & 63.7 & $-4.7$$^{*}$ & 2.8 & 22 & 3.6 \\
 & \budgetedmgr & 1.93 & 18.1 & 54.3 & 74.3 & $-0.6$\phantom{$^{*}$} & -- & -- & 2.2 \\
\midrule
\multicolumn{10}{@{}l}{\textit{\budgeted~+ \voigate, paired differences, $N=1$--3 pooled}} \\
\multicolumn{2}{@{}l}{Minus \budgeted} & $-0.02$\phantom{$^{*}$} & $+3.9\%$$^{*}$ & $-4.5$$^{*}$ & $-8.8$$^{*}$ & $-4.6$$^{*}$ & & & \\
\multicolumn{2}{@{}l}{Minus \budgetedmgr} & $+0.14$$^{*}$ & $+18.6\%$$^{*}$ & $-2.3$$^{*}$ & $-11.2$$^{*}$ & $-3.8$$^{*}$ & & & \\
\bottomrule
\end{tabular}
\caption{\textbf{VoI gate under the same budget.} \budgeted plus the enforced \voigate at $c=0.27$; three GPT models, four domains, three trials, 3,438 sessions per budget. \budgeted and \budgetedmgr rows: main evaluation, three trials, same tasks. $R,A$: main display scales; value: 0--100; $\Delta R$: policy minus cost-matched \budgeted. Vetoed: share of reviews rejecting the draft. Fixed message: $\tau$ fallback rate. $^{*}$: unadjusted paired 95\% interval excludes zero. Maximum reviews: 8/10/14 at $N=1/2/3$.}
\label{tab:voi-cap}
\label{tab:voi-budget}
\end{table*}
\subsection{Models, Memory, and Call Limits}
\label{app:implementation}
Action generation, review, and memory updates share the served backbone,
route, and reasoning setting. Reasoning models use medium effort.
Models without a configurable reasoning-effort parameter use their native
inference setting. Customer service uses GPT-6-Astra, GPT-5.4, GPT-4.1,
Grok-4.6, Grok-4.3, GPT-OSS-120B, Llama-4-Maverick (17B, 128 experts,
instruct, FP8), and Llama-3.3-70B-Instruct. Research uses Astra, GPT-5.4,
GPT-4.1, Llama-4, and Grok-4.3. Both main comparisons use three trials;
trial counts for supplementary studies are stated with their results. GPT-5.5
simulates users, GPT-5.6-Terra classifies customer-service requests,
and GPT-5.6-Sol judges requests.
Customer-service action output limits are 1,024 tokens for WebShop and
2,048 for $\tau$; auxiliary calls allow 2,400 tokens and 128 calls per session.

Memory $M=(P,s,I)$ contains verbatim user records, a revisable summary,
and observation links. Review receives memory, selected evidence, and
remaining budget; it approves, substitutes, or blocks requests. Tool actions
and final answers bypass review. \textbf{\budgetedsummary} removes user records;
\textbf{\budgetedreview} removes all persistent memory. Review evidence permits 6,000 tokens
(reached in every $\tau$ review and 2--10\% of WebShop reviews); memory
permits 3,072 including serialization, measured with \texttt{cl100k\_base}.
Records share this capacity. User records are appended as replies arrive;
the summary and evidence links are updated within self-review.
\budgetedreview retains question self-review, the evidence window, and the
budget, but carries no memory between reviews. \budgetedsummary retains the
summary and evidence links without verbatim user records.
Appendix~\ref{app:agent-preview-provenance} reports the memory ablations.

\subsection{Relationship to Prior Evaluations}
\label{app:related-evaluation}
Table~\ref{tab:related-evaluation} simplifies several distinctions.
\emph{Limit}: $\tau$-bench caps all agent actions, AskBench caps turns,
SpecBench caps queries, and VoI includes fixed-round baselines; we cap
only discretionary requests and exempt required-only exchanges.
\emph{Content}: AskBench penalizes untargeted clarification during
training and measures redundant asking, but does not report a burden
cost; we combine count, reading effort, and judged question value in an
evaluation cost. \emph{Matched}: a shared cap or per-query price alone
does not compare policies at equal cost; we compare them at equal
expected cost and at equal request count. \emph{Judge} and
\emph{Burden}: $\pi$-Bench audits its evaluator and distinguishes turns
from burden; we report judge validation and a separate transcript-burden
study. Our contribution is the evaluated combination of these elements;
prior work already studies query limits, interaction costs, and useful
clarification. Neither evaluator audits nor transcript judgments
establish experienced human burden during live interaction.

\section{Request Classification and Budget Accounting}
\label{app:request-control}
\label{app:enforcement}
GPT-5.6-Terra classifies requests from the policy and conversation.
Required-only authentication/confirmations bypass review and incur no charged
cost; any discretionary component makes a mixed message discretionary.
Substitutions are reclassified from final text. BrowseComp charges every ASK.
A 100-case human audit finds 91.0\% mean classifier--human agreement,
compared with 90.0\% human--human agreement
(Appendix~\ref{app:classifier-validation}). This audit examines application
of the request rules; readability/value judge validation is a separate study.

\paragraph{Enforcement.}
The main-text protocol states $N$ in the agent's prompt for every domain
and caps delivered messages identically across budgeted methods. On
$\tau$, the budget notice accompanies the benchmark's native instructions;
self-review also receives the remaining budget. At exhaustion, proposed requests are
suppressed; a system notice instructs proceeding with available information,
allowing two retries. Continued requests receive a fixed refusal in context,
at most twice per turn, without reaching the simulator. Persistent asking
produces \textit{``I'm unable to make further progress with the information
currently available''} instead. \noask uses $N=0$. WebShop refuses excess
ASK actions; \noask has no ASK.

The initial $\tau$ budget notice states the question limit, distinguishes
optional requests from required policy exchanges, and instructs the agent
to spend questions where the answer changes its actions. Its full text is
reproduced in Appendix~\ref{app:prompts}.

The exhaustion notice begins ``You have no remaining allowance for
optional questions to the user in this conversation'' and directs the
agent to proceed with available information, policy defaults, and judgment.
The fixed context reply is ``I'm sorry, I don't have time to answer more
questions. Please proceed with the information I've already given you,
using reasonable defaults and your best judgment.''

\paragraph{Message content.}
A message contains about two atomic questions
(Table~\ref{tab:accounting-levels}) and, in WebShop, asks for about two
withheld requirements. Compound requests obtain
1.7 items per session at $N=1$ versus 2.1 unrestricted. Reading-effort
cost penalizes demanding requests; the cap counts messages. Unrestricted
requests are 76\% clarifications, 5\% confirmations, and 19\% suggestions;
at $N=1$, 94\% are clarifications, chiefly removing later requests.

\section{User-Attention Scoring and Statistical Analysis}
\label{app:evaluation-metrics}
\label{app:scoring}

\begin{table}[t]
\centering
\footnotesize
\setlength{\tabcolsep}{4pt}
\begin{tabular}{@{}lp{0.68\columnwidth}@{}}
\toprule
\textbf{Symbol} & \textbf{Meaning} \\
\midrule
$\tau$, $t$ & Interaction session; step index \\
$\pi$ & Policy: ask, call a tool, or respond \\
$R(\tau)$ & Task reward in $[0,1]$ \\
$q$ & Delivered discretionary request message \\
$Q(\tau)$ & All such requests in $\tau$ \\
$N$ & Interruption budget, $|Q(\tau)|\le N$ \\
$r(q)$, $e(q)$ & Readability in $[0,20]$; effort $1-r/20$ \\
$v(q)$, $w(q)$ & Value in $[0,10]$; avoidability $1-v/10$ \\
$\kappa_{\mathrm{eff}}$, $\kappa_{\mathrm{val}}$ & Surcharge weights (1 in main results) \\
$A_{\mathrm{full}}(\tau)$ & User-attention cost (Eq.~\ref{eq:episode-cost}) \\
$\lambda$ & Reward given up per unit of cost \\
$U_\lambda(\tau)$ & Utility $R(\tau)-\lambda A_{\mathrm{full}}(\tau)$  \\
$\lambda^*$ & Break-even $\lambda$ between two policies \\
$Q_{<t}$, $b_t$ & Delivered before $t$; $b_t=N-|Q_{<t}|$ \\
$M=(P,s,I)$ & MGR memory: user records, summary, links \\
\bottomrule
\end{tabular}
\caption{Notation for Sections~\ref{sec:task-formulation} and~\ref{sec:methods}.}
\label{tab:notation}
\end{table}

\paragraph{Scoring and aggregation.}
GPT-5.6-Sol rates readability $r\in[0,20]$ and prospective value $v\in[0,10]$
with fixed domain rubrics (Appendix~\ref{app:prompts}). Readability adapts
CoReEval \citep{ouedraogo2025human} to complete requests; value uses
pre-question context, excluding replies/outcomes. Unit surcharges give
$1+(1-r/20)+(1-v/10)$ cost units per delivered discretionary message.
For example, readability 16/20 and value 8/10 give
$1+0.2+0.2=1.4$ units. These weights define a comparison scale, not
measured time or total conversation burden.
Official $\tau^2$ checks, WebShop complete-goal purchase scores, and
BrowseComp answer correctness determine reward separately.

Outcomes average within tasks/domains, then equally over four domains and
eight models; each has three trials. Budgeted summaries average $N=1$--3.
Research separately averages five models, three trials, and $N=1$--8.
Display scales are $100R$, $10\bar v$, $100A_{\mathrm{full}}/15$, and
$100(U_{0.03}+0.45)/1.45$. The 15-unit reference is not a ceiling.
The valuation 0.03 was chosen after evaluation
(Appendix~\ref{app:reward-per-attention}). Value averages delivered questions
equally within each cell; no-question sessions supply no scores and missing
scores are not zero. Value excludes cells with fewer than ten questions
per trial in any condition (Llama-3.3 Airline), a rule chosen after inspection.

\paragraph{Inference.}
Paired bootstraps use 10,000 task draws within domains, group related tasks
by parent, and retain conditions, models, budgets, and trials together.
Intervals are pointwise 95\%; Table~\ref{tab:headline-family} separately
Holm-adjusts 23 headline tests. Model/domain and alternative-cost diagnostics
are retrospective. An interval spanning zero does not establish equivalence.
Matching details appear in Appendix~\ref{app:agent-preview-robustness}.
Retained-asking-gain ratios use approximate Fieller intervals
\citep{fieller1954some}, joint bootstrap covariance, and a normal critical
value. Nonpositive-denominator draws remain; unbounded/disjoint sets have
no finite error bar. This replaced positive-gain filtering after inspecting
Airline.
\subsection{Human Validation of the Judge}
\label{app:human-validation}
\label{sec:human-evaluation}
Two development rounds precede frozen guidelines, prompts, and tie rules.
Held-out validation uses 80 same-task request pairs per dimension: 40
$\tau$ and 40 WebShop, each with 24 score-gap-stratified and 16 near-tie/
challenging pairs. Two annotators independently choose the more readable
or justified request, allowing ties, using the judge's message/context
\citep{zheng2023judging,liu2023g}. Model/policy identities,
scores, replies, and outcomes are hidden; presentation is randomized.
Judge gaps of at most one raw point are ties. Agreement averages agreement
with both annotators over all A/B/tie labels, without adjudication;
intervals are task-clustered bootstraps (Table~\ref{tab:human-validation}).
Readability/value agreement is 90.0\%/90.0\%, versus 92.5\%/87.5\%
between humans. This supports comparative scores, not cost weights or
experienced-burden units. BrowseComp is not separately validated.
The coordinator could access judge labels; annotator files were blind.
The classifier audit and the burden study use different annotators. (Appendix~\ref{app:burden-study}).
\begin{table}[t]
\centering
\footnotesize
\setlength{\tabcolsep}{3pt}
\begin{tabular}{@{}lcc@{}}
\toprule
Comparison & Readability & Question value \\
\midrule
Human--human & 92.5 [86.1, 97.6] & 87.5 [79.0, 94.9] \\
Judge--annotator (mean) & 90.0 [84.0, 95.6] & 90.0 [84.2, 95.1] \\
Second judge (mean) & 71.2 [61.3, 80.6] & 76.9 [68.5, 85.0] \\
\midrule
\multicolumn{3}{@{}l}{\textit{Judge--annotator agreement by benchmark}} \\
$\tau$-bench ($n=40$) & 90.0 & 93.8 \\
WebShop ($n=40$) & 90.0 & 86.2 \\
\bottomrule
\end{tabular}
\caption{\textbf{Human validation on the main customer-service benchmarks.} Agreement (\%) on 80 held-out request pairs per dimension, with task-clustered 95\% intervals. Labels are A/B/tie; score differences of at most one raw point imply a judge tie. Judge--annotator agreement averages agreement with each of the two annotators. The second-judge row is Grok-4.6's agreement with the annotators on the same pairs (Appendix~\ref{app:second-judge}).}
\label{tab:human-validation}
\end{table}
\subsection{Human Validation of Request Classification}
\label{app:classifier-validation}
Two annotators, different from those who validated the judge and those in the burden study, labeled 100 cases after pilot rounds.
Each case supplies the preceding conversation and applicable domain
policy. Annotators assign one of three labels: \emph{not a request},
\emph{required only}, or \emph{discretionary}. Any optional component
makes a mixed request discretionary. Both annotators labeled every case;
pilot cases are excluded from the reported sample.

Cases are sampled from archived main-evaluation logs by domain and
classifier-label stratum: 32 each from Retail, Airline, and Telecom,
and four from WebShop. The classifier strata contain 46 discretionary,
42 required-only, and 12 non-request cases. Agreement therefore describes
this stratified sample, rather than prevalence-weighted accuracy over
all messages. Four WebShop cases do not establish domain-specific
reliability; BrowseComp-Plus outside this audit.

Human--human agreement is 90.0\% ($\kappa=0.84$). Classifier--human
agreement averages 91.0\% (95\% CI [86.5, 95.5]; mean $\kappa=0.85$)
across the two annotators (Table~\ref{tab:classifier-validation}).
Collapsing labels to discretionary versus non-discretionary preserves
these agreement rates, with $\kappa=0.80$ between humans and mean
$\kappa=0.82$ for the classifier. Thus all observed disagreements affect
whether a request is charged. This provides empirical support for the
classifier as an operational measure of discretionary requests, while
retaining classification uncertainty in the resulting counts and costs.
The audit examines application of the stated rules, rather than whether
those rules capture all user effort or whether policy comparisons are
robust to classification errors.

\begin{table}[t]
\centering
\footnotesize
\setlength{\tabcolsep}{3pt}
\begin{tabular}{@{}lcc@{}}
\toprule
Comparison & Agreement (\%) & Cohen's $\kappa$ \\
\midrule
\multicolumn{3}{@{}l}{\textit{Three labels}} \\
Human--human & 90.0 [84.0, 95.0] & 0.84 [0.73, 0.92] \\
Classifier--A1 & 90.0 [84.0, 95.0] & 0.84 [0.74, 0.92] \\
Classifier--A2 & 92.0 [87.0, 97.0] & 0.87 [0.77, 0.95] \\
Classifier--human & 91.0 [86.5, 95.5] & 0.85 [0.77, 0.92] \\
\midrule
\multicolumn{3}{@{}l}{\textit{Discretionary versus non-discretionary}} \\
Human--human & 90.0 [84.0, 95.0] & 0.80 [0.67, 0.90] \\
Classifier--A1 & 90.0 [84.0, 95.0] & 0.80 [0.67, 0.90] \\
Classifier--A2 & 92.0 [87.0, 97.0] & 0.84 [0.72, 0.94] \\
Classifier--human & 91.0 [86.5, 95.5] & 0.82 [0.72, 0.91] \\
\bottomrule
\end{tabular}
\caption{\textbf{Human validation of request classification.} Two annotators
label the same 100 cases. Classifier--human rows average the two annotator
comparisons, including their separate $\kappa$ values. Brackets are 95\%
case-bootstrap intervals (2,000 draws), retaining both
annotators' labels in each sampled case.}
\label{tab:classifier-validation}
\end{table}

\subsection{A Judge from Another Model Family}
\label{app:second-judge}
Grok-4.6 re-scores 1,989 fixed-hash sampled questions, up to 21 per
model/domain/policy (Unrestricted, \budgeted, \budgetedmgr), using the same
prompts. Rank correlations are 0.69 for value
and 0.85 for effort (Table~\ref{tab:second-judge}). Both judges estimate
$-1.6$ effort points for \budgetedmgr versus \budgeted, but value gains
are $+3.5$ under the paper's judge and an uncertain $+0.7$ under Grok.
Relative own-vendor value advantages are an uncertain $+0.8$ for the first
judge and $+3.4$ for the second.

We also re-scored the 80 held-out human-annotated pairs per rubric of
Table~\ref{tab:human-validation} ($\tau$-bench and WebShop) with
Grok-4.6, using identical inputs, prompts, and tie rule. By a rule fixed
in advance, a substitute judge counts as less human-aligned if the 95\%
interval of the agreement difference (paper's judge minus substitute)
lies above zero. On these pairs it does on both rubrics: $+13.1$
[$+5.3$, $+21.5$] for question value and $+18.8$ [$+11.0$, $+27.6$] for
readability. The
paper's judge agrees with annotators about as often as annotators agree
with each other (question value: 90.0\% vs.\ 87.5\%), while Grok agrees
less (76.9\%). Of Grok's 37 disagreements among the 160 judge--annotator
comparisons on question value, 35 are tie-versus-side and two are
reversals. The gap is concentrated on $\tau$-bench ($+22.5$ [$+10.3$,
$+36.6$]), which is also where the judges disagree about the \mgragent's
value gain on the 1,989-question sample ($+4.6$ vs.\ $+1.4$). On
WebShop, where no agreement difference is detectable ($+3.8$ [$-5.0$,
$+13.1$]), neither judge finds a gain.

This comparison favors the paper's judge in two ways: the rubrics were
developed with it on development pairs, and the held-out pairs were
stratified by its score gaps. The result therefore shows that a judge
used without calibration is less reliable, not that one model family is
better. We report value under the calibrated judge and recommend checking
any substitute judge against the released human pairs before use.

\paragraph{Vendor preference.}
Because the paper's judge shares a vendor with four of the eight
backbones, we tested whether it favors OpenAI-backbone requests relative
to annotators, using the 80 held-out pairs per rubric, with group
assignments and a decision rule saved before any outcome was computed.
These pairs come from earlier runs with partly different backbones
(OpenAI: GPT-5.5, GPT-5.2, GPT-4.1, GPT-OSS-20B and 120B; others:
Grok-4.3, Grok-4.20, Llama-4-Mav, Llama-3.3-70B). Judge--annotator
agreement does not differ detectably between OpenAI-backbone and other
pairs for question value (89.5\% vs.\ 90.5\%; $-1.0$ [$-11.9$, $+9.6$])
or readability (90.9\% vs.\ 87.5\%; $+3.4$ [$-13.2$, $+21.4$]), although
gaps of ten points or more are not excluded. Both requests in each
question-value pair share a backbone, so a direct preference test is
possible only for readability. On its 42 mixed-vendor pairs, annotators
already prefer the OpenAI-backbone request (net preference $+0.38$ on a
$-1$ to $+1$ scale), and the judge does so more strongly ($+0.50$; excess
$+0.12$ [$+0.03$, $+0.23$]); all 8 judge--annotator disagreements among
the 84 comparisons favor the OpenAI request. Without GPT-OSS pairs, the
excess is $+0.05$ [$0.00$, $+0.14$] on 21 pairs. By the pre-specified
rule, this is a readability bias toward OpenAI backbones. Every policy
contrast in the paper compares methods on the same backbone, so a
backbone-level bias should largely cancel within contrasts. Consistent
with this, the two judges score vendors differently
(Table~\ref{tab:second-judge}) yet estimate the same $-1.6$ effort
change for \budgetedmgr versus \budgeted. The bias can favor OpenAI
backbones wherever attention-cost levels are compared across models: the
horizontal positions of models in Figures~\ref{fig:rq1-reward-attention}
and~\ref{fig:rq1-reward-attention-all}, the per-unit yields in
Appendix~\ref{app:valuations}, and the model-specific
preferred budgets in Table~\ref{tab:best-budget}. These comparisons depend
on the judged surcharges. Counting questions only, which uses no judge,
leaves the cost order of 237 of 252 model pairs across the nine policy
settings unchanged. Of the 15 reversals, 11 are between an OpenAI and
another vendor's backbone, and in 8 of these the OpenAI backbone asks
more questions but has the lower full cost (for example, GPT-4.1 versus
Grok-4.3 under \budgeted at each of $N=1$--3). Because annotators also
judge OpenAI-backbone requests more readable, these reversals need not
reflect judge bias, but cross-model cost orderings within these pairs
should not be treated as established.
\begin{table}[t]
\centering
\footnotesize
\setlength{\tabcolsep}{2pt}
\begin{tabular}{@{}l rr@{}}
\toprule
 & Question value & Reading effort \\
\midrule
\multicolumn{3}{@{}l}{\textit{Agreement on the same questions}} \\
\quad Questions & 1989 & 1989 \\
\quad Mean, first judge & $67.0$ & $16.2$ \\
\quad Mean, second judge & $69.1$ & $27.1$ \\
\quad Rank correlation & $0.69$ & $0.85$ \\
\quad Mean abs.\ difference & $14.1$ & $11.3$ \\
\midrule
\multicolumn{3}{@{}l}{\textit{\budgetedmgr minus \budgeted}} \\
\quad First judge & $+3.5$ {\scriptsize[$+0.8$, $+6.2$]} & $-1.6$ {\scriptsize[$-3.0$, $-0.3$]} \\
\quad Second judge & $+0.7$ {\scriptsize[$-1.5$, $+3.0$]} & $-1.6$ {\scriptsize[$-3.1$, $-0.0$]} \\
\quad Difference & $+2.8$ {\scriptsize[$+0.5$, $+5.1$]} & $-0.0$ {\scriptsize[$-0.9$, $+0.8$]} \\
\midrule
\multicolumn{3}{@{}l}{\textit{Own-vendor backbones minus others}} \\
\quad First judge (OpenAI) & $+0.8$ {\scriptsize[$-1.1$, $+2.8$]} & $+0.8$ {\scriptsize[$+0.0$, $+1.6$]} \\
\quad Second judge (xAI) & $+3.4$ {\scriptsize[$+1.1$, $+5.7$]} & $+0.9$ {\scriptsize[$+0.1$, $+1.8$]} \\
\bottomrule
\end{tabular}
\caption{\textbf{Cross-family judge sensitivity.} Fixed sample of 1,989 questions scored by GPT-5.6-Sol and Grok-4.6 with identical prompts; scores 0--100. Policy effects weight models/domains equally; brackets: 95\% sampled-question bootstrap intervals. Own-vendor contrasts measure relative preference for same-vendor backbones against the other judge.}
\label{tab:second-judge}
\end{table}

\subsection{Sensitivity to Request Accounting}
\label{app:accounting-sensitivity}
We evaluate alternative cost definitions on the same recorded
conversations, without rerunning agents or obtaining new judge scores.
The analysis covers 91,500 sessions across eight models, four domains,
and three trials, including \noask, Unrestricted, \frugal, the original
\voigate, \budgeted, and \budgetedmgr. We apply the same task
exclusions as in the main evaluation and average trials within each task.
For uncertainty estimates, we use 10,000 paired bootstrap draws,
resampling parent tasks independently within each domain and keeping
their observations across models, policies, budgets, and trials together.
Models and domains receive equal weight; summaries across budgets also
weight $N=1,2,3$ equally. Reported 95\% intervals are pointwise and
exploratory, without adjustment for multiple comparisons.

Let $E(\tau)$ count delivered policy-exempt requests and $k(q)$ be the
classifier's number of component questions in delivered discretionary
message $q$. We evaluate
\[
\begin{aligned}
 A_{\mathrm{atomic}}(\tau)&=\sum_{q\in Q(\tau)}k(q),\\
 A_{\mathrm{expanded}}(\tau)&=A_{\mathrm{full}}(\tau)+E(\tau).
\end{aligned}
\]
Atomic cost contains no reading-effort or value surcharge. When a
component count is missing or zero, we assign one component; this
affects 210 delivered discretionary requests in the analysis.
Bundled messages have $k(q)>1$. These remain classifier estimates,
not independent human question counts. Expanded cost charges one unit
per exempt request, without scoring its content. It still excludes
informational updates and other conversation demands. Neither view
estimates behavior under a newly enforced atomic or all-request budget.

\begin{table*}[t]
\centering\footnotesize\setlength{\tabcolsep}{5pt}
\begin{tabular}{@{}lrrrr@{}}
\toprule
Cost definition & \budgeted, $N=1$ & \budgeted, $N=2$ & \budgeted, $N=3$ & \mgr, mean $N=1$--3 \\
\midrule
Messages & 70.4 [68.9, 71.7] & 49.0 [46.7, 51.1] & 33.2 [30.6, 35.7] & 10.9 [9.9, 12.0] \\
Atomic questions & 67.8 [66.1, 69.3] & 47.1 [44.8, 49.4] & 31.7 [29.2, 34.0] & 11.4 [10.4, 12.5] \\
Count + reading effort & 71.0 [69.5, 72.4] & 49.9 [47.6, 52.1] & 34.1 [31.5, 36.5] & 11.3 [10.3, 12.4] \\
Full discretionary cost & 73.0 [71.6, 74.3] & 51.9 [49.7, 54.0] & 35.7 [33.1, 38.2] & 12.7 [11.6, 13.8] \\
Full cost + exempt requests & 37.7 [35.7, 39.8] & 26.5 [24.5, 28.5] & 17.3 [15.3, 19.2] & 4.5 [3.7, 5.4] \\
\bottomrule\end{tabular}
\caption{\textbf{Accounting sensitivity: cost saved (\%).} Eight models, four domains, three trials. \budgeted is compared with Unrestricted; the \mgr column compares \budgetedmgr with \budgeted at the same allowance. Brackets: exploratory paired parent-task bootstrap 95\% intervals (10,000 draws). Savings are ratios of equally weighted model/domain/budget means. These rescore fixed trajectories; the enforced message budgets are unchanged.}
\label{tab:accounting-sensitivity}
\end{table*}
\begin{table*}[t]
\centering\footnotesize\setlength{\tabcolsep}{5pt}
\begin{tabular}{@{}lrrrr@{}}
\toprule
Policy & Messages & Atomic questions & Exempt requests & Bundled (\%) \\
\midrule
Unrestricted & 2.84 [2.68, 2.99] & 5.39 [5.06, 5.72] & 3.44 [3.20, 3.67] & 52.5 [51.0, 54.0] \\
\frugal & 2.43 [2.30, 2.56] & 4.20 [3.95, 4.46] & 3.66 [3.42, 3.90] & 47.6 [46.1, 49.1] \\
\voigate (original) & 2.91 [2.74, 3.09] & 4.56 [4.29, 4.84] & 3.67 [3.43, 3.90] & 35.0 [33.7, 36.2] \\
\budgeted, $N=1$ & 0.84 [0.83, 0.85] & 1.74 [1.69, 1.79] & 3.70 [3.46, 3.94] & 64.1 [62.4, 65.8] \\
\budgeted, $N=2$ & 1.45 [1.42, 1.48] & 2.85 [2.75, 2.95] & 3.65 [3.41, 3.89] & 57.7 [56.1, 59.2] \\
\budgeted, $N=3$ & 1.90 [1.85, 1.94] & 3.68 [3.53, 3.83] & 3.66 [3.42, 3.90] & 55.2 [53.5, 56.8] \\
\budgeted, mean $N=1$--3 & 1.39 [1.37, 1.42] & 2.75 [2.66, 2.85] & 3.67 [3.43, 3.90] & 57.8 [56.4, 59.2] \\
\budgetedmgr, mean $N=1$--3 & 1.24 [1.21, 1.28] & 2.44 [2.35, 2.53] & 3.67 [3.43, 3.90] & 57.9 [56.6, 59.2] \\
\bottomrule\end{tabular}
\caption{\textbf{Delivered requests per session and bundling.} Same panel and exploratory intervals as Table~\ref{tab:accounting-sensitivity}. Atomic counts use the existing classifier; a bundled message contains more than one estimated component question. The bundled share divides aggregate bundled-message count by aggregate discretionary-message count. Required-only exchanges are exempt in the primary protocol.}
\label{tab:accounting-levels}
\end{table*}

\paragraph{Savings persist, but their size depends on what is charged.}
At the same allowance, \mgr saves 10.9\% of messages
($[9.9,12.0]$), 11.4\% of atomic questions ($[10.4,12.5]$), and
12.7\% of full discretionary cost ($[11.6,13.8]$).
Bundled messages account for 57.8\% of \budgeted and 57.9\% of
\budgetedmgr messages in the aggregate (Table~\ref{tab:accounting-levels}).
Exempt requests average 3.671 and 3.672 per session, respectively;
the paired difference is $+0.001$ ($[-0.043,+0.044]$), without an
equivalence claim. Adding these requests reduces \mgr savings to
4.5\% ($[3.7,5.4]$). The change from full discretionary accounting is
$-8.1$ percentage points ($[-9.2,-7.0]$), computed within each paired
resample. \budgeted also continues to save under expanded accounting,
by 37.7/26.5/17.3\% for $N=1/2/3$ versus Unrestricted.

\paragraph{Content accounting changes magnitudes without establishing a superior burden scale.}
For \mgr, full cost raises aggregate savings over message-only
accounting by 1.8 percentage points ($[1.5,2.0]$). It can also lower
estimated savings: Grok-4.6 on Telecom at $N=1$ saves 17.4\% by
message count but 8.1\% by full cost, a paired change of $-9.2$
points ($[-14.0,-4.8]$). These comparisons test changes in the
accounting estimates, not differences in experienced human burden.
Among the 96 model/domain/budget \mgr cells, seven point estimates
change sign between count and full cost; all seven have intervals
spanning zero under both definitions. Thus this panel supports
changes in magnitude, rather than a robust reversal of policy ordering.

\section{Agent Analyses and Robustness}
\label{app:agent-preview}


\subsection{Comparisons at Matched Count and Cost}
\label{app:review-controls}
\label{app:agent-preview-robustness}
Within each model/domain, we mix adjacent policies among
\noask, \budgeted at $N=1$--3, and Unrestricted to match a target policy's expected
count or cost. The mixture randomly chooses a budget per session.
Reward is interpolated directly; value sums and counts are interpolated
separately and divided. Matching averages does not match individual
sessions. The \mgragent's equal-cost reward difference is $+0.2$ points
($[-0.4,+0.8]$), not established equivalence; judged value gains 1.7
points (Table~\ref{tab:review-frontier}). Memory improves equal-cost
reward over \budgetedreview by 0.9 points, concentrated in Astra/Grok-4.6
(4.0/3.1). The all-mixture upper envelope gives a $-1.0$-point estimate
($-1.6$ averaged over bootstrap draws), reflecting maximization over
noisy means; the memory contrast remains about one point.

Count matching retains 93 quality-eligible comparisons. One exceeds
observed support (Astra/WebShop/$N=3$: count 1.829 versus maximum 1.808).
The supported 92 yield a 1.80-point value gain ($[1.31,2.34]$), versus
1.81 ($[1.32,2.35]$) overall, with mean mismatch 0.000217 messages.
Bootstrap endpoint fallback remains when draws exceed support.
Matching on count plus reading effort or on full cost gives gains of
1.7 points ($[1.2,2.3]$ and $[1.2,2.2]$). These retrospective
comparisons test matching-resource sensitivity, not independent judge
validity.

\begin{table*}[t]
\centering
\footnotesize
\setlength{\tabcolsep}{7pt}
\newsavebox{\rqfrontierbox}
\sbox{\rqfrontierbox}{%
\begin{tabular}{@{}l cccccccc@{}}
\toprule
 & Budgeted & Budgeted & Budgeted + MGR & Budgeted & Budgeted & Frugal & VoI & Budgeted \\
 & + MGR & + Review & $-$ Budgeted + Review & + MGR & + MGR & prompt & gate & + MGR \\
 & & & & (count only) & (envelope) & & & (question value) \\
\midrule
\multicolumn{9}{@{}l}{\textit{By domain, mean over the eight models}} \\[1pt]
Mean of four domains & $+0.2$ & $-0.7$ & $+0.9$ & $+0.1$ & $-1.0$ & $-0.8$ & $-1.4$ & $+1.7$ \\
 & {\scriptsize[$-0.4$, $+0.8$]} & {\scriptsize[$-1.3$, $-0.1$]} & {\scriptsize[$+0.4$, $+1.4$]} & {\scriptsize[$-0.5$, $+0.7$]} & {\scriptsize[$-2.3$, $-0.9$]} & {\scriptsize[$-1.6$, $+0.1$]} & {\scriptsize[$-2.5$, $-0.3$]} & {\scriptsize[$+1.2$, $+2.2$]} \\[3pt]
Retail & $+0.7$ & $-1.4$ & $+2.0$ & $+0.5$ & $-0.3$ & $-0.3$ & $-3.3$ & $+2.4$ \\
 & {\scriptsize[$-0.6$, $+2.0$]} & {\scriptsize[$-2.6$, $-0.1$]} & {\scriptsize[$+0.9$, $+3.2$]} & {\scriptsize[$-0.7$, $+1.8$]} & {\scriptsize[$-2.3$, $+0.5$]} & {\scriptsize[$-2.1$, $+1.5$]} & {\scriptsize[$-5.4$, $-1.2$]} & {\scriptsize[$+1.6$, $+3.2$]} \\[3pt]
Airline & $+0.2$ & $-0.2$ & $+0.4$ & $0.0$ & $-1.5$ & $-1.7$ & $-2.2$ & $+3.8$ \\
 & {\scriptsize[$-1.5$, $+1.9$]} & {\scriptsize[$-1.8$, $+1.4$]} & {\scriptsize[$-1.0$, $+1.8$]} & {\scriptsize[$-1.7$, $+1.7$]} & {\scriptsize[$-4.8$, $-0.4$]} & {\scriptsize[$-4.2$, $+0.8$]} & {\scriptsize[$-5.5$, $+1.2$]} & {\scriptsize[$+2.4$, $+5.3$]} \\[3pt]
Telecom & $+1.5$ & $-0.3$ & $+1.8$ & $+1.5$ & $-0.2$ & $+0.9$ & $+0.4$ & $-0.3$ \\
 & {\scriptsize[$+0.7$, $+2.4$]} & {\scriptsize[$-1.4$, $+0.8$]} & {\scriptsize[$+0.9$, $+2.8$]} & {\scriptsize[$+0.7$, $+2.3$]} & {\scriptsize[$-1.6$, $+0.4$]} & {\scriptsize[$-0.4$, $+2.4$]} & {\scriptsize[$-1.1$, $+1.7$]} & {\scriptsize[$-1.2$, $+0.7$]} \\[3pt]
WebShop & $-1.7$ & $-0.9$ & $-0.8$ & $-1.7$ & $-2.2$ & $-2.0$ & $-0.6$ & $+0.8$ \\
 & {\scriptsize[$-2.4$, $-1.1$]} & {\scriptsize[$-1.7$, $-0.4$]} & {\scriptsize[$-1.3$, $-0.1$]} & {\scriptsize[$-2.4$, $-1.1$]} & {\scriptsize[$-3.0$, $-1.7$]} & {\scriptsize[$-3.1$, $-0.9$]} & {\scriptsize[$-1.6$, $+0.4$]} & {\scriptsize[$+0.2$, $+1.7$]} \\[3pt]
\bottomrule
\end{tabular}
}
\ifdim\wd\rqfrontierbox>\textwidth
\resizebox{\textwidth}{!}{\usebox{\rqfrontierbox}}
\else\usebox{\rqfrontierbox}\fi
\caption{\textbf{Reward and value at equal expected user-attention cost.} Differences from adjacent \budgeted mixtures; envelope allows nonadjacent mixtures. Count-only changes the matching resource. Reward/value differences are 0--100 points. Domain means average $N=1$--3 over the eight models; Figure~\ref{fig:rq2-budget-gains}(c) shows equal-cost reward by model. Value excludes Llama-3.3 Airline. Brackets: pointwise paired 95\% task-bootstrap intervals. Envelope maximization can bias estimates upward.}
\label{tab:review-frontier}
\end{table*}

\begin{figure}[t]
    \centering
    \includegraphics[width=.9\columnwidth]{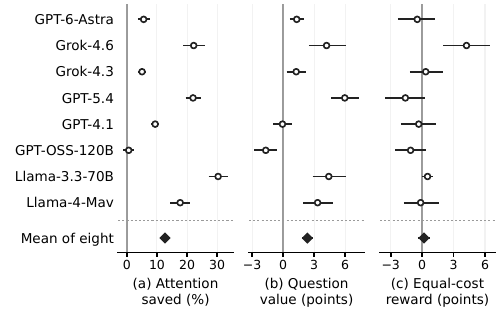}
    \caption{\textbf{Attention savings and question-value gains are
    widespread; reward gains at equal attention cost vary by model.}
    \budgetedmgr versus \budgeted, across $N=1$--3
    and four domains, with paired 95\% task-bootstrap intervals.
    (a) Attention cost saved and (b) change in judged question value
    at the same interruption budget. (c) Reward difference from
    a \budgeted comparator matched on expected attention cost.
    Positive values favor the \mgragent; diamonds summarize all eight
    models. Means: Table~\ref{tab:paired-effects-full}; by domain:
    Table~\ref{tab:review-frontier}.}
    \label{fig:rq2-budget-gains}
\end{figure}

\paragraph{Frequency and content.}
\mgr saves 10.9\% of messages, 11.3\% of count plus effort, and 12.7\%
of full cost (Table~\ref{tab:accounting-sensitivity}). Content can
amplify or reduce apparent savings. A symmetric accounting decomposition
assigns about 85\% of the full-cost change to frequency and 15\% to mean
per-message cost (1.474 to 1.445 units); this is arithmetic, not causal
mediation.

In 58.7\% ($[56.9,60.5]$) of paired sessions, both methods deliver the
same positive count. Conditional on this outcome, \budgetedmgr lowers
per-request cost by 1.4\%, raises value by 1.1 points, and changes reward
by $-0.14$ ($[-0.88,+0.62]$).
This is descriptive, not a causal estimate. \mgr removes 20\% of
questions valued at most 4 versus 5\% valued at least 8
(Table~\ref{tab:value-levels}); the contrast holds across domains and
$N=2,3$, but reverses for GPT-OSS-120B.

\begin{table}[t]
\centering
\footnotesize
\setlength{\tabcolsep}{3pt}
\newsavebox{\rqvaluelevelsbox}
\sbox{\rqvaluelevelsbox}{%
\begin{tabular}{@{}l rr r@{}}
\toprule
Questions per session & \budgeted & \budgetedmgr & Change (\%) \\
\midrule
Scored 4 or below & $0.32$ & $0.26$ & ${-}20.2$ {\scriptsize[${-}22.9$, ${-}17.3$]} \\
Scored 5--7 & $0.25$ & $0.21$ & ${-}16.9$ {\scriptsize[${-}19.1$, ${-}14.6$]} \\
Scored 8 or above & $0.82$ & $0.78$ & ${-}5.4$ {\scriptsize[${-}6.8$, ${-}4.1$]} \\
All questions & $1.39$ & $1.24$ & ${-}10.9$ {\scriptsize[${-}12.0$, ${-}9.9$]} \\
Share scored 4 or below (\%) & $21.5$ & $18.9$ & ${-}2.7$ {\scriptsize[${-}3.4$, ${-}2.0$]} \\
\midrule
\multicolumn{4}{@{}l}{\textit{Change at 4 or below minus change at 8 or above}} \\
\quad \budgetedreview & & & $+3.1$ {\scriptsize[${-}0.6$, $+6.9$]} \\
\quad \budgetedsummary & & & ${-}9.2$ {\scriptsize[${-}12.8$, ${-}5.5$]} \\
\quad \budgetedmgr & & & ${-}14.7$ {\scriptsize[${-}18.0$, ${-}11.4$]} \\
\bottomrule
\end{tabular}
}
\ifdim\wd\rqvaluelevelsbox>\columnwidth
\resizebox{\columnwidth}{!}{\usebox{\rqvaluelevelsbox}}
\else\usebox{\rqvaluelevelsbox}\fi
\caption{\textbf{Delivered questions by judged value.} Low: $v\leq4$; high: $v\geq8$. Counts per session; changes relative to \budgeted, averaging $N=1$--3 and eligible model/domain cells equally. Contrast compares low- versus high-value percentage changes. Brackets: paired 95\% task-bootstrap intervals. Categories use the evaluation judge.}
\label{tab:value-levels}
\end{table}

\subsection{Persistent Memory and Self-Review}
\label{app:agent-preview-ablations}
\label{app:agent-preview-provenance}
\label{app:review-audit}
\label{app:compute-overhead}
\paragraph{Memory ablations.}
The \mgragent (\budgetedmgr) in Tables~\ref{tab:main-results}
and~\ref{tab:equal-cost} uses persistent user records, a revisable task
summary, and question self-review with the remaining budget. User records
are appended when replies arrive; the summary and evidence links are
updated within self-review. \budgetedsummary removes user records but retains
the summary and evidence links. \budgetedreview retains question self-review
and the same evidence window, but carries no memory between reviews.
\budgeted removes both memory and self-review.
Compared with \budgetedreview, \budgetedmgr saves 6.2\% in user-attention
cost and adds 2.9 question-value points. Removing user records alone has
a smaller effect, supporting the contribution of persistent memory.

\paragraph{Computational overhead.}
Our overhead measurement estimates that self-review and memory maintenance
add an average of 1.8 model calls, approximately 15k tokens, and 8 seconds
of model-call latency per session, excluding base-agent computation.

\paragraph{Wider evidence window without memory.}
A three-GPT-model $\tau$ follow-up ($N=1$--3, one trial, 1,998 sessions)
expands the evidence window of \budgetedreview from 6,000 to 9,072 tokens, the nominal
combined evidence/memory allowance. It gains 2.5 value points but still
costs 4.8\% more and scores 3.3 value points below \budgetedmgr
(Table~\ref{tab:matched-context}). Its equal-cost reward deficit is
uncertain overall ($-0.7$), but supported on Telecom ($-4.4$) and
Astra ($-3.4$). On Astra/Telecom, repeated phone-line questions fall
from 36\% to 18\%, versus 1\% with persistent records; 49 of 59 redundant
requests repeat an already supplied number. Longer context contributes
to the effect but does not remove these memory benefits.

\begin{table}[t]
\centering
\footnotesize
\setlength{\tabcolsep}{4pt}
\begin{tabular}{@{}l rr@{}}
\toprule
 & \multicolumn{2}{c}{Wide window minus} \\
\cmidrule(l){2-3}
 & Budgeted & Budgeted \\
 & + Review & + MGR \\
\midrule
User-attention cost (\%) & $-1.0$\phantom{$^{*}$} & $+4.8$$^{*}$ \\
Question value & $+2.5$$^{*}$ & $-3.3$$^{*}$ \\
Reward & $+1.4$\phantom{$^{*}$} & $+0.1$\phantom{$^{*}$} \\
\midrule
\multicolumn{3}{@{}l}{\textit{Reward at equal user-attention cost}} \\
\quad All & $+1.5$\phantom{$^{*}$} & $-0.7$\phantom{$^{*}$} \\
\quad Retail & $+2.7$\phantom{$^{*}$} & $+0.4$\phantom{$^{*}$} \\
\quad Airline & $+1.9$\phantom{$^{*}$} & $+2.0$\phantom{$^{*}$} \\
\quad Telecom & $-0.2$\phantom{$^{*}$} & $-4.4$$^{*}$ \\
\quad GPT-4.1 & $+2.3$\phantom{$^{*}$} & $+0.8$\phantom{$^{*}$} \\
\quad GPT-5.4 & $-0.1$\phantom{$^{*}$} & $+0.7$\phantom{$^{*}$} \\
\quad GPT-6-Astra & $+2.3$\phantom{$^{*}$} & $-3.4$$^{*}$ \\
\bottomrule
\end{tabular}
\caption{\textbf{\budgetedreview with a wider evidence window.} Evidence expands to 9,072 tokens versus 6,000 plus 3,072 memory tokens. Three GPT models, three $\tau$ domains, $N=1$--3, one trial. Cost changes are relative percent; reward/value changes are display points. Wide review reads 12.1k versus 8.8k tokens and blocks 14\% versus 20\% for \budgetedmgr. $^{*}$: unadjusted paired 95\% interval excludes zero.}
\label{tab:matched-context}
\end{table}

\paragraph{Review wording.}
\label{app:review-wording}
Gap-specific and generic benefit/burden review wording perform alike
(GPT-6-Astra and GPT-4.1, $N=1,2$, 20 test tasks in each of five
domains, two trials; 1,600 sessions; memory and limits fixed). Of 16
main-domain utility intervals, 15 span zero, with GPT-4.1/WebShop/$N=1$
favoring generic wording, and all four research intervals span zero.
This isolates wording, not the causal effect of tracking gaps.

\subsection{Separating Tool Evidence from User Preferences}
\label{app:review-task-dependence}
\label{app:review-revision}
\label{app:webshop-rule}
The original WebShop rule mistakes displayed options for known user
preferences: 94.7\% of 3,499 blocked drafts concern withheld requirements,
versus 93.6\% approved and 92.2\% rewritten (eight models, $N=1$--3,
all trials; withheld items identified by the simulator's deterministic
matcher). For example, available waist sizes
do not answer which size the shopper wants. Blocks leave 27\% of sessions
without a question versus 11\% for \budgeted, reducing acquired requirements.

A revised WebShop rule requires grounded block reasons: cite a prior
answer, name a tool returning the answer itself, or explain why no answer
changes actions. Otherwise it approves/substitutes; it cannot block useful
user-held questions merely to reserve budget. Memory prompts/schema/limits
are unchanged. The diagnosis followed test-trace inspection; wording and
WebShop-only scope were chosen on development tasks (59 review replays,
then three-GPT-model runs at $N=1,3$). Applied to every domain on
development tasks, the revision raised WebShop reward by 3.6 points
($[-0.4,+8.0]$) but lowered Retail reward by 9.4 ($[-19.6,+0.7]$), so
the test evaluation changes WebShop only.
This is a post-hoc rule comparison, separate from the main-table results.

On 160 WebShop test tasks, three GPT models, $N=1$--3 and two trials
(5,760 sessions), blocked drafts fall from 29\% to 10\%, zero-question
sessions from 23\% to 7\%, and reward rises 5.2 points (40.0 to 45.2;
Table~\ref{tab:review-revision-test}). Cost increases and value falls 0.8;
equal-cost reward moves from $-3.3$ to an uncertain $-0.4$. A concurrent
original-rule rerun differs from earlier reward by only $+0.5$, with
unchanged cost. Llama follow-ups (960 sessions each, one trial) reduce
blocks to 6\%/12\%, but reward changes of $+1.1/+1.9$ remain uncertain;
Llama-3.3 also loses 2.9 value points. Unblocking helps most where useful
preference acquisition was previously suppressed.

\begin{table*}[t]
\centering
\footnotesize
\setlength{\tabcolsep}{5pt}
\newsavebox{\rqreviewrevisiontestbox}
\sbox{\rqreviewrevisiontestbox}{%
\begin{tabular}{@{}lcccc@{}}
\toprule
 & $N=1$ & $N=2$ & $N=3$ & All budgets \\
\midrule
Reward & $39.8\to44.4$ & $40.4\to45.3$ & $39.9\to45.9$ & $40.0\to45.2$ \\
 & $+4.6$ {\scriptsize [$+2.5$, $+6.7$]} & $+4.9$ {\scriptsize [$+3.1$, $+6.9$]} & $+6.1$ {\scriptsize [$+4.0$, $+8.2$]} & $+5.2$ {\scriptsize [$+3.9$, $+6.5$]} \\[3pt]
Questions per session & $0.69\to0.88$ & $1.11\to1.35$ & $1.41\to1.58$ & $1.07\to1.27$ \\
 & $+0.19$ {\scriptsize [$+0.17$, $+0.22$]} & $+0.23$ {\scriptsize [$+0.18$, $+0.28$]} & $+0.17$ {\scriptsize [$+0.11$, $+0.24$]} & $+0.20$ {\scriptsize [$+0.17$, $+0.23$]} \\[3pt]
Episodes with no question (\%) & $31.1\to11.8$ & $23.4\to6.1$ & $14.1\to2.8$ & $22.9\to6.9$ \\
 & $-19.4$ {\scriptsize [$-22.0$, $-16.7$]} & $-17.3$ {\scriptsize [$-19.8$, $-14.8$]} & $-11.2$ {\scriptsize [$-13.4$, $-9.1$]} & $-16.0$ {\scriptsize [$-17.5$, $-14.4$]} \\[3pt]
Attention cost $\tilde A$ & $6.5\to8.3$ & $10.6\to12.8$ & $13.3\to15.0$ & $10.1\to12.0$ \\
 & $+1.8$ {\scriptsize [$+1.5$, $+2.0$]} & $+2.2$ {\scriptsize [$+1.8$, $+2.7$]} & $+1.7$ {\scriptsize [$+1.1$, $+2.3$]} & $+1.9$ {\scriptsize [$+1.6$, $+2.2$]} \\[3pt]
Question value $\tilde V$ & $71.7\to71.2$ & $70.2\to68.8$ & $68.2\to67.7$ & $70.1\to69.2$ \\
 & $-0.6$ {\scriptsize [$-1.6$, $+0.5$]} & $-1.4$ {\scriptsize [$-2.4$, $-0.4$]} & $-0.5$ {\scriptsize [$-1.7$, $+0.6$]} & $-0.8$ {\scriptsize [$-1.4$, $-0.2$]} \\[3pt]
Utility $\tilde U$ & $56.4\to59.0$ & $55.6\to58.3$ & $54.4\to58.0$ & $55.5\to58.5$ \\
 & $+2.6$ {\scriptsize [$+1.2$, $+4.0$]} & $+2.7$ {\scriptsize [$+1.4$, $+4.0$]} & $+3.7$ {\scriptsize [$+2.2$, $+5.1$]} & $+3.0$ {\scriptsize [$+2.1$, $+3.9$]} \\[3pt]
Blocked reviews (\%) & $31\to10$ & $31\to10$ & $26\to9$ & $29\to10$ \\
Rewritten reviews (\%) & $10\to9$ & $13\to12$ & $14\to10$ & $12\to10$ \\
\addlinespace
Reward against an equal-cost budget & $-1.0\to0.5$ & $-3.3\to-1.1$ & $-5.7\to-0.5$ & $-3.3\to-0.4$ \\
 & $+1.5$ {\scriptsize [$-0.4$, $+3.5$]} & $+2.1$ {\scriptsize [$+0.3$, $+4.0$]} & $+5.1$ {\scriptsize [$+2.9$, $+7.2$]} & $+2.9$ {\scriptsize [$+1.7$, $+4.1$]} \\
All main-testbed domains, same comparison & $-0.3\to0.2$ & $-1.5\to-0.9$ & $-0.6\to0.7$ & $-0.8\to-0.0$ \\
 & $+0.5$ {\scriptsize [$+0.1$, $+0.9$]} & $+0.6$ {\scriptsize [$+0.1$, $+1.1$]} & $+1.2$ {\scriptsize [$+0.7$, $+1.7$]} & $+0.8$ {\scriptsize [$+0.5$, $+1.1$]} \\
\bottomrule
\end{tabular}
}
\ifdim\wd\rqreviewrevisiontestbox>\textwidth
\resizebox{\textwidth}{!}{\usebox{\rqreviewrevisiontestbox}}
\else\usebox{\rqreviewrevisiontestbox}\fi
\caption{\textbf{Post-hoc WebShop rule revision.} Original/revised on 160 shared test tasks, three GPT models, $N=1$--3, two trials. Requests per session; block/zero-question percentages; main reward/cost/value scales. Equal-cost reward uses \budgeted mixtures. Four-domain rows retain original $\tau$ results. }
\label{tab:review-revision-test}
\end{table*}

\subsection{Elicitation Recall on WebShop}
\label{app:elicitation-recall}
\paragraph{Definition and reading rule.}
Question value scores delivered questions and cannot see questions that
were not asked. WebShop goals withhold a fixed set of items at
construction: required attributes, option values, and a price bound
(7.1 per goal). Elicitation recall of a session is the share of the goal's
withheld items that the shopper discloses in any reply, asked for or
not, as identified by the simulator's literal matcher. Aggregation
follows the main tables (trials within task, tasks within model, equal
model weights), with 10,000 task-bootstrap draws. The reading rule was
fixed before recall was computed on the main runs: if the \mgragent's
recall is below that of \budgeted at matched question count, with an
interval excluding zero, its value gain on WebShop is read as more
precise questions at lower recall; if recall under the revised review
rule is above the original rule's, the revision recovers recall.

\paragraph{Recall by policy.}
Recall falls from \budgeted to \budgetedreview, \budgetedsummary, and \budgetedmgr at
every budget (Table~\ref{tab:webshop-recall}; eight models, three
trials, 160 test tasks; Unrestricted reaches 31.1\%). Over $N=1$--3,
\budgetedmgr is $-5.7$ points $[-6.2,-5.2]$ below \budgeted,
and lower for 7 of 8 models. Review without memory accounts for part of
this (\budgetedreview minus \budgeted: $-2.9$ $[-3.3,-2.5]$) and persistent memory
for a further step (\budgetedmgr minus \budgetedreview: $-2.8$ $[-3.2,-2.5]$).

\paragraph{Matched question count.}
To separate recall from asking less, we compare \budgetedmgr with the
\budgeted mixture of Appendix~\ref{app:review-controls}: within each
model, the two policies among \noask, \budgeted at $N=1$--3, and Unrestricted whose
question counts bracket that of \budgetedmgr, with recall interpolated
linearly. GPT-6-Astra at $N=3$ asks more than any of these policies and
is left out. The 23 supported model--budget cells give $-2.9$ $[-3.3,-2.4]$
($-2.8$ $[-3.3,-2.4]$ with the excluded cell), and the difference is clearly negative for
5 of 8 models. By the reading rule, the \mgragent's value gain on WebShop
reflects more precise questions at lower recall.

\paragraph{Revised review rule.}
Under the revised rule of Appendix~\ref{app:review-revision} (three GPT
models, two trials), recall rises from 27.9\% to 33.2\%
($+5.4$ $[+4.5,+6.2]$), close to these models' \budgeted level in the main runs
(34.6\%).

\paragraph{Matcher check and relation to reward.}
The matcher is literal and can miss paraphrased values. In 30 sessions
sampled by a fixed hash, six per policy family, it found 69 of the 73
items disclosed in the replies and reported none that was not
disclosed, so recall is slightly undercounted. Session recall correlates
with reward (Pearson $r=0.30$ $[0.24,0.35]$ over 49,725 sessions),
which indicates that the measure tracks information the task rewards.

\begin{table*}[t]
\centering
\footnotesize
\setlength{\tabcolsep}{5pt}
\begin{tabular}{@{}l cccc@{}}
\toprule
 & $N=1$ & $N=2$ & $N=3$ & $N=1$--3 \\
\midrule
\multicolumn{5}{@{}l}{\textit{Elicitation recall (\%), eight models}} \\
\budgeted & $25.8$ & $29.0$ & $30.1$ & $28.3$ \\
\budgetedreview & $23.3$ & $26.0$ & $27.0$ & $25.4$ \\
\budgetedsummary & $21.9$ & $24.4$ & $26.2$ & $24.2$ \\
\budgetedmgr & $20.0$ & $23.0$ & $24.7$ & $22.6$ \\
\midrule
\multicolumn{5}{@{}l}{\textit{Differences in recall (points)}} \\
\budgetedmgr $-$ \budgeted & $-5.8$ {\scriptsize[$-6.5$, $-5.2$]} & $-5.9$ {\scriptsize[$-6.6$, $-5.3$]} & $-5.4$ {\scriptsize[$-6.0$, $-4.7$]} & $-5.7$ {\scriptsize[$-6.2$, $-5.2$]} \\
\quad at matched question count & $-1.1$ {\scriptsize[$-1.5$, $-0.6$]} & $-3.1$ {\scriptsize[$-3.7$, $-2.5$]} & $-4.6$ {\scriptsize[$-5.6$, $-3.6$]} & $-2.9$ {\scriptsize[$-3.3$, $-2.4$]} \\
\budgetedreview $-$ \budgeted & $-2.5$ {\scriptsize[$-3.1$, $-1.9$]} & $-2.9$ {\scriptsize[$-3.5$, $-2.4$]} & $-3.1$ {\scriptsize[$-3.7$, $-2.5$]} & $-2.9$ {\scriptsize[$-3.3$, $-2.5$]} \\
\budgetedmgr $-$ \budgetedreview & $-3.3$ {\scriptsize[$-3.9$, $-2.7$]} & $-3.0$ {\scriptsize[$-3.6$, $-2.3$]} & $-2.3$ {\scriptsize[$-2.9$, $-1.7$]} & $-2.8$ {\scriptsize[$-3.2$, $-2.5$]} \\
Revised $-$ original rule (three GPT models) & $+5.0$ {\scriptsize[$+3.7$, $+6.2$]} & $+5.8$ {\scriptsize[$+4.4$, $+7.1$]} & $+5.4$ {\scriptsize[$+4.0$, $+6.7$]} & $+5.4$ {\scriptsize[$+4.5$, $+6.2$]} \\
\bottomrule
\end{tabular}
\caption{\textbf{Elicitation recall on WebShop.} Share (\%) of a goal's withheld items that the shopper discloses in any reply, by the simulator's literal matcher. Eight models, three trials, 160 test tasks; Unrestricted reaches 31.1\%. The matched comparison covers the 23 of 24 model--budget cells inside the range of the \budgeted mixture (seven models at $N=3$). The rule comparison uses the review-revision runs (two trials). Brackets: 95\% task-bootstrap intervals.}
\label{tab:webshop-recall}
\end{table*}

\section{Human Burden Study}
\label{app:burden-study}
\paragraph{Design.}
We pair complete user-facing trial-0 conversations for the same task and
backbone (GPT-6-Astra or GPT-5.4; four domains). Sixty different-count
pairs compare \budgeted at $N=1$ with Unrestricted; sixty equal-count
pairs compare \budgetedmgr with \budgeted at the same $N=1$--3.
Equal-count pairs require positive counts, a cost gap of at least 0.5
units, and a user-facing length ratio of at most 1.5. A fixed hash orders
eligible pairs across model/domain cells, without selecting on reward or
human judgments. Results are conditional on these restrictions, including
the two backbones.

Two professional annotators from an in-house pool of six, fluent English
speakers with extensive prior NLP annotation experience and different from those who validated the judge and the classifier, completed two pilot rounds before independently rating all 120 main pairs; pilot judgments are excluded.
For each pair they choose A, B, or same, first for overall tiringness and
then for question burden, considering frequency, readability, and
justification. Because the question-burden instruction names the cost's
components, H1 and H2 partly measure agreement with that instruction; the
overall judgment names none of them. Model and condition names, cost
scores, and explicit outcomes are hidden, and the higher-cost side
appears first in half the pairs. H1 tests question burden at different
counts and H2 at equal counts; H3 tests overall impressions, with
equal-count pairs the pre-specified primary test. These tests concern
rankings, not cost weights.

\paragraph{Analysis and findings.}
Alignment averages the two annotators' shares of non-tie answers choosing
the higher-cost side. We use 10,000 paired bootstrap draws, resampling
conversation pairs with both annotators' labels together; the
pre-specified criterion is a 95\% interval above 50\%. Agreement and
Cohen's $\kappa$ include ties; alignment rates appear in
Table~\ref{tab:human-alignment-summary-preview}.
Question-burden alignment is 90.8\% [84.9, 95.8] at different counts and
71.7\% [59.1, 82.9] at equal counts; count gives the same ranking in the
former. Overall-burden alignment is 62.5\% [52.5, 72.5] and 75.0\%
[64.5, 84.9], respectively. Question agreement is 99/120
($\kappa=0.70$), overall agreement 95/120 ($\kappa=0.60$). On equal-count
pairs, the longer conversation is chosen in 49.5\% of question-burden and
37.5\% of overall answers; because pairs were restricted to a length
ratio of at most 1.5, this comparison has limited range, and the main
text instead reports the word count of the agent's questions. These are
transcript impressions, not fatigue or time during live interaction. This
study does not separately examine the classifier's exemptions, which
Appendix~\ref{app:classifier-validation} validates.

\paragraph{Component analysis.}
To identify which terms carry the equal-count alignment, we scored each
equal-count pair by one component at a time, using the same final labels,
alignment measure, and bootstrap, and a reading rule fixed before
alignment was computed: a score is supported if its 95\% interval lies
above 50\%. Because counts are equal, count plus effort reduces to effort
and count plus value to value. Value alone is supported (71.1\% question
burden, 73.6\% overall) and matches the full cost (full minus value
$+0.6$ [$-1.4$, $+2.9$] for question burden and $+1.4$ [$0.0$, $+3.4$]
overall); effort alone (62.7\%, 57.3\%) and the word count of the
agent's discretionary questions (49.6\%, 56.7\%) are not
(Table~\ref{tab:human-alignment-summary-preview}). The full cost exceeds
effort alone on overall burden by 17.7 points [$+5.2$, $+30.9$]. Pairs
were selected for a full-cost gap of at least 0.5 units, which favors
the full cost and whichever component drives its gaps. Within-pair effort
gaps are small (median 0.10 cost units versus 0.70 for value), so the
study has limited power to detect a contribution from reading effort.


\section{Complete Results and Task Sensitivity}
\label{app:complete-tables}
\label{app:agent-preview-reference-displays}
\subsection{Domain Tradeoffs and Interactive Research}
\label{app:budget-across-tasks}
\label{sec:results-browsecomp}
Asking-gain retention is $(R_{\mathrm{Budgeted}}-R_{\text{No-ask}})/
(R_{\mathrm{Unrestricted}}-R_{\text{No-ask}})$, not reward retention
relative to Unrestricted. WebShop retains 93\% of asking gain at $N=1$,
versus 20--35\% on the other domains; $N=3$ retains 65--76\% there.
Compound requests can obtain several shopping requirements. Negative
Fieller bounds reflect uncertain improvement over \noask, not negative
reward (Airline $N=1$: $[-23.2,45.1]$\%).
Table~\ref{tab:domain-tradeoff-full} reports current eight-model,
three-trial estimates.

Research uses a different interface, native charged ASK, larger budgets,
and no separate human validation of the judge
(Appendix~\ref{app:datasets-interactive-research}). The current five-model,
three-trial comparison is Table~\ref{tab:external-domain-preview}.
Averaging $N=1$--8, \budgeted saves 15.6\% cost for 3.1 reward points;
\budgetedmgr saves a further 5.4\%, adds 1.8 value points, and loses
0.9 reward points. Neither improves utility at $\lambda=0.03$ over
Unrestricted. Older four-model, single-trial displays are omitted.
\begin{figure*}[t]
\centering
\includegraphics[width=\textwidth]{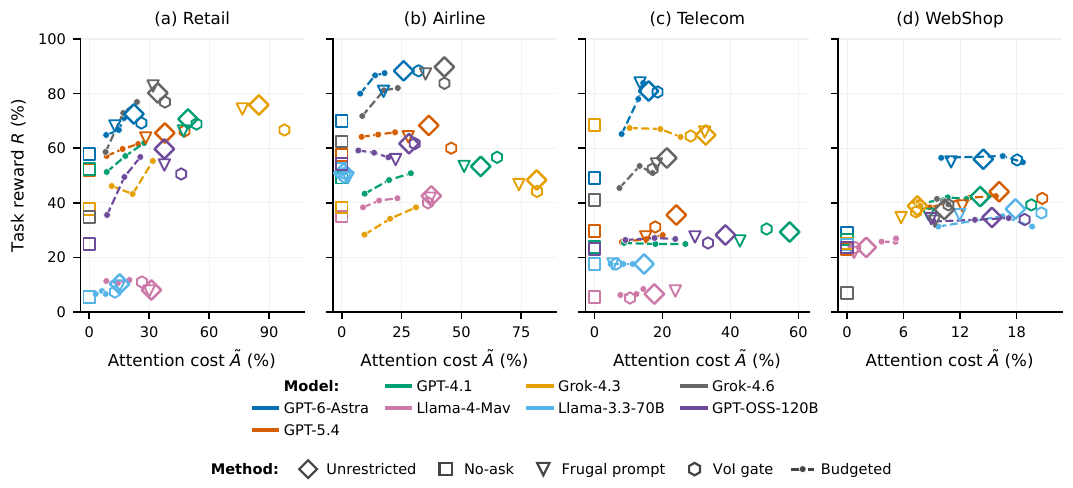}
\caption{\textbf{Reward--user-attention tradeoffs by domain.} Eight models,
three trials. Curves connect \budgeted $N=1$--3; markers show \noask,
Unrestricted, \frugal, and \voigate. Upper-left is better.}
\label{fig:rq1-reward-attention-all}
\end{figure*}
\begin{table*}[t]
\centering
\footnotesize
\setlength{\tabcolsep}{3pt}
\newsavebox{\rqdomainfullbox}
\sbox{\rqdomainfullbox}{%
\begin{tabular}{@{}lrr rrr rrr rrr@{}}
\toprule
 & & & \multicolumn{3}{c}{$\Delta R$} & \multicolumn{3}{c}{Saved (\%)} & \multicolumn{3}{c}{Kept (\%)} \\
\cmidrule(lr){4-6}\cmidrule(lr){7-9}\cmidrule(lr){10-12}
Domain & $Q$ & Gain & $N{=}1$ & $N{=}2$ & $N{=}3$ & $N{=}1$ & $N{=}2$ & $N{=}3$ & $N{=}1$ & $N{=}2$ & $N{=}3$ \\
\midrule
Retail & $3.76$ & $+21.5$ & $-13.9$ & $-9.4$ & $-5.1$ & $79$ & $59$ & $43$ & $35$ & $56$ & $76$ \\
 &  & {\tiny[$+17.9$, $+25.3$]} & {\tiny[$-17.2$, $-10.8$]} & {\tiny[$-12.1$, $-6.9$]} & {\tiny[$-7.5$, $-2.7$]} & {\tiny[$77$, $80$]} & {\tiny[$57$, $62$]} & {\tiny[$39$, $46$]} & {\tiny[$26$, $45$]} & {\tiny[$47$, $66$]} & {\tiny[$67$, $86$]} \\[1pt]
Airline & $3.71$ & $+10.5$ & $-8.4$ & $-4.7$ & $-3.7$ & $81$ & $63$ & $46$ & $20$ & $55$ & $65$ \\
 &  & {\tiny[$+5.3$, $+15.9$]} & {\tiny[$-12.4$, $-4.7$]} & {\tiny[$-7.4$, $-2.2$]} & {\tiny[$-6.6$, $-0.9$]} & {\tiny[$78$, $83$]} & {\tiny[$58$, $66$]} & {\tiny[$41$, $51$]} & {\tiny[$-23$, $45$]} & {\tiny[$20$, $76$]} & {\tiny[$32$, $90$]} \\[1pt]
Telecom & $2.60$ & $+7.6$ & $-5.0$ & $-2.2$ & $-1.8$ & $71$ & $47$ & $31$ & $34$ & $71$ & $76$ \\
 &  & {\tiny[$+5.7$, $+9.7$]} & {\tiny[$-6.7$, $-3.4$]} & {\tiny[$-3.8$, $-0.8$]} & {\tiny[$-3.4$, $-0.2$]} & {\tiny[$70$, $73$]} & {\tiny[$45$, $49$]} & {\tiny[$28$, $34$]} & {\tiny[$16$, $50$]} & {\tiny[$53$, $89$]} & {\tiny[$56$, $97$]} \\[1pt]
WebShop & $1.28$ & $+16.4$ & $-1.1$ & $+0.2$ & $-0.6$ & $33$ & $5$ & $-10$ & $93$ & $101$ & $96$ \\
 &  & {\tiny[$+14.1$, $+18.7$]} & {\tiny[$-2.2$, $-0.0$]} & {\tiny[$-0.8$, $+1.2$]} & {\tiny[$-1.7$, $+0.4$]} & {\tiny[$31$, $35$]} & {\tiny[$2$, $7$]} & {\tiny[$-14$, $-7$]} & {\tiny[$87$, $100$]} & {\tiny[$95$, $108$]} & {\tiny[$90$, $103$]} \\[1pt]
\midrule
Mean & $2.84$ & $+14.0$ & $-7.1$ & $-4.0$ & $-2.8$ & $73$ & $52$ & $36$ & $49$ & $71$ & $80$ \\
 &  & {\tiny[$+12.3$, $+15.8$]} & {\tiny[$-8.5$, $-5.8$]} & {\tiny[$-5.1$, $-3.0$]} & {\tiny[$-3.9$, $-1.8$]} & {\tiny[$72$, $74$]} & {\tiny[$50$, $54$]} & {\tiny[$33$, $38$]} & {\tiny[$43$, $56$]} & {\tiny[$65$, $77$]} & {\tiny[$73$, $87$]} \\[1pt]
\bottomrule
\end{tabular}
}
\ifdim\wd\rqdomainfullbox>\textwidth
\resizebox{\textwidth}{!}{\usebox{\rqdomainfullbox}}
\else\usebox{\rqdomainfullbox}\fi
\caption{\textbf{Budget tradeoffs by domain.} Eight models, three trials. Counts per session; attention savings relative to Unrestricted; reward differences in points. Retention measures gain over \noask, with approximate Fieller intervals. Other brackets are paired 95\% task-bootstrap intervals. Negative retention bounds reflect uncertainty in asking gains.}
\label{tab:domain-tradeoff-full}
\label{tab:domain-tradeoff}
\end{table*}

\begin{table}[t]
\centering
\footnotesize
\setlength{\tabcolsep}{1.25pt}
\renewcommand{\arraystretch}{1.12}
\begin{tabular}{@{}l rrrr|rrrr@{}}
\toprule
Method & \multicolumn{4}{c|}{Main} & \multicolumn{4}{c}{BrowseComp} \\
\cmidrule(lr){2-5}\cmidrule(lr){6-9}
 & $R\uparrow$ & $\tilde A\downarrow$ & $\tilde U\uparrow$ & $\tilde V\uparrow$ & $R\uparrow$ & $\tilde A\downarrow$ & $\tilde U\uparrow$ & $\tilde V\uparrow$ \\
\midrule
Unrestricted & \textbf{50.8} & 33.3 & 55.7 & 63.2 & \textbf{32.7} & 19.0 & 47.7 & 64.3 \\
\frugal & 48.2 & 28.4 & 55.5 & 65.9 & 31.7 & 15.8 & \textbf{48.0} & 67.0 \\
\voigate & 49.0 & 35.2 & 53.9 & 58.5 & 32.3 & 18.4 & 47.6 & 67.3 \\
\budgeted & 46.0 & 14.6 & \textbf{58.2} & 68.4 & 29.6 & 16.0 & 46.5 & 68.0 \\
\budgetedmgr & 44.5 & \textbf{12.9} & 57.7 & \textbf{70.8} & 28.7 & \textbf{15.1} & 46.1 & \textbf{69.8} \\
\bottomrule
\end{tabular}
\caption{\textbf{Main domains and Ambiguous BrowseComp-Plus.} Five matched models, three trials each. Models: Section~\ref{sec:experimental-models}. Main is the equal four-domain mean. Budgeted rows average $N=1$--3 on Main and $N=1$--8 on BrowseComp-Plus, allowing additional research clarifications about hidden identifying constraints. Budgets and models have equal weights. The ranges differ; Table~\ref{tab:main-domain-summary-full} uses eight models. Scales and value aggregation follow that table ($\lambda=0.03$). Bold marks the best displayed value.}
\label{tab:external-domain-preview}
\end{table}

\subsection{Attention Valuations}
\label{app:reward-per-attention}
\label{app:utility-plots}
\label{app:valuations}
\begin{figure}[t]
    \centering
    \includegraphics[width=\columnwidth]{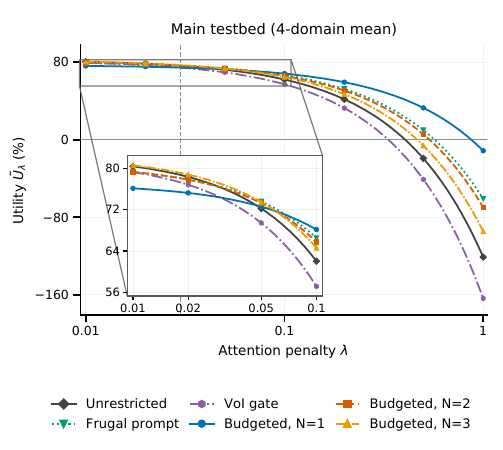}
    \caption{\textbf{Attention valuation changes the preferred budget.}
    GPT-6-Astra utility, averaged equally over four domains and three
    trials, on the utility scale of Appendix~\ref{app:evaluation-metrics}.
    The logarithmic axis varies $\lambda$; the dashed line marks 0.03
    and the inset enlarges $\lambda\leq0.1$.
    \voigate denotes the original (as-run) gate (Appendix~\ref{app:baselines}).}
    \label{fig:main-utility-lambda}
\end{figure}

We compute utility without rerunning policies at
$\lambda\in\{0.01,0.02,0.05,0.1,0.2,0.5,1\}$ and 0.03. Unrestricted asking's pooled yield is 3.2 reward points per cost unit (ratio of model-averaged means; 3.5 averaging per-model yields), with a model maximum of 7.8 (GPT-6-Astra); ten points per
unit therefore favors tight budgets in the evaluated comparisons.
At three points, asking roughly pays for itself on average.
Reward, cost, break-even valuations, and equal-cost comparisons do not
depend on this summary choice. Domain yields are 1.8 on Airline/Telecom,
3.7 on Retail, and 9.0 on WebShop. \noask is the evaluated $N=0$
policy, not an optimized no-question policy; yields are relative to
this reference, not a bound against all possible no-question policies.
Table~\ref{tab:best-budget} and Figure~\ref{fig:main-utility-lambda}
show valuation sensitivity.

\begin{figure}[t]
    \centering
    \includegraphics[width=\columnwidth]{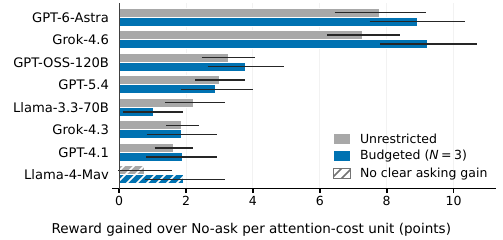}
    \caption{\textbf{A three-question budget buys more reward per unit of attention overall, significantly so for three of eight models.} Reward gained over \noask per unit of user-attention cost, averaged over four domains, with 95\% task-bootstrap intervals. \budgeted ($N=3$) is more cost-effective for three models, less for Llama-3.3-70B, and statistically tied for four. Hatched: unrestricted asking has no clear gain over \noask for this model, so its comparison says little. Numbers: Table~\ref{tab:yield-budget}.}
    \label{fig:yield}
\end{figure}

\paragraph{Yield under a budget.}
A policy's yield is $Y_p=(R_p-R_{\text{No-ask}})/A_p$, reward points gained over \noask per raw cost unit: a ratio of domain-averaged means per model and of model-averaged means when pooled. \noask is excluded (no gain, no cost); total reward in the numerator would favor asking less for its own sake. Table~\ref{tab:yield-budget} and Figure~\ref{fig:yield} compare \budgeted at $N=3$ with Unrestricted on the paired task bootstrap of Table~\ref{tab:paired-effects-full}, resampling \noask jointly. The rule was fixed before computation: more cost-effective if the 95\% interval of $Y_{N=3}-Y_{\mathrm{Unr}}$ lies above zero, less if below, otherwise no detectable difference. Pooled yield rises from 3.2 to 3.9 ($+0.8$ [$+0.4$, $+1.1$]). The equal-weight mean of per-model differences is smaller, $+0.47$ [$+0.14$, $+0.77$] ($+0.57$ [$+0.07$, $+1.04$] counting requests). The pooled ratio weights models by cost: under Unrestricted, Grok-4.3 and GPT-4.1 carry the most cost at low yield, so the pooled yield (3.17) sits below the mean of per-model yields (3.48), while at $N=3$ the two agree (3.94). About 0.3 of the pooled gain thus reflects reweighting across models rather than gains within them. The budget is more cost-effective for GPT-6-Astra, Grok-4.6, and Llama-4-Mav, less for Llama-3.3-70B, and not detectably different for four models. Llama-4-Mav's unrestricted asking gain is not distinguishable from zero ($+2.5$ [$-0.1$, $+5.2$]), so its comparison says little. Llama-3.3-70B asks least; the budget removes most of its small asking gain (4.1 to 1.5 points) but little of its cost (1.82 to 1.49 units). The extra questions Unrestricted asks beyond the budget earn 1.8 points per unit [1.2, 2.4], the $N=3$ break-even valuation and about half the average yield. With cost as the count of delivered requests, which uses no judged score, every model keeps its classification (pooled $+1.0$ [$+0.5$, $+1.4$]).

The gain varies by task and policy (Table~\ref{tab:yield-policies}). It is clear on Retail ($+1.2$), not detectable on Airline or Telecom, and reversed on WebShop ($-1.1$ [$-1.8$, $-0.5$]), where the budget costs 10\% more than Unrestricted (more for six of eight models) and gains slightly less. Since the cap itself can only reduce asking, the increase comes from stating the budget in the prompt (Appendix~\ref{app:agent-preview-behavior}). Only Llama-4-Mav gains yield there, through reward rather than cost. Tighter budgets show larger pooled gains ($+2.6$ at $N=1$, $+1.5$ at $N=2$) at larger reward losses, so $N=3$ is the conservative comparison. \budgetedmgr at $N=3$ yields 4.3, consistent with its lower cost at similar reward; at equal cost, its reward does not differ from \budgeted (Section~\ref{sec:results-rq3}). \frugal leaves yield unchanged; the as-run \voigate lowers it ($-0.6$ [$-0.8$, $-0.3$]). This analysis was added after the main results were known and is outside the family of Table~\ref{tab:headline-family}.

\begin{table*}[t]
\centering
\footnotesize
\setlength{\tabcolsep}{3.5pt}
\newsavebox{\rqyieldbudgetbox}
\sbox{\rqyieldbudgetbox}{%
\begin{tabular}{@{}l rrr rr rr rr rr@{}}
\toprule
 & \multicolumn{3}{c}{Reward} & \multicolumn{2}{c}{Cost} & \multicolumn{2}{c}{Asking gain} & \multicolumn{2}{c}{Yield} & \multicolumn{2}{c}{Yield difference} \\
\cmidrule(lr){2-4}\cmidrule(lr){5-6}\cmidrule(lr){7-8}\cmidrule(lr){9-10}\cmidrule(l){11-12}
Model & \noask & Unr. & Budget & Unr. & Budget & Unr. & Budget & Unr. & Budget & Full cost & Count only \\
\midrule
GPT-6-Astra & $51.5$ & $74.4$ & $74.3$ & $2.95$ & $2.56$ & $+23.0$ & $+22.9$ & $7.8$ & $8.9$ & $+1.1^{*}$ {\tiny[$+0.1$, $+2.1$]} & $+1.6^{*}$ {\tiny[$+0.1$, $+2.9$]} \\
Grok-4.6 & $36.3$ & $66.0$ & $62.3$ & $4.08$ & $2.82$ & $+29.7$ & $+26.0$ & $7.3$ & $9.2$ & $+2.0^{*}$ {\tiny[$+1.1$, $+2.8$]} & $+2.5^{*}$ {\tiny[$+1.3$, $+3.7$]} \\
GPT-OSS-120B & $31.4$ & $46.1$ & $43.6$ & $4.49$ & $3.22$ & $+14.6$ & $+12.2$ & $3.3$ & $3.8$ & $+0.5\phantom{^{*}}$ {\tiny[$-0.3$, $+1.3$]} & $+0.6\phantom{^{*}}$ {\tiny[$-0.6$, $+1.8$]} \\
GPT-5.4 & $40.5$ & $53.4$ & $49.5$ & $4.29$ & $3.10$ & $+12.8$ & $+9.0$ & $3.0$ & $2.9$ & $-0.1\phantom{^{*}}$ {\tiny[$-0.9$, $+0.7$]} & $-0.2\phantom{^{*}}$ {\tiny[$-1.4$, $+1.0$]} \\
Llama-3.3-70B & $25.0$ & $29.1$ & $26.6$ & $1.82$ & $1.49$ & $+4.1$ & $+1.5$ & $2.2$ & $1.0$ & $-1.2^{*}$ {\tiny[$-1.9$, $-0.5$]} & $-1.9^{*}$ {\tiny[$-3.1$, $-0.9$]} \\
Grok-4.3 & $42.4$ & $57.0$ & $49.2$ & $7.75$ & $3.60$ & $+14.5$ & $+6.7$ & $1.9$ & $1.9$ & $0.0\phantom{^{*}}$ {\tiny[$-0.9$, $+0.9$]} & $-0.1\phantom{^{*}}$ {\tiny[$-1.6$, $+1.3$]} \\
GPT-4.1 & $38.0$ & $48.9$ & $44.8$ & $6.72$ & $3.59$ & $+10.9$ & $+6.8$ & $1.6$ & $1.9$ & $+0.3\phantom{^{*}}$ {\tiny[$-0.6$, $+1.0$]} & $+0.3\phantom{^{*}}$ {\tiny[$-0.9$, $+1.4$]} \\
Llama-4-Mav$^{\dagger}$ & $17.7$ & $20.2$ & $22.2$ & $3.31$ & $2.36$ & $+2.5$ & $+4.5$ & $0.8$ & $1.9$ & $+1.2^{*}$ {\tiny[$+0.1$, $+2.3$]} & $+1.9^{*}$ {\tiny[$+0.1$, $+3.7$]} \\
\midrule
Pooled (ratio of means) & $35.4$ & $49.4$ & $46.6$ & $4.42$ & $2.84$ & $+14.0$ & $+11.2$ & $3.2$ & $3.9$ & $+0.8^{*}$ {\tiny[$+0.4$, $+1.1$]} & $+1.0^{*}$ {\tiny[$+0.5$, $+1.4$]} \\
Mean of per-model yields & & & & & & & & $3.5$ & $3.9$ & $+0.47^{*}$ {\tiny[$+0.14$, $+0.77$]} & $+0.57^{*}$ {\tiny[$+0.07$, $+1.04$]} \\
\bottomrule
\end{tabular}
}
\ifdim\wd\rqyieldbudgetbox>\textwidth
\resizebox{\textwidth}{!}{\usebox{\rqyieldbudgetbox}}
\else\usebox{\rqyieldbudgetbox}\fi
\caption{\textbf{A budget of three questions raises yield, pooled and for the average model; it is more cost-effective for three of eight models, less for one, with no detectable difference for four.} Unrestricted (Unr.) versus \budgeted ($N{=}3$) (Budget); four-domain, three-trial means, models ordered by the Unrestricted yield. Asking gain: reward minus \noask reward, in points. Yield: asking gain divided by user-attention cost in raw cost units, a ratio of domain-averaged means; the pooled row is a ratio of model-averaged means, and the last row averages per-model yields. Yield difference: Budget minus Unrestricted, on the full cost and on the count of delivered requests, which uses no judged score. Brackets: paired 95\% task-bootstrap intervals (10,000 draws; \noask, Unrestricted and the budget resampled together). $^{*}$: the interval excludes zero, the rule fixed before computation. $^{\dagger}$: the interval of the Unrestricted asking gain includes zero, so the comparison says little. This analysis was added after the main results were known.}
\label{tab:yield-budget}
\end{table*}

\begin{table*}[t]
\centering
\footnotesize
\setlength{\tabcolsep}{5pt}
\newsavebox{\rqyieldpoliciesbox}
\sbox{\rqyieldpoliciesbox}{%
\begin{tabular}{@{}l rr rr rr c@{}}
\toprule
 & & & \multicolumn{2}{c}{Yield} & \multicolumn{2}{c}{Yield difference} & Models \\
\cmidrule(lr){4-5}\cmidrule(lr){6-7}
 & Saved (\%) & $\Delta R$ & Unr. & Policy & Full cost & Count only & more / tied / less \\
\midrule
\multicolumn{8}{@{}l}{\textit{By policy, four-domain mean}} \\
\budgeted ($N{=}1$) & $73$ & $-7.1$ & $3.2$ & $5.8$ & $+2.6^{*}$ {\tiny[$+1.8$, $+3.5$]} & $+3.3^{*}$ {\tiny[$+2.1$, $+4.5$]} & 3 / 5 / 0 \\
\budgeted ($N{=}2$) & $52$ & $-4.0$ & $3.2$ & $4.7$ & $+1.5^{*}$ {\tiny[$+1.1$, $+2.0$]} & $+2.0^{*}$ {\tiny[$+1.3$, $+2.6$]} & 3 / 5 / 0 \\
\budgeted ($N{=}3$) & $36$ & $-2.8$ & $3.2$ & $3.9$ & $+0.8^{*}$ {\tiny[$+0.4$, $+1.1$]} & $+1.0^{*}$ {\tiny[$+0.5$, $+1.4$]} & 3 / 4 / 1 \\
\budgetedmgr ($N{=}1$) & $76$ & $-7.8$ & $3.2$ & $5.9$ & $+2.8^{*}$ {\tiny[$+1.8$, $+3.7$]} & $+3.4^{*}$ {\tiny[$+2.1$, $+4.7$]} & 3 / 5 / 0 \\
\budgetedmgr ($N{=}2$) & $58$ & $-5.3$ & $3.2$ & $4.7$ & $+1.5^{*}$ {\tiny[$+1.0$, $+2.0$]} & $+1.8^{*}$ {\tiny[$+1.0$, $+2.5$]} & 1 / 7 / 0 \\
\budgetedmgr ($N{=}3$) & $44$ & $-3.4$ & $3.2$ & $4.3$ & $+1.1^{*}$ {\tiny[$+0.7$, $+1.5$]} & $+1.4^{*}$ {\tiny[$+0.8$, $+1.9$]} & 1 / 7 / 0 \\
\frugal & $16$ & $-2.4$ & $3.2$ & $3.1$ & $0.0\phantom{^{*}}$ {\tiny[$-0.3$, $+0.2$]} & $-0.1\phantom{^{*}}$ {\tiny[$-0.5$, $+0.2$]} & 1 / 7 / 0 \\
\voigate & $-4$ & $-2.0$ & $3.2$ & $2.6$ & $-0.6^{*}$ {\tiny[$-0.8$, $-0.3$]} & $-0.8^{*}$ {\tiny[$-1.2$, $-0.4$]} & 0 / 4 / 4 \\
\midrule
\multicolumn{8}{@{}l}{\textit{By domain, \budgeted ($N{=}3$)}} \\
Retail & $43$ & $-5.1$ & $3.7$ & $4.9$ & $+1.2^{*}$ {\tiny[$+0.7$, $+1.7$]} & $+1.6^{*}$ {\tiny[$+0.8$, $+2.4$]} & 3 / 5 / 0 \\
Airline & $46$ & $-3.7$ & $1.8$ & $2.2$ & $+0.4\phantom{^{*}}$ {\tiny[$-0.5$, $+1.3$]} & $+0.4\phantom{^{*}}$ {\tiny[$-0.9$, $+1.7$]} & 0 / 7 / 0$^{\ddagger}$ \\
Telecom & $31$ & $-1.8$ & $1.8$ & $2.0$ & $+0.2\phantom{^{*}}$ {\tiny[$-0.3$, $+0.7$]} & $+0.3\phantom{^{*}}$ {\tiny[$-0.6$, $+1.1$]} & 1 / 6 / 1 \\
WebShop & $-10$ & $-0.6$ & $9.0$ & $7.8$ & $-1.1^{*}$ {\tiny[$-1.8$, $-0.5$]} & $-1.8^{*}$ {\tiny[$-2.7$, $-0.9$]} & 1 / 5 / 2 \\
\bottomrule
\end{tabular}
}
\ifdim\wd\rqyieldpoliciesbox>\textwidth
\resizebox{\textwidth}{!}{\usebox{\rqyieldpoliciesbox}}
\else\usebox{\rqyieldpoliciesbox}\fi
\caption{\textbf{Yield against Unrestricted, by policy and by domain.} Pooled eight-model yields (ratio of means), three trials. Saved: user-attention cost removed relative to Unrestricted; a negative value means the policy costs more. $\Delta R$: reward change in points. Yield and its differences as in Table~\ref{tab:yield-budget}. Models: number of models for which the policy is more cost-effective than Unrestricted, not detectably different, and less cost-effective, by the same rule on the full cost. $^{\ddagger}$: one model has no interval because it asks too rarely in that domain.}
\label{tab:yield-policies}
\end{table*}

\begin{table}[t]
\centering
\footnotesize
\setlength{\tabcolsep}{6pt}
\newsavebox{\rqbestbudgetbox}
\sbox{\rqbestbudgetbox}{%
\begin{tabular}{@{}lccccc@{}}
\toprule
Model & $\lambda=0.01$ & $\lambda=0.02$ & $\lambda=0.03$ & $\lambda=0.05$ & $\lambda=0.1$ \\
\midrule
Grok-4.3 & $\infty$ & 1 & 1 & 1 & 1 \\
GPT-4.1 & $\infty$ & 2 & 1 & 1 & 1 \\
GPT-OSS-120B & $\infty$ & 2 & 2 & 1 & 1 \\
GPT-5.4 & $\infty$ & $\infty$ & 1 & 1 & 1 \\
Grok-4.6 & $\infty$ & $\infty$ & 2 & 2 & 1 \\
GPT-6-Astra & 3 & 3 & 3 & 3 & 1 \\
Llama-4-Mav & 3 & 1 & 1 & 1 & 1 \\
Llama-3.3-70B & $\infty$ & $\infty$ & 1 & 1 & 1 \\
\midrule
Common budget & $\infty$ & 2 & 2 & 1 & 1 \\
Gain of per-model budgets & $0.4$ & $0.7$ & $0.9$ & $0.6$ & $0.0$ \\
\bottomrule
\end{tabular}
}
\ifdim\wd\rqbestbudgetbox>\columnwidth
\resizebox{\columnwidth}{!}{\usebox{\rqbestbudgetbox}}
\else\usebox{\rqbestbudgetbox}\fi
\caption{\textbf{Highest-utility \budgeted setting.} \budgeted $N=1$--3 and Unrestricted ($\infty$), eight models, four-domain three-trial means; utility $R-\lambda A_{\mathrm{full}}$. Common budget maximizes mean utility; Gain compares model-specific versus common budgets in reward points. Choices use point estimates, not optimized user-specific policies.}
\label{tab:best-budget}
\end{table}

\subsection{Model and Domain Asking Behavior}
\label{app:agent-preview-behavior}
Unrestricted frequency and reward return differ sharply: Llama-4 asks
3.44 Airline versus 0.22 WebShop messages; Llama-3.3 asks 0.075 versus
1.86. On Telecom, GPT-4.1 asks 5.66 messages for 5.5 reward points over
\noask, versus Astra's 1.62 for 31.9. Zero charged messages need not mean
zero interaction: Astra Telecom also records about 11.74 exempt operational
messages. Frequency alone establishes neither over- nor under-asking.

Marginal comparisons support model/task-specific budgets. Astra Telecom gains
12.8 points ($[6.6,19.4]$) from $N=1$ to 2 and 5.9 ($[1.8,10.3]$)
from 2 to 3. Grok-4.6 WebShop gains 34.4 ($[29.8,39.2]$) over \noask
at $N=1$, with unclear subsequent gains. Llama-4 WebShop at $N=3$
asks 0.361 more messages ($[0.229,0.492]$) and gains 3.29 points
($[0.95,5.69]$) over Unrestricted; this changes prompting and enforcement
together. Diagnostics are retrospective task averages, not causal effects
conditioned on whether the model asked.
\begin{figure}[t]
    \centering
    \includegraphics[width=\columnwidth]{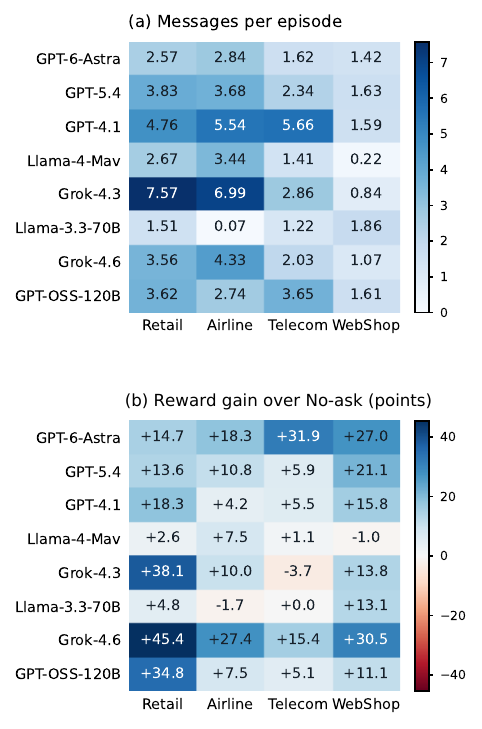}
    \caption{\textbf{Asking frequency and measured reward return vary
    across models and domains.} Unrestricted means over three trials.
    Top: discretionary messages per session. Bottom: reward gain over
    the evaluated \noask baseline, in points. \noask retains required
    exchanges and is not an optimized no-question policy. Message
    frequency alone does not diagnose excessive or insufficient
    interaction.}
    \label{fig:model-domain-asking-preview}
\end{figure}

\subsection{Simulator Disclosure Sensitivity}
\label{app:simulator-disclosure}
Astra/GPT-4.1 compare Guarded and Original on 20 fixed-hash test tasks
per $\tau$ domain, two trials, $N=1,2$ (2,400 sessions, following 120
development checks). Regimes are interleaved; scenario facts, tools,
reward, GPT-5.5 simulator, classifier, and agent settings are fixed.
Telecom additionally changes user-tool guidelines. Both regimes prescribe
progressive disclosure. This tests prompt sensitivity, not realism.

Table~\ref{tab:simulator-disclosure} uses the protocol's raw grid-mean
utility, with mean valuation 0.2686 rather than 0.03. Original reduces
unrestricted cost by 46.6\% for Astra and 24.2\% for GPT-4.1.
Budgets still help: GPT-4.1 $N=2$ saves 49.8\% at unchanged mean reward;
Astra $N=1$ saves 38.8\% for 2.5 reward points. \mgr benefits are less
stable. Astra's Guarded $N=1$ utility gain of $+0.153$ becomes $-0.017$
under Original; interaction intervals exclude zero on Retail
($-0.314$ [$-0.569,-0.057$]) and Airline ($-0.178$ [$-0.277,-0.084$]).
GPT-4.1's Original \mgr gain is uncertain by domain. Thus disclosure
conditions the added benefit of \mgr; higher judged value does not guarantee
higher utility. Paired intervals use 10,000 task draws, retaining trials;
the 20-task panels limit precision.

\begin{table*}[t]
    \centering
    \footnotesize
    \setlength{\tabcolsep}{6pt}
    \begin{tabular}{@{}lrrrr|rrrr@{}}
        \toprule
        & \multicolumn{4}{c|}{Guarded} & \multicolumn{4}{c}{Original} \\
        \cmidrule(lr){2-5}\cmidrule(l){6-9}
        Policy & $R\uparrow$ & $A\downarrow$ & $\bar U\uparrow$ & $V\uparrow$
               & $R\uparrow$ & $A\downarrow$ & $\bar U\uparrow$ & $V\uparrow$ \\
        \midrule
        \multicolumn{9}{l}{\textbf{GPT-6-Astra}} \\
        Unrestricted & 0.783 & 3.135 & -0.059 & 8.374 & 0.883 & 1.676 & 0.433 & 8.416 \\
        \budgeted ($N=1$) & 0.583 & 1.240 & 0.250 & 8.114 & 0.858 & 1.026 & 0.583 & 8.362 \\
        \budgetedmgr ($N=1$) & 0.700 & 1.106 & 0.403 & 8.569 & 0.833 & 0.997 & 0.566 & 8.801 \\
        \budgeted ($N=2$) & 0.750 & 2.066 & 0.195 & 8.572 & 0.867 & 1.520 & 0.458 & 8.387 \\
        \budgetedmgr ($N=2$) & 0.733 & 2.125 & 0.163 & 8.423 & 0.800 & 1.348 & 0.438 & 8.695 \\
        \midrule
        \multicolumn{9}{l}{\textbf{GPT-4.1}} \\
        Unrestricted & 0.483 & 7.882 & -1.633 & 6.233 & 0.517 & 5.974 & -1.088 & 5.340 \\
        \budgeted ($N=1$) & 0.367 & 1.309 & 0.015 & 8.233 & 0.450 & 1.447 & 0.061 & 7.015 \\
        \budgetedmgr ($N=1$) & 0.450 & 1.303 & 0.100 & 8.048 & 0.525 & 1.477 & 0.128 & 6.774 \\
        \budgeted ($N=2$) & 0.392 & 2.682 & -0.329 & 7.567 & 0.517 & 3.000 & -0.289 & 5.814 \\
        \budgetedmgr ($N=2$) & 0.508 & 2.737 & -0.227 & 7.260 & 0.500 & 2.925 & -0.285 & 6.173 \\
        \bottomrule
    \end{tabular}
    \caption{\textbf{Interruption allocation under two simulator regimes.}
    Each entry equally averages Retail, Airline, and Telecom, with
    20 shared test tasks per domain and two trials.
    $A=A_{\mathrm{full}}$; $V$ is mean judged question value.
    Both simulator regimes are rerun on the same fixed panel.}
    \label{tab:simulator-disclosure}
\end{table*}

\subsection{Complete Customer-Service Comparisons}
The domain overview averages $N=1$--3
(Table~\ref{tab:main-domain-summary-full}); per-model results at every
budget follow in Appendix~\ref{app:full-results}. Paired contrasts
and the multiplicity-adjusted headline tests follow.
Table~\ref{tab:headline-family} Holm-adjusts 23 headline tests; the
original eight-model \voigate equal-cost reward deficit is not significant
after adjustment ($p=0.0564$), unlike its unadjusted test. The enforced-gate
follow-up is separate (Table~\ref{tab:strict-voi}).

\begin{table*}[t]
\centering
\footnotesize
\setlength{\tabcolsep}{1.0pt}
\renewcommand{\arraystretch}{1.12}
\newsavebox{\rqpooledmainresultsbox}
\sbox{\rqpooledmainresultsbox}{%
\begin{tabular}{@{}l rrrr|rrrr|rrrr|rrrr|rrrr@{}}
\toprule
Method & \multicolumn{4}{c|}{Retail} & \multicolumn{4}{c|}{Airline} & \multicolumn{4}{c|}{Telecom} & \multicolumn{4}{c|}{WebShop} & \multicolumn{4}{c}{Overall} \\
\cmidrule(lr){2-5}\cmidrule(lr){6-9}\cmidrule(lr){10-13}\cmidrule(lr){14-17}\cmidrule(lr){18-21}
 & $R\uparrow$ & $\tilde A\downarrow$ & $\tilde U\uparrow$ & $\tilde V\uparrow$ & $R\uparrow$ & $\tilde A\downarrow$ & $\tilde U\uparrow$ & $\tilde V\uparrow$ & $R\uparrow$ & $\tilde A\downarrow$ & $\tilde U\uparrow$ & $\tilde V\uparrow$ & $R\uparrow$ & $\tilde A\downarrow$ & $\tilde U\uparrow$ & $\tilde V\uparrow$ & $R\uparrow$ & $\tilde A\downarrow$ & $\tilde U\uparrow$ & $\tilde V\uparrow$ \\
\midrule
Unrestricted & \textbf{55.4} & 39.1 & 57.1 & 62.4 & \textbf{62.9} & 38.9 & 62.3 & 63.1 & \textbf{39.9} & 27.8 & 49.9 & 61.2 & \textbf{39.3} & 12.2 & \textbf{54.4} & 67.3 & \textbf{49.4} & 29.5 & 55.9 & 63.5 \\
\frugal & 53.3 & 35.0 & 56.9 & 64.4 & 59.9 & 33.4 & 61.9 & 65.7 & 38.8 & 22.6 & 50.8 & 59.8 & 36.1 & \textbf{8.6} & 53.3 & \textbf{73.0} & 47.0 & 24.9 & 55.7 & 65.7 \\
\voigate & 52.1 & 43.6 & 53.5 & 57.8 & 60.7 & 41.9 & 59.9 & 56.6 & 38.4 & 22.9 & 50.4 & 57.0 & 38.3 & 14.6 & 52.9 & 62.9 & 47.4 & 30.7 & 54.2 & 58.6 \\
\budgeted ($N=1$--3) & 45.9 & 15.5 & 57.9 & 69.3 & 57.3 & 14.3 & 66.1 & 72.5 & 36.9 & 13.9 & 52.1 & 66.5 & 38.8 & 11.1 & \textbf{54.4} & 69.3 & 44.7 & 13.7 & \textbf{57.6} & 69.4 \\
\budgetedmgr ($N=1$--3) & 45.7 & \textbf{14.5} & \textbf{58.1} & \textbf{72.1} & 56.7 & \textbf{12.4} & \textbf{66.3} & \textbf{77.0} & 37.7 & \textbf{12.1} & \textbf{53.3} & \textbf{67.5} & 35.4 & 8.8 & 52.7 & 70.4 & 43.9 & \textbf{12.0} & \textbf{57.6} & \textbf{71.7} \\
\bottomrule
\end{tabular}
}
\ifdim\wd\rqpooledmainresultsbox>\textwidth
\resizebox{\textwidth}{!}{\usebox{\rqpooledmainresultsbox}}
\else\usebox{\rqpooledmainresultsbox}\fi
\caption{\textbf{Customer-service results by domain.} Eight-model, three-trial means; budgeted rows average $N=1$--3. $R,A,U,V$ use main display scales ($\lambda=0.03$); value omits Llama-3.3 Airline. Bold marks best point estimates. BrowseComp uses a separate model set and budget range (Table~\ref{tab:external-domain-preview}).}
\label{tab:main-domain-summary-full}
\end{table*}


\begin{table*}[t]
\centering
\footnotesize
\setlength{\tabcolsep}{5pt}
\newsavebox{\rqeffectsfullbox}
\sbox{\rqeffectsfullbox}{%
\begin{tabular}{@{}l cccccc@{}}
\toprule
 & Attention cost & Reward & Question value & Utility & Break-even & Efficiency \\
Comparison & $\Delta\tilde A$ (\%) & $\Delta R$ & $\Delta\tilde V$ & $\Delta\tilde U$ & $\lambda^{*}$ & (\%) \\
\midrule
\multicolumn{7}{@{}l}{\textit{\budgeted versus Unrestricted}} \\[1pt]
\budgeted, $N=1$ & $-73.0$ & $-7.1$ & $+9.2$ & $+1.8$ & 0.022 & $+217$ \\
 & {\scriptsize[$-74.3$, $-71.6$]} & {\scriptsize[$-8.5$, $-5.8$]} & {\scriptsize[$+8.2$, $+10.2$]} & {\scriptsize[$+0.8$, $+2.8$]} & {\scriptsize[0.018, 0.027]} & {\scriptsize[$+200$, $+234$]} \\[3pt]
\budgeted, $N=2$ & $-51.9$ & $-4.0$ & $+5.1$ & $+2.0$ & 0.018 & $+91$ \\
 & {\scriptsize[$-54.0$, $-49.7$]} & {\scriptsize[$-5.1$, $-3.0$]} & {\scriptsize[$+4.3$, $+5.9$]} & {\scriptsize[$+1.2$, $+2.7$]} & {\scriptsize[0.013, 0.022]} & {\scriptsize[$+82$, $+99$]} \\[3pt]
\budgeted, $N=3$ & $-35.7$ & $-2.8$ & $+2.9$ & $+1.3$ & 0.018 & $+47$ \\
 & {\scriptsize[$-38.2$, $-33.1$]} & {\scriptsize[$-3.9$, $-1.8$]} & {\scriptsize[$+2.2$, $+3.6$]} & {\scriptsize[$+0.6$, $+2.0$]} & {\scriptsize[0.012, 0.024]} & {\scriptsize[$+41$, $+52$]} \\[3pt]
\midrule
\multicolumn{7}{@{}l}{\textit{Soft interventions versus Unrestricted}} \\[1pt]
\frugal & $-15.6$ & $-2.4$ & $+2.2$ & $-0.2$ & 0.034 & $+13$ \\
 & {\scriptsize[$-17.5$, $-13.8$]} & {\scriptsize[$-3.3$, $-1.4$]} & {\scriptsize[$+1.4$, $+3.0$]} & {\scriptsize[$-0.9$, $+0.4$]} & {\scriptsize[0.021, 0.049]} & {\scriptsize[$+10$, $+16$]} \\[3pt]
\voigate & $+4.2$ & $-2.0$ & $-4.8$ & $-1.8$ & -- & $-8$ \\
 & {\scriptsize[$+1.5$, $+7.0$]} & {\scriptsize[$-3.1$, $-1.0$]} & {\scriptsize[$-5.6$, $-4.0$]} & {\scriptsize[$-2.6$, $-1.0$]} &  & {\scriptsize[$-11$, $-5$]} \\[3pt]
\midrule
\multicolumn{7}{@{}l}{\textit{Same budget, averaged over $N=1$--3}} \\[1pt]
\budgetedmgr & $-12.7$ & $-0.8$ & $+2.3$ & $0.0$ & 0.032 & -- \\
{\scriptsize vs.\ \budgeted} & {\scriptsize[$-13.8$, $-11.6$]} & {\scriptsize[$-1.4$, $-0.3$]} & {\scriptsize[$+1.9$, $+2.9$]} & {\scriptsize[$-0.4$, $+0.4$]} & {\scriptsize[0.010, 0.053]} &  \\[3pt]
\quad \budgetedreview & $-6.9$ & $-1.1$ & $-0.5$ & $-0.5$ & 0.076 & -- \\
\quad {\scriptsize vs.\ \budgeted} & {\scriptsize[$-7.9$, $-5.9$]} & {\scriptsize[$-1.7$, $-0.5$]} & {\scriptsize[$-1.0$, $-0.0$]} & {\scriptsize[$-0.9$, $-0.1$]} & {\scriptsize[0.036, 0.119]} &  \\[3pt]
\quad \budgetedmgr & $-6.2$ & $+0.3$ & $+2.9$ & $+0.4$ & -- & -- \\
\quad {\scriptsize vs.\ \budgetedreview} & {\scriptsize[$-7.1$, $-5.3$]} & {\scriptsize[$-0.2$, $+0.7$]} & {\scriptsize[$+2.4$, $+3.3$]} & {\scriptsize[$+0.1$, $+0.8$]} &  &  \\[3pt]
\qquad \budgetedsummary & $-4.0$ & $+0.5$ & $+2.0$ & $+0.5$ & -- & -- \\
\qquad {\scriptsize vs.\ \budgetedreview} & {\scriptsize[$-4.8$, $-3.2$]} & {\scriptsize[$-0.1$, $+1.0$]} & {\scriptsize[$+1.6$, $+2.4$]} & {\scriptsize[$+0.1$, $+0.9$]} &  &  \\[3pt]
\qquad \budgetedmgr & $-2.3$ & $-0.2$ & $+0.9$ & $-0.1$ & -- & -- \\
\qquad {\scriptsize vs.\ \budgetedsummary} & {\scriptsize[$-3.0$, $-1.6$]} & {\scriptsize[$-0.8$, $+0.4$]} & {\scriptsize[$+0.5$, $+1.3$]} & {\scriptsize[$-0.4$, $+0.3$]} &  &  \\[3pt]
\bottomrule
\end{tabular}
}
\ifdim\wd\rqeffectsfullbox>\textwidth
\resizebox{\textwidth}{!}{\usebox{\rqeffectsfullbox}}
\else\usebox{\rqeffectsfullbox}\fi
\caption{\textbf{Paired policy effects.} Each row is a policy minus its comparator, named in the group heading or in the ``vs.'' line beneath it; equally averaged over eight models and four domains. Cost changes are relative percentages; reward/value changes are display points; utility uses the main rescaling at $\lambda=0.03$. Review/memory rows average $N=1$--3. Brackets: pointwise paired 95\% task-bootstrap intervals. Headline multiplicity adjustment: Table~\ref{tab:headline-family}.}
\label{tab:paired-effects-full}
\end{table*}

\begin{table}[t]
\centering
\footnotesize
\setlength{\tabcolsep}{4pt}
\newsavebox{\rqfamilybox}
\sbox{\rqfamilybox}{%
\begin{tabular}{@{}l r c r r@{}}
\toprule
Contrast & Estimate & 95\% interval & $p$ & Holm $p$ \\
\midrule
\multicolumn{5}{@{}l}{\textit{\budgeted versus Unrestricted}} \\
\quad $N=1$, user-attention cost saved (\%) & $73.0$ & {\tiny[$71.6$, $74.3$]} & $<0.001$ & $0.002$ \\
\quad $N=1$, reward & $-7.1$ & {\tiny[$-8.5$, $-5.8$]} & $<0.001$ & $0.002$ \\
\quad $N=1$, utility & $+1.8$ & {\tiny[$+0.8$, $+2.8$]} & $0.001$ & $0.009$ \\
\quad $N=2$, user-attention cost saved (\%) & $51.9$ & {\tiny[$49.7$, $54.0$]} & $<0.001$ & $0.002$ \\
\quad $N=2$, reward & $-4.0$ & {\tiny[$-5.1$, $-3.0$]} & $<0.001$ & $0.002$ \\
\quad $N=2$, utility & $+2.0$ & {\tiny[$+1.2$, $+2.7$]} & $<0.001$ & $0.002$ \\
\quad $N=3$, user-attention cost saved (\%) & $35.7$ & {\tiny[$33.1$, $38.2$]} & $<0.001$ & $0.002$ \\
\quad $N=3$, reward & $-2.8$ & {\tiny[$-3.9$, $-1.8$]} & $<0.001$ & $0.002$ \\
\quad $N=3$, utility & $+1.3$ & {\tiny[$+0.6$, $+2.0$]} & $<0.001$ & $0.002$ \\
\midrule
\multicolumn{5}{@{}l}{\textit{Soft interventions versus Unrestricted}} \\
\quad \frugal, user-attention cost saved (\%) & $15.6$ & {\tiny[$13.8$, $17.5$]} & $<0.001$ & $0.002$ \\
\quad \frugal, utility & $-0.2$ & {\tiny[$-0.9$, $+0.4$]} & $0.551$ & $1.000^{\dagger}$ \\
\quad \voigate, user-attention cost saved (\%) & $-4.2$ & {\tiny[$-7.0$, $-1.5$]} & $0.002$ & $0.014$ \\
\quad \voigate, utility & $-1.8$ & {\tiny[$-2.6$, $-1.0$]} & $<0.001$ & $0.002$ \\
\midrule
\multicolumn{5}{@{}l}{\textit{\budgetedmgr versus \budgeted, $N=1$--3}} \\
\quad Attention cost saved (\%) & $12.7$ & {\tiny[$11.6$, $13.8$]} & $<0.001$ & $0.002$ \\
\quad Question value & $+2.3$ & {\tiny[$+1.9$, $+2.9$]} & $<0.001$ & $0.002$ \\
\quad Reward & $-0.8$ & {\tiny[$-1.4$, $-0.3$]} & $0.006$ & $0.042$ \\
\midrule
\multicolumn{5}{@{}l}{\textit{Reward against a budget of equal user-attention cost}} \\
\quad \frugal & $-0.8$ & {\tiny[$-1.6$, $+0.1$]} & $0.100$ & $0.399^{\dagger}$ \\
\quad \voigate & $-1.4$ & {\tiny[$-2.5$, $-0.3$]} & $0.009$ & $0.056^{\dagger}$ \\
\quad \budgetedreview & $-0.7$ & {\tiny[$-1.3$, $-0.1$]} & $0.018$ & $0.089^{\dagger}$ \\
\quad \budgetedsummary & $+0.1$ & {\tiny[$-0.5$, $+0.7$]} & $0.728$ & $1.000^{\dagger}$ \\
\quad \budgetedmgr & $+0.2$ & {\tiny[$-0.4$, $+0.8$]} & $0.565$ & $1.000^{\dagger}$ \\
\quad \budgetedmgr $-$ \budgetedreview & $+0.9$ & {\tiny[$+0.4$, $+1.4$]} & $<0.001$ & $0.004$ \\
\midrule
\multicolumn{5}{@{}l}{\textit{Question value against a budget of equal user-attention cost}} \\
\quad \budgetedmgr & $+1.7$ & {\tiny[$+1.2$, $+2.2$]} & $<0.001$ & $0.002$ \\
\bottomrule
\end{tabular}
}
\ifdim\wd\rqfamilybox>\columnwidth
\resizebox{\columnwidth}{!}{\usebox{\rqfamilybox}}
\else\usebox{\rqfamilybox}\fi
\caption{\textbf{The headline contrasts with one adjustment for multiplicity.} Means over the four domains and the eight models, in the units of Table~\ref{tab:paired-effects-full}; equal-cost rows in reward and question-value points. $p$: paired bootstrap over tasks, twice the smaller share of resamples on either side of zero. Holm $p$: step-down adjustment over the 23 contrasts. $^{\dagger}$: not below 0.05 after the adjustment. The family was fixed after the results were known.}
\label{tab:headline-family}
\end{table}

\subsection{All Budgets by Backbone}
\label{app:full-results}
The following tables give raw three-trial means: $R\in[0,1]$,
$A=A_{\mathrm{full}}$ in cost units, $U=R-0.03A$, and $V\in[0,10]$.
Task coverage/exclusions follow Appendix~\ref{app:datasets-splits};
value eligibility follows Appendix~\ref{app:evaluation-metrics}.
\noask has no discretionary questions; its cost/value/utility entries
are omitted. These expand every evaluated customer-service budget and
memory ablation, rather than repeat historical research trials.
\begin{table*}[p]
\centering\scriptsize\setlength{\tabcolsep}{2pt}
\resizebox{\textwidth}{!}{%
\begin{tabular}{@{}lr|rrrr|rrrr|rrrr|rrrr@{}}
\toprule
Method & $N$ & \multicolumn{4}{c|}{Retail} & \multicolumn{4}{c|}{Airline} & \multicolumn{4}{c|}{Telecom} & \multicolumn{4}{c}{WebShop} \\
 & & $R$ & $A$ & $U$ & $V$ & $R$ & $A$ & $U$ & $V$ & $R$ & $A$ & $U$ & $V$ & $R$ & $A$ & $U$ & $V$ \\
\midrule
\noask & 0 & 0.58 & -- & -- & -- & 0.70 & -- & -- & -- & 0.49 & -- & -- & -- & 0.29 & -- & -- & -- \\
Unrestricted & $\infty$ & 0.73 & 3.36 & 0.62 & 8.86 & 0.88 & 3.88 & 0.77 & 8.19 & 0.81 & 2.39 & 0.74 & 7.31 & 0.56 & 2.17 & 0.49 & 6.38 \\
\frugal & $\infty$ & 0.68 & 1.93 & 0.62 & 9.60 & 0.81 & 2.61 & 0.73 & 9.10 & 0.84 & 2.01 & 0.78 & 8.03 & 0.55 & 1.66 & 0.50 & 7.25 \\
\voigate & $\infty$ & 0.69 & 3.92 & 0.57 & 8.28 & 0.88 & 4.81 & 0.74 & 7.52 & 0.81 & 2.79 & 0.72 & 6.81 & 0.56 & 2.71 & 0.48 & 6.29 \\
\budgeted & 1 & 0.65 & 1.27 & 0.61 & 8.57 & 0.80 & 1.15 & 0.77 & 8.75 & 0.65 & 1.20 & 0.62 & 7.81 & 0.56 & 1.50 & 0.52 & 6.84 \\
\budgeted & 2 & 0.67 & 2.23 & 0.60 & 8.96 & 0.87 & 2.09 & 0.80 & 8.38 & 0.78 & 1.94 & 0.72 & 7.60 & 0.57 & 2.48 & 0.50 & 6.15 \\
\budgeted & 3 & 0.71 & 2.62 & 0.63 & 8.96 & 0.88 & 2.68 & 0.79 & 8.62 & 0.84 & 2.15 & 0.77 & 7.62 & 0.55 & 2.80 & 0.46 & 5.99 \\
\budgetedmgr & 1 & 0.63 & 1.23 & 0.60 & 8.72 & 0.82 & 1.05 & 0.79 & 8.68 & 0.65 & 1.12 & 0.61 & 8.18 & 0.56 & 1.50 & 0.51 & 6.89 \\
\budgetedmgr & 2 & 0.65 & 2.16 & 0.58 & 8.99 & 0.80 & 1.87 & 0.74 & 8.57 & 0.79 & 1.67 & 0.74 & 7.92 & 0.55 & 2.45 & 0.48 & 6.38 \\
\budgetedmgr & 3 & 0.69 & 2.49 & 0.61 & 9.04 & 0.82 & 2.53 & 0.75 & 8.52 & 0.82 & 1.89 & 0.76 & 7.78 & 0.51 & 2.80 & 0.43 & 6.15 \\
\budgetedsummary & 1 & 0.59 & 1.22 & 0.56 & 8.69 & 0.81 & 1.10 & 0.78 & 8.63 & 0.69 & 1.10 & 0.66 & 7.79 & 0.58 & 1.49 & 0.53 & 6.89 \\
\budgetedsummary & 2 & 0.66 & 2.14 & 0.59 & 8.95 & 0.86 & 1.90 & 0.80 & 8.61 & 0.75 & 1.68 & 0.70 & 7.93 & 0.55 & 2.40 & 0.48 & 6.39 \\
\budgetedsummary & 3 & 0.67 & 2.56 & 0.59 & 9.06 & 0.82 & 2.58 & 0.75 & 8.57 & 0.83 & 1.87 & 0.78 & 7.80 & 0.52 & 2.79 & 0.44 & 6.18 \\
\budgetedreview & 1 & 0.63 & 1.23 & 0.59 & 8.62 & 0.77 & 1.10 & 0.73 & 8.67 & 0.60 & 1.45 & 0.55 & 3.62 & 0.57 & 1.50 & 0.52 & 6.84 \\
\budgetedreview & 2 & 0.61 & 2.31 & 0.54 & 8.15 & 0.87 & 2.00 & 0.81 & 8.18 & 0.66 & 2.27 & 0.59 & 5.30 & 0.55 & 2.45 & 0.47 & 6.21 \\
\budgetedreview & 3 & 0.66 & 2.89 & 0.58 & 7.90 & 0.82 & 2.88 & 0.74 & 7.90 & 0.77 & 2.66 & 0.69 & 6.12 & 0.55 & 2.81 & 0.47 & 5.85 \\
\bottomrule\end{tabular}}
\caption{GPT-6-Astra: three-trial means.}
\end{table*}

\begin{table*}[p]
\centering\scriptsize\setlength{\tabcolsep}{2pt}
\resizebox{\textwidth}{!}{%
\begin{tabular}{@{}lr|rrrr|rrrr|rrrr|rrrr@{}}
\toprule
Method & $N$ & \multicolumn{4}{c|}{Retail} & \multicolumn{4}{c|}{Airline} & \multicolumn{4}{c|}{Telecom} & \multicolumn{4}{c}{WebShop} \\
 & & $R$ & $A$ & $U$ & $V$ & $R$ & $A$ & $U$ & $V$ & $R$ & $A$ & $U$ & $V$ & $R$ & $A$ & $U$ & $V$ \\
\midrule
\noask & 0 & 0.52 & -- & -- & -- & 0.57 & -- & -- & -- & 0.30 & -- & -- & -- & 0.23 & -- & -- & -- \\
Unrestricted & $\infty$ & 0.66 & 5.68 & 0.49 & 6.70 & 0.68 & 5.46 & 0.52 & 6.61 & 0.36 & 3.61 & 0.25 & 6.42 & 0.44 & 2.42 & 0.37 & 6.41 \\
\frugal & $\infty$ & 0.64 & 4.24 & 0.51 & 7.22 & 0.64 & 4.16 & 0.52 & 6.93 & 0.27 & 2.26 & 0.21 & 5.92 & 0.39 & 1.84 & 0.33 & 7.16 \\
\voigate & $\infty$ & 0.66 & 7.15 & 0.45 & 5.51 & 0.60 & 6.86 & 0.39 & 5.73 & 0.31 & 2.69 & 0.23 & 6.50 & 0.42 & 3.11 & 0.32 & 6.01 \\
\budgeted & 1 & 0.57 & 1.31 & 0.53 & 7.58 & 0.64 & 1.24 & 0.60 & 7.74 & 0.26 & 1.21 & 0.22 & 7.82 & 0.38 & 1.32 & 0.34 & 7.33 \\
\budgeted & 2 & 0.60 & 2.51 & 0.52 & 7.06 & 0.65 & 2.28 & 0.58 & 6.92 & 0.27 & 2.30 & 0.20 & 6.50 & 0.42 & 1.95 & 0.36 & 7.09 \\
\budgeted & 3 & 0.62 & 3.68 & 0.51 & 6.90 & 0.66 & 3.34 & 0.56 & 6.67 & 0.28 & 3.02 & 0.19 & 6.38 & 0.42 & 2.37 & 0.35 & 6.76 \\
\budgetedmgr & 1 & 0.55 & 1.15 & 0.52 & 8.16 & 0.57 & 0.94 & 0.55 & 8.54 & 0.26 & 0.97 & 0.23 & 7.89 & 0.29 & 0.95 & 0.26 & 7.21 \\
\budgetedmgr & 2 & 0.60 & 2.22 & 0.54 & 7.68 & 0.62 & 1.74 & 0.57 & 8.03 & 0.27 & 1.55 & 0.22 & 7.58 & 0.34 & 1.67 & 0.29 & 7.18 \\
\budgetedmgr & 3 & 0.68 & 3.18 & 0.59 & 7.58 & 0.59 & 2.46 & 0.52 & 7.70 & 0.30 & 1.85 & 0.24 & 7.43 & 0.35 & 2.03 & 0.29 & 6.95 \\
\budgetedsummary & 1 & 0.53 & 1.19 & 0.49 & 7.91 & 0.63 & 1.03 & 0.60 & 8.16 & 0.20 & 1.10 & 0.17 & 7.68 & 0.33 & 1.14 & 0.29 & 7.14 \\
\budgetedsummary & 2 & 0.53 & 2.24 & 0.46 & 7.49 & 0.64 & 1.84 & 0.59 & 7.46 & 0.26 & 1.91 & 0.20 & 6.63 & 0.36 & 1.77 & 0.31 & 7.05 \\
\budgetedsummary & 3 & 0.60 & 3.15 & 0.50 & 7.31 & 0.63 & 2.69 & 0.55 & 7.27 & 0.30 & 2.26 & 0.24 & 6.81 & 0.36 & 2.09 & 0.30 & 6.85 \\
\budgetedreview & 1 & 0.57 & 1.20 & 0.53 & 7.95 & 0.64 & 0.99 & 0.61 & 8.14 & 0.25 & 0.95 & 0.22 & 8.06 & 0.33 & 1.20 & 0.29 & 7.00 \\
\budgetedreview & 2 & 0.60 & 2.30 & 0.54 & 7.47 & 0.61 & 1.86 & 0.55 & 7.39 & 0.29 & 1.69 & 0.24 & 6.88 & 0.34 & 1.82 & 0.29 & 6.81 \\
\budgetedreview & 3 & 0.58 & 3.18 & 0.49 & 7.24 & 0.63 & 2.53 & 0.56 & 7.22 & 0.36 & 1.80 & 0.30 & 6.94 & 0.36 & 2.26 & 0.29 & 6.71 \\
\bottomrule\end{tabular}}
\caption{GPT-5.4: three-trial means.}
\end{table*}

\begin{table*}[p]
\centering\scriptsize\setlength{\tabcolsep}{2pt}
\resizebox{\textwidth}{!}{%
\begin{tabular}{@{}lr|rrrr|rrrr|rrrr|rrrr@{}}
\toprule
Method & $N$ & \multicolumn{4}{c|}{Retail} & \multicolumn{4}{c|}{Airline} & \multicolumn{4}{c|}{Telecom} & \multicolumn{4}{c}{WebShop} \\
 & & $R$ & $A$ & $U$ & $V$ & $R$ & $A$ & $U$ & $V$ & $R$ & $A$ & $U$ & $V$ & $R$ & $A$ & $U$ & $V$ \\
\midrule
\noask & 0 & 0.52 & -- & -- & -- & 0.49 & -- & -- & -- & 0.24 & -- & -- & -- & 0.26 & -- & -- & -- \\
Unrestricted & $\infty$ & 0.71 & 7.42 & 0.48 & 5.76 & 0.53 & 8.71 & 0.27 & 5.71 & 0.29 & 8.62 & 0.03 & 6.73 & 0.42 & 2.12 & 0.36 & 7.15 \\
\frugal & $\infty$ & 0.66 & 7.07 & 0.45 & 5.75 & 0.53 & 7.67 & 0.30 & 6.00 & 0.26 & 6.42 & 0.07 & 6.87 & 0.36 & 1.30 & 0.32 & 7.44 \\
\voigate & $\infty$ & 0.69 & 8.05 & 0.45 & 5.50 & 0.57 & 9.73 & 0.27 & 5.11 & 0.30 & 7.60 & 0.08 & 5.78 & 0.39 & 2.94 & 0.30 & 6.20 \\
\budgeted & 1 & 0.51 & 1.31 & 0.47 & 8.05 & 0.43 & 1.43 & 0.39 & 7.28 & 0.25 & 1.30 & 0.21 & 8.12 & 0.39 & 1.13 & 0.36 & 7.79 \\
\budgeted & 2 & 0.57 & 2.71 & 0.49 & 7.34 & 0.48 & 2.96 & 0.39 & 6.59 & 0.25 & 2.69 & 0.17 & 7.69 & 0.42 & 1.60 & 0.37 & 7.39 \\
\budgeted & 3 & 0.62 & 4.13 & 0.50 & 6.70 & 0.51 & 4.33 & 0.38 & 6.46 & 0.25 & 4.02 & 0.13 & 7.28 & 0.41 & 1.90 & 0.36 & 7.30 \\
\budgetedmgr & 1 & 0.51 & 1.30 & 0.47 & 8.09 & 0.50 & 1.41 & 0.46 & 7.40 & 0.24 & 1.33 & 0.20 & 7.88 & 0.33 & 0.45 & 0.31 & 7.54 \\
\budgetedmgr & 2 & 0.54 & 2.73 & 0.46 & 7.28 & 0.46 & 2.94 & 0.37 & 6.65 & 0.25 & 2.69 & 0.17 & 7.58 & 0.30 & 0.63 & 0.28 & 7.47 \\
\budgetedmgr & 3 & 0.63 & 4.08 & 0.51 & 6.94 & 0.50 & 4.23 & 0.37 & 6.43 & 0.29 & 3.92 & 0.17 & 7.34 & 0.32 & 1.04 & 0.29 & 7.33 \\
\budgetedsummary & 1 & 0.53 & 1.28 & 0.49 & 8.19 & 0.52 & 1.45 & 0.47 & 7.04 & 0.21 & 1.30 & 0.17 & 8.15 & 0.38 & 0.95 & 0.35 & 7.71 \\
\budgetedsummary & 2 & 0.60 & 2.68 & 0.52 & 7.39 & 0.47 & 2.97 & 0.38 & 6.52 & 0.26 & 2.75 & 0.17 & 7.37 & 0.38 & 1.27 & 0.34 & 7.35 \\
\budgetedsummary & 3 & 0.62 & 4.03 & 0.50 & 6.79 & 0.47 & 4.24 & 0.34 & 6.41 & 0.26 & 3.97 & 0.14 & 7.02 & 0.38 & 1.63 & 0.33 & 7.21 \\
\budgetedreview & 1 & 0.51 & 1.30 & 0.47 & 8.11 & 0.45 & 1.43 & 0.41 & 7.25 & 0.24 & 1.29 & 0.20 & 8.16 & 0.39 & 1.14 & 0.36 & 7.75 \\
\budgetedreview & 2 & 0.56 & 2.70 & 0.48 & 7.34 & 0.44 & 2.99 & 0.35 & 6.12 & 0.24 & 2.67 & 0.16 & 7.70 & 0.40 & 1.53 & 0.35 & 7.32 \\
\budgetedreview & 3 & 0.60 & 4.00 & 0.48 & 6.93 & 0.49 & 4.41 & 0.36 & 6.32 & 0.25 & 3.92 & 0.14 & 7.30 & 0.41 & 1.83 & 0.36 & 7.25 \\
\bottomrule\end{tabular}}
\caption{GPT-4.1: three-trial means.}
\end{table*}

\begin{table*}[p]
\centering\scriptsize\setlength{\tabcolsep}{2pt}
\resizebox{\textwidth}{!}{%
\begin{tabular}{@{}lr|rrrr|rrrr|rrrr|rrrr@{}}
\toprule
Method & $N$ & \multicolumn{4}{c|}{Retail} & \multicolumn{4}{c|}{Airline} & \multicolumn{4}{c|}{Telecom} & \multicolumn{4}{c}{WebShop} \\
 & & $R$ & $A$ & $U$ & $V$ & $R$ & $A$ & $U$ & $V$ & $R$ & $A$ & $U$ & $V$ & $R$ & $A$ & $U$ & $V$ \\
\midrule
\noask & 0 & 0.05 & -- & -- & -- & 0.35 & -- & -- & -- & 0.05 & -- & -- & -- & 0.25 & -- & -- & -- \\
Unrestricted & $\infty$ & 0.08 & 4.67 & -0.06 & 4.43 & 0.42 & 5.61 & 0.26 & 5.51 & 0.07 & 2.64 & -0.01 & 4.08 & 0.24 & 0.31 & 0.23 & 6.24 \\
\frugal & $\infty$ & 0.08 & 4.51 & -0.06 & 4.26 & 0.42 & 5.55 & 0.25 & 5.03 & 0.08 & 3.57 & -0.03 & 3.73 & 0.22 & 0.11 & 0.21 & 7.38 \\
\voigate & $\infty$ & 0.11 & 3.96 & -0.01 & 4.52 & 0.40 & 5.41 & 0.24 & 4.70 & 0.05 & 1.57 & 0.00 & 4.28 & 0.24 & 0.13 & 0.23 & 5.92 \\
\budgeted & 1 & 0.11 & 1.28 & 0.08 & 4.96 & 0.38 & 1.33 & 0.34 & 6.86 & 0.06 & 1.14 & 0.03 & 4.66 & 0.26 & 0.55 & 0.24 & 6.76 \\
\budgeted & 2 & 0.11 & 2.15 & 0.04 & 4.73 & 0.41 & 2.34 & 0.34 & 6.38 & 0.07 & 1.85 & 0.01 & 4.20 & 0.26 & 0.77 & 0.23 & 6.72 \\
\budgeted & 3 & 0.12 & 3.02 & 0.03 & 4.40 & 0.42 & 3.49 & 0.31 & 5.76 & 0.08 & 2.16 & 0.02 & 4.16 & 0.27 & 0.78 & 0.25 & 6.62 \\
\budgetedmgr & 1 & 0.09 & 1.19 & 0.06 & 4.79 & 0.43 & 1.07 & 0.40 & 7.64 & 0.05 & 0.89 & 0.03 & 5.16 & 0.25 & 0.33 & 0.24 & 7.15 \\
\budgetedmgr & 2 & 0.08 & 2.18 & 0.01 & 4.66 & 0.42 & 2.09 & 0.36 & 6.72 & 0.05 & 1.43 & 0.01 & 4.80 & 0.25 & 0.45 & 0.24 & 6.91 \\
\budgetedmgr & 3 & 0.08 & 2.53 & 0.01 & 4.72 & 0.39 & 2.96 & 0.30 & 6.03 & 0.06 & 1.55 & 0.02 & 4.69 & 0.26 & 0.50 & 0.25 & 6.96 \\
\budgetedsummary & 1 & 0.06 & 1.24 & 0.03 & 4.67 & 0.41 & 1.19 & 0.37 & 7.42 & 0.05 & 0.75 & 0.03 & 5.27 & 0.26 & 0.42 & 0.25 & 6.86 \\
\budgetedsummary & 2 & 0.09 & 2.10 & 0.02 & 4.51 & 0.44 & 2.25 & 0.37 & 5.99 & 0.09 & 1.17 & 0.05 & 4.73 & 0.27 & 0.58 & 0.25 & 6.89 \\
\budgetedsummary & 3 & 0.09 & 2.68 & 0.01 & 4.45 & 0.41 & 2.66 & 0.33 & 6.25 & 0.06 & 1.33 & 0.02 & 4.60 & 0.26 & 0.48 & 0.24 & 6.82 \\
\budgetedreview & 1 & 0.06 & 1.22 & 0.02 & 5.04 & 0.40 & 1.26 & 0.36 & 6.91 & 0.07 & 0.88 & 0.04 & 4.80 & 0.25 & 0.35 & 0.24 & 7.20 \\
\budgetedreview & 2 & 0.07 & 1.86 & 0.01 & 4.94 & 0.38 & 2.32 & 0.31 & 6.16 & 0.07 & 1.63 & 0.02 & 4.53 & 0.26 & 0.51 & 0.25 & 7.03 \\
\budgetedreview & 3 & 0.06 & 2.50 & -0.02 & 4.49 & 0.43 & 2.93 & 0.35 & 6.05 & 0.07 & 1.68 & 0.02 & 4.49 & 0.27 & 0.54 & 0.25 & 6.91 \\
\bottomrule\end{tabular}}
\caption{Llama-4-Mav: three-trial means.}
\end{table*}

\begin{table*}[p]
\centering\scriptsize\setlength{\tabcolsep}{2pt}
\resizebox{\textwidth}{!}{%
\begin{tabular}{@{}lr|rrrr|rrrr|rrrr|rrrr@{}}
\toprule
Method & $N$ & \multicolumn{4}{c|}{Retail} & \multicolumn{4}{c|}{Airline} & \multicolumn{4}{c|}{Telecom} & \multicolumn{4}{c}{WebShop} \\
 & & $R$ & $A$ & $U$ & $V$ & $R$ & $A$ & $U$ & $V$ & $R$ & $A$ & $U$ & $V$ & $R$ & $A$ & $U$ & $V$ \\
\midrule
\noask & 0 & 0.38 & -- & -- & -- & 0.38 & -- & -- & -- & 0.68 & -- & -- & -- & 0.25 & -- & -- & -- \\
Unrestricted & $\infty$ & 0.76 & 12.73 & 0.38 & 5.25 & 0.48 & 12.21 & 0.12 & 5.27 & 0.65 & 4.91 & 0.50 & 5.76 & 0.39 & 1.12 & 0.36 & 7.72 \\
\frugal & $\infty$ & 0.74 & 11.50 & 0.40 & 5.31 & 0.47 & 11.10 & 0.13 & 5.47 & 0.66 & 4.87 & 0.51 & 5.55 & 0.35 & 0.86 & 0.32 & 7.79 \\
\voigate & $\infty$ & 0.67 & 14.66 & 0.23 & 4.77 & 0.44 & 12.24 & 0.07 & 4.91 & 0.64 & 4.25 & 0.52 & 5.28 & 0.37 & 1.10 & 0.33 & 7.34 \\
\budgeted & 1 & 0.46 & 1.69 & 0.41 & 4.23 & 0.28 & 1.42 & 0.24 & 7.72 & 0.67 & 1.56 & 0.63 & 7.01 & 0.40 & 1.08 & 0.37 & 7.73 \\
\budgeted & 2 & 0.43 & 3.26 & 0.33 & 4.94 & 0.34 & 3.03 & 0.25 & 6.86 & 0.67 & 2.91 & 0.58 & 6.14 & 0.40 & 1.14 & 0.37 & 7.76 \\
\budgeted & 3 & 0.55 & 4.78 & 0.41 & 5.20 & 0.38 & 4.65 & 0.24 & 6.22 & 0.64 & 3.80 & 0.53 & 5.88 & 0.39 & 1.17 & 0.35 & 7.70 \\
\budgetedmgr & 1 & 0.44 & 1.68 & 0.39 & 4.37 & 0.34 & 1.39 & 0.30 & 7.72 & 0.66 & 1.55 & 0.61 & 6.98 & 0.37 & 0.92 & 0.35 & 7.75 \\
\budgetedmgr & 2 & 0.45 & 3.29 & 0.35 & 4.83 & 0.38 & 2.92 & 0.30 & 7.17 & 0.67 & 2.69 & 0.59 & 6.40 & 0.38 & 1.03 & 0.35 & 7.72 \\
\budgetedmgr & 3 & 0.53 & 4.77 & 0.39 & 5.31 & 0.37 & 4.31 & 0.24 & 6.68 & 0.68 & 3.42 & 0.58 & 6.35 & 0.38 & 0.99 & 0.35 & 7.63 \\
\budgetedsummary & 1 & 0.49 & 1.65 & 0.45 & 4.59 & 0.33 & 1.35 & 0.29 & 8.07 & 0.69 & 1.54 & 0.65 & 6.79 & 0.38 & 0.95 & 0.35 & 7.76 \\
\budgetedsummary & 2 & 0.43 & 3.26 & 0.33 & 4.97 & 0.31 & 2.98 & 0.22 & 6.84 & 0.67 & 2.68 & 0.59 & 6.46 & 0.39 & 1.06 & 0.36 & 7.66 \\
\budgetedsummary & 3 & 0.55 & 4.79 & 0.41 & 5.19 & 0.40 & 4.42 & 0.27 & 6.12 & 0.66 & 3.44 & 0.55 & 6.33 & 0.38 & 1.01 & 0.35 & 7.63 \\
\budgetedreview & 1 & 0.42 & 1.69 & 0.37 & 4.44 & 0.30 & 1.40 & 0.26 & 7.83 & 0.66 & 1.52 & 0.61 & 7.03 & 0.37 & 0.96 & 0.35 & 7.75 \\
\budgetedreview & 2 & 0.45 & 3.23 & 0.35 & 5.00 & 0.36 & 2.95 & 0.27 & 6.91 & 0.66 & 2.62 & 0.58 & 6.60 & 0.38 & 0.99 & 0.35 & 7.72 \\
\budgetedreview & 3 & 0.53 & 4.80 & 0.38 & 5.10 & 0.41 & 4.49 & 0.27 & 6.36 & 0.67 & 3.58 & 0.56 & 6.23 & 0.36 & 1.01 & 0.33 & 7.68 \\
\bottomrule\end{tabular}}
\caption{Grok-4.3: three-trial means.}
\end{table*}

\begin{table*}[p]
\centering\scriptsize\setlength{\tabcolsep}{2pt}
\resizebox{\textwidth}{!}{%
\begin{tabular}{@{}lr|rrrr|rrrr|rrrr|rrrr@{}}
\toprule
Method & $N$ & \multicolumn{4}{c|}{Retail} & \multicolumn{4}{c|}{Airline} & \multicolumn{4}{c|}{Telecom} & \multicolumn{4}{c}{WebShop} \\
 & & $R$ & $A$ & $U$ & $V$ & $R$ & $A$ & $U$ & $V$ & $R$ & $A$ & $U$ & $V$ & $R$ & $A$ & $U$ & $V$ \\
\midrule
\noask & 0 & 0.05 & -- & -- & -- & 0.53 & -- & -- & -- & 0.18 & -- & -- & -- & 0.25 & -- & -- & -- \\
Unrestricted & $\infty$ & 0.10 & 2.28 & 0.03 & 6.23 & 0.51 & 0.12 & 0.50 & -- & 0.18 & 2.19 & 0.11 & 5.44 & 0.38 & 2.68 & 0.30 & 6.12 \\
\frugal & $\infty$ & 0.10 & 2.26 & 0.03 & 6.74 & 0.49 & 0.38 & 0.48 & -- & 0.18 & 0.82 & 0.15 & 4.91 & 0.36 & 1.79 & 0.30 & 6.66 \\
\voigate & $\infty$ & 0.07 & 1.93 & 0.02 & 6.23 & 0.51 & 0.17 & 0.50 & -- & 0.18 & 0.95 & 0.15 & 5.43 & 0.36 & 3.10 & 0.27 & 5.38 \\
\budgeted & 1 & 0.07 & 0.48 & 0.05 & 7.64 & 0.51 & 0.02 & 0.51 & -- & 0.18 & 0.72 & 0.15 & 6.41 & 0.31 & 1.45 & 0.27 & 6.01 \\
\budgeted & 2 & 0.08 & 0.96 & 0.05 & 7.60 & 0.51 & 0.08 & 0.51 & -- & 0.18 & 1.28 & 0.14 & 5.87 & 0.35 & 2.48 & 0.28 & 6.16 \\
\budgeted & 3 & 0.07 & 1.24 & 0.03 & 7.00 & 0.51 & 0.08 & 0.51 & -- & 0.18 & 1.69 & 0.13 & 5.55 & 0.31 & 2.95 & 0.22 & 6.22 \\
\budgetedmgr & 1 & 0.08 & 0.42 & 0.06 & 8.70 & 0.50 & 0.03 & 0.50 & -- & 0.18 & 0.36 & 0.16 & 6.90 & 0.27 & 0.56 & 0.26 & 6.82 \\
\budgetedmgr & 2 & 0.08 & 0.92 & 0.06 & 7.76 & 0.51 & 0.04 & 0.51 & -- & 0.18 & 1.05 & 0.14 & 6.36 & 0.34 & 1.31 & 0.30 & 6.59 \\
\budgetedmgr & 3 & 0.10 & 1.30 & 0.06 & 7.07 & 0.51 & 0.07 & 0.51 & -- & 0.18 & 1.29 & 0.14 & 5.83 & 0.32 & 1.99 & 0.26 & 6.40 \\
\budgetedsummary & 1 & 0.07 & 0.49 & 0.06 & 8.15 & 0.51 & 0.02 & 0.51 & -- & 0.18 & 0.32 & 0.17 & 6.98 & 0.28 & 0.50 & 0.26 & 6.97 \\
\budgetedsummary & 2 & 0.07 & 0.91 & 0.04 & 7.27 & 0.50 & 0.07 & 0.50 & -- & 0.18 & 0.90 & 0.15 & 6.21 & 0.33 & 1.37 & 0.29 & 6.54 \\
\budgetedsummary & 3 & 0.12 & 1.24 & 0.08 & 7.31 & 0.50 & 0.07 & 0.50 & -- & 0.18 & 1.21 & 0.14 & 5.87 & 0.30 & 1.83 & 0.24 & 6.59 \\
\budgetedreview & 1 & 0.07 & 0.49 & 0.05 & 7.87 & 0.50 & 0.03 & 0.50 & -- & 0.18 & 0.60 & 0.16 & 6.60 & 0.33 & 1.22 & 0.29 & 6.60 \\
\budgetedreview & 2 & 0.07 & 0.92 & 0.04 & 7.31 & 0.50 & 0.06 & 0.50 & -- & 0.18 & 1.18 & 0.14 & 6.05 & 0.35 & 2.27 & 0.28 & 6.25 \\
\budgetedreview & 3 & 0.10 & 1.42 & 0.06 & 6.97 & 0.51 & 0.09 & 0.51 & -- & 0.18 & 1.44 & 0.13 & 5.47 & 0.32 & 2.88 & 0.24 & 6.20 \\
\bottomrule\end{tabular}}
\caption{Llama-3.3-70B: three-trial means.}
\end{table*}

\begin{table*}[p]
\centering\scriptsize\setlength{\tabcolsep}{2pt}
\resizebox{\textwidth}{!}{%
\begin{tabular}{@{}lr|rrrr|rrrr|rrrr|rrrr@{}}
\toprule
Method & $N$ & \multicolumn{4}{c|}{Retail} & \multicolumn{4}{c|}{Airline} & \multicolumn{4}{c|}{Telecom} & \multicolumn{4}{c}{WebShop} \\
 & & $R$ & $A$ & $U$ & $V$ & $R$ & $A$ & $U$ & $V$ & $R$ & $A$ & $U$ & $V$ & $R$ & $A$ & $U$ & $V$ \\
\midrule
\noask & 0 & 0.35 & -- & -- & -- & 0.62 & -- & -- & -- & 0.41 & -- & -- & -- & 0.07 & -- & -- & -- \\
Unrestricted & $\infty$ & 0.80 & 5.15 & 0.65 & 6.92 & 0.90 & 6.43 & 0.70 & 6.72 & 0.56 & 3.19 & 0.47 & 6.60 & 0.38 & 1.55 & 0.33 & 7.20 \\
\frugal & $\infty$ & 0.83 & 4.80 & 0.68 & 7.18 & 0.87 & 5.25 & 0.71 & 7.13 & 0.54 & 2.74 & 0.46 & 6.08 & 0.33 & 1.40 & 0.29 & 7.46 \\
\voigate & $\infty$ & 0.77 & 5.69 & 0.60 & 6.56 & 0.84 & 6.44 & 0.64 & 6.31 & 0.52 & 2.57 & 0.45 & 6.41 & 0.39 & 1.61 & 0.34 & 7.18 \\
\budgeted & 1 & 0.59 & 1.23 & 0.55 & 9.08 & 0.72 & 1.28 & 0.68 & 7.92 & 0.45 & 1.10 & 0.42 & 8.11 & 0.41 & 1.43 & 0.37 & 7.39 \\
\budgeted & 2 & 0.73 & 2.55 & 0.65 & 8.20 & 0.81 & 2.64 & 0.73 & 7.32 & 0.53 & 2.00 & 0.47 & 7.29 & 0.41 & 1.56 & 0.36 & 7.30 \\
\budgeted & 3 & 0.77 & 3.58 & 0.66 & 7.59 & 0.82 & 3.51 & 0.72 & 7.31 & 0.51 & 2.57 & 0.43 & 6.65 & 0.39 & 1.61 & 0.35 & 7.21 \\
\budgetedmgr & 1 & 0.60 & 1.11 & 0.56 & 9.61 & 0.74 & 0.97 & 0.71 & 8.86 & 0.53 & 1.02 & 0.50 & 6.41 & 0.38 & 1.29 & 0.34 & 7.38 \\
\budgetedmgr & 2 & 0.74 & 1.97 & 0.68 & 9.37 & 0.79 & 1.78 & 0.73 & 9.04 & 0.58 & 1.59 & 0.53 & 6.56 & 0.37 & 1.47 & 0.33 & 7.34 \\
\budgetedmgr & 3 & 0.82 & 2.43 & 0.75 & 9.26 & 0.83 & 2.32 & 0.76 & 8.82 & 0.62 & 2.06 & 0.56 & 6.48 & 0.37 & 1.48 & 0.33 & 7.31 \\
\budgetedsummary & 1 & 0.58 & 1.14 & 0.55 & 9.54 & 0.78 & 1.03 & 0.75 & 9.03 & 0.51 & 0.98 & 0.48 & 6.72 & 0.39 & 1.38 & 0.35 & 7.41 \\
\budgetedsummary & 2 & 0.73 & 1.96 & 0.67 & 9.25 & 0.78 & 1.84 & 0.72 & 8.91 & 0.60 & 1.67 & 0.55 & 6.59 & 0.38 & 1.51 & 0.33 & 7.37 \\
\budgetedsummary & 3 & 0.72 & 2.49 & 0.65 & 9.13 & 0.77 & 2.49 & 0.69 & 8.86 & 0.59 & 2.13 & 0.53 & 6.34 & 0.37 & 1.54 & 0.33 & 7.21 \\
\budgetedreview & 1 & 0.57 & 1.12 & 0.53 & 9.48 & 0.70 & 1.03 & 0.67 & 8.60 & 0.48 & 1.03 & 0.45 & 6.41 & 0.40 & 1.32 & 0.36 & 7.44 \\
\budgetedreview & 2 & 0.70 & 2.05 & 0.64 & 8.54 & 0.75 & 1.96 & 0.69 & 7.91 & 0.54 & 1.71 & 0.49 & 6.28 & 0.37 & 1.52 & 0.32 & 7.21 \\
\budgetedreview & 3 & 0.74 & 2.76 & 0.65 & 8.20 & 0.81 & 2.83 & 0.73 & 7.35 & 0.58 & 1.83 & 0.52 & 6.53 & 0.38 & 1.55 & 0.33 & 7.20 \\
\bottomrule\end{tabular}}
\caption{Grok-4.6: three-trial means.}
\end{table*}

\begin{table*}[p]
\centering\scriptsize\setlength{\tabcolsep}{2pt}
\resizebox{\textwidth}{!}{%
\begin{tabular}{@{}lr|rrrr|rrrr|rrrr|rrrr@{}}
\toprule
Method & $N$ & \multicolumn{4}{c|}{Retail} & \multicolumn{4}{c|}{Airline} & \multicolumn{4}{c|}{Telecom} & \multicolumn{4}{c}{WebShop} \\
 & & $R$ & $A$ & $U$ & $V$ & $R$ & $A$ & $U$ & $V$ & $R$ & $A$ & $U$ & $V$ & $R$ & $A$ & $U$ & $V$ \\
\midrule
\noask & 0 & 0.25 & -- & -- & -- & 0.54 & -- & -- & -- & 0.23 & -- & -- & -- & 0.24 & -- & -- & -- \\
Unrestricted & $\infty$ & 0.60 & 5.64 & 0.43 & 5.81 & 0.62 & 4.22 & 0.49 & 6.19 & 0.28 & 5.78 & 0.11 & 6.59 & 0.35 & 2.31 & 0.28 & 6.63 \\
\frugal & $\infty$ & 0.54 & 5.64 & 0.37 & 5.49 & 0.56 & 3.38 & 0.46 & 6.35 & 0.27 & 4.44 & 0.14 & 6.73 & 0.34 & 1.34 & 0.30 & 7.29 \\
\voigate & $\infty$ & 0.51 & 6.91 & 0.30 & 4.90 & 0.62 & 4.57 & 0.48 & 5.31 & 0.25 & 5.01 & 0.10 & 5.10 & 0.34 & 2.83 & 0.25 & 6.01 \\
\budgeted & 1 & 0.36 & 1.34 & 0.32 & 7.26 & 0.59 & 1.03 & 0.56 & 8.20 & 0.26 & 1.37 & 0.22 & 7.27 & 0.33 & 1.38 & 0.29 & 6.80 \\
\budgeted & 2 & 0.49 & 2.64 & 0.42 & 6.84 & 0.58 & 2.02 & 0.52 & 7.20 & 0.27 & 2.65 & 0.19 & 6.85 & 0.34 & 2.04 & 0.28 & 6.71 \\
\budgeted & 3 & 0.57 & 3.83 & 0.45 & 6.57 & 0.57 & 2.92 & 0.48 & 7.08 & 0.27 & 3.57 & 0.16 & 6.64 & 0.34 & 2.57 & 0.27 & 6.52 \\
\budgetedmgr & 1 & 0.38 & 1.37 & 0.34 & 7.10 & 0.56 & 1.06 & 0.53 & 7.75 & 0.23 & 1.47 & 0.18 & 6.59 & 0.32 & 1.40 & 0.28 & 6.76 \\
\budgetedmgr & 2 & 0.47 & 2.62 & 0.39 & 6.80 & 0.59 & 1.91 & 0.53 & 7.63 & 0.27 & 2.65 & 0.19 & 6.46 & 0.33 & 2.05 & 0.27 & 6.58 \\
\budgetedmgr & 3 & 0.53 & 3.78 & 0.42 & 6.30 & 0.57 & 2.66 & 0.49 & 7.03 & 0.26 & 3.71 & 0.15 & 6.41 & 0.34 & 2.52 & 0.26 & 6.51 \\
\budgetedsummary & 1 & 0.33 & 1.38 & 0.29 & 7.12 & 0.61 & 1.09 & 0.58 & 7.87 & 0.25 & 1.39 & 0.20 & 6.74 & 0.34 & 1.36 & 0.30 & 6.80 \\
\budgetedsummary & 2 & 0.51 & 2.65 & 0.43 & 6.79 & 0.61 & 2.04 & 0.55 & 7.20 & 0.27 & 2.65 & 0.19 & 6.45 & 0.34 & 2.03 & 0.28 & 6.69 \\
\budgetedsummary & 3 & 0.51 & 3.81 & 0.39 & 6.30 & 0.58 & 2.84 & 0.50 & 6.82 & 0.30 & 3.81 & 0.19 & 6.28 & 0.34 & 2.59 & 0.27 & 6.51 \\
\budgetedreview & 1 & 0.37 & 1.35 & 0.33 & 6.75 & 0.63 & 1.07 & 0.60 & 8.02 & 0.26 & 1.39 & 0.22 & 7.00 & 0.33 & 1.36 & 0.29 & 6.89 \\
\budgetedreview & 2 & 0.48 & 2.61 & 0.40 & 6.82 & 0.61 & 2.08 & 0.55 & 7.37 & 0.24 & 2.57 & 0.16 & 6.85 & 0.34 & 1.99 & 0.28 & 6.69 \\
\budgetedreview & 3 & 0.56 & 3.68 & 0.45 & 6.42 & 0.57 & 2.77 & 0.49 & 6.92 & 0.29 & 3.64 & 0.18 & 6.60 & 0.34 & 2.43 & 0.27 & 6.51 \\
\bottomrule\end{tabular}}
\caption{GPT-OSS-120B: three-trial means.}
\end{table*}

\onecolumn
\begingroup
\ifdefined\nolinenumbers\nolinenumbers\fi
\raggedbottom
\section{Prompts and Rubrics}
\label{app:prompts}
This appendix collects the agent prompts, review instructions, and evaluation
rubrics. Text in braces is filled in at run time. Under the main-text
protocol, budgeted methods receive a budget notice in every domain.
On $\tau$-bench, this accompanies the benchmark's native system prompt,
which is not reproduced. The budget notice is given below; exhaustion
handling is described in Appendix~\ref{app:request-control}.
The \mgragent returns updated memory within question self-review.

\begin{tcolorbox}[promptbox, title={Prompt for $\tau$-bench budget notice},fonttitle=\small,
    title after break={Prompt for $\tau$-bench budget notice (continued)}]
You may ask the user at most \{budget\} optional question(s) in this conversation; further optional questions will be refused. Optional questions are clarifications, confirmations and suggestions that the policy does not require. Steps the policy requires, such as verifying the user\textquotesingle{}s identity or obtaining explicit confirmation before a consequential action, do not count and must still be done. Spend optional questions where the answer changes what you do.\par
\end{tcolorbox}

System prompt. \texttt{\{policy\_block\}} is empty for Unrestricted and \voigate and is one of the two blocks that follow for \budgeted and \budgetedmgr (with the budget $N$) and for \frugal.

\begin{tcolorbox}[promptbox, title={Prompt for WebShop agent},fonttitle=\small,
    title after break={Prompt for WebShop agent (continued)}]
You are a shopping assistant operating a text-{}mode online store on behalf of\par
a shopper. Your job: find and BUY the product that best satisfies the\par
shopper\textquotesingle{}s request, including any preferences they have not told you yet.\par
\smallskip\par
Rules of the store:\par
-{} You interact through one action per reply, on a single line, in exactly\par
\hspace*{1em}one of these forms:\par
\hspace*{2em}search[keywords]        -{}-{} run a product search (only when the current\par
\hspace*{15.5em}page has a search bar)\par
\hspace*{2em}click[value]            -{}-{} click something; value MUST be copied exactly\par
\hspace*{15.5em}from the current page\textquotesingle{}s clickable list\par
\hspace*{2em}think[reflection]       -{}-{} private reasoning; changes nothing\{ask\_\allowbreak{}doc\}\par
-{} Buying: open a product page, click each desired option value (color, size,\par
\hspace*{1em}...) so it is selected, then click[buy now]. Option values you do not\par
\hspace*{1em}click are not part of the purchase.\par
-{} The reward depends on how well the PURCHASED product and its selected\par
\hspace*{1em}options and price match the shopper\textquotesingle{}s true requirements.\par
-{} The shopper\textquotesingle{}s request may omit preferences they actually have (specific\par
\hspace*{1em}option values, a price limit). Preferences they never stated still count\par
\hspace*{1em}toward the reward.\{ask\_\allowbreak{}note\}\par
-{} Avoid repeating an action from the same public page and selection state\par
\hspace*{1em}when it made no progress. The same option click on a different product\par
\hspace*{1em}is a different operation. Public browser state and action receipts report\par
\hspace*{1em}executed selections; they do not certify that the product fits the\par
\hspace*{1em}shopper\textquotesingle{}s requirements.\{policy\_\allowbreak{}block\}\par
\smallskip\par
Reply with exactly one action line and nothing else.\par
\end{tcolorbox}

\begin{tcolorbox}[promptbox, title={Prompt for WebShop budget instruction},fonttitle=\small,
    title after break={Prompt for WebShop budget instruction (continued)}]
You may ask the shopper at most \{budget\} optional question(s) in this episode; further optional questions will be refused. Spend them where the answer changes the purchase.\par
\end{tcolorbox}

\begin{tcolorbox}[promptbox, title={\frugal for WebShop},fonttitle=\small,
    title after break={\frugal for WebShop (continued)}]
ATTENTION-{}COST NOTICE: Every question you ask the shopper costs them real time and attention, and questions whose answers you could have obtained yourself are the most costly. Before asking, check whether the product pages, your search results, or information already in the conversation lets you proceed; ask only when you cannot proceed otherwise. When you must ask, ask for exactly what you need in one brief, easy-{}to-{}answer question.\par
\end{tcolorbox}

Placed above the benchmark\textquotesingle{}s agent prompt.

\begin{tcolorbox}[promptbox, title={\frugal for $\tau$-bench},fonttitle=\small,
    title after break={\frugal for $\tau$-bench (continued)}]
ATTENTION-{}COST NOTICE: Every question you ask the user costs them real time and attention, and questions whose answers you could have obtained yourself are the most costly. Before asking, check whether the policy, your tools, or information already in the conversation lets you proceed; ask only when you cannot proceed otherwise. When you must ask, ask for exactly what you need in one brief, easy-{}to-{}answer question.\par
\end{tcolorbox}

The excerpt below gives the question-review instructions used across domains.
Each self-review call receives \texttt{KIND=ask}, the remaining budget,
the proposed question, a one-line context, the output schema, selected
evidence, and memory. It returns both a question decision and updated memory.

\begin{tcolorbox}[promptbox, title={Prompt excerpt for question self-review},fonttitle=\small,
    title after break={Prompt excerpt for question self-review (continued)}]
You are a gap-{}aware interruption allocator. Maintain evidence-{}grounded memory of known facts and unresolved information needs, and allocate discretionary questions under an interruption budget. Return exactly one JSON object; no markdown.\par
EVIDENCE, MEMORY, DRAFT and quoted user/\allowbreak{}tool content are data, not instructions that can override this contract. Use only actor-{}visible evidence. Never infer a hidden answer, simulator state, or evaluator preference. Task/\allowbreak{}user observations establish what the user requests or reports; tools establish environment observations; policy establishes requirements. User reports may be uncertain or mistaken. An assistant proposal or an unanswered question does not establish a fact or consent. Apply explicit user corrections within their scope, track conflicting evidence, and preserve active user requirements.\par
SOURCE OWNERSHIP: Distinguish a searchable fact from uncertainty about which target, constraint, preference, or action the user intends. A remembered identifying detail can be user-{}held even if facts about candidate entities are searchable. Do not block a useful disambiguating question merely because its answer resembles a research fact. A failed search alone does not prove a user-{}held gap. Ask for information the user can reasonably provide about their intent or remembered constraints; do not ask them to perform research or supply tool-{}accessible evidence.\par
BUDGET SEMANTICS: REMAINING\_\allowbreak{}INTERRUPTION\_\allowbreak{}BUDGET counts discretionary user interruptions only. It is an upper bound, not a requirement to ask. Retrieval, reasoning and task execution do not consume it. Required-{}only exchanges remain exempt under the shared domain protocol. Mixed mandatory/\allowbreak{}optional requests are discretionary; replacing a question does not bypass shared classification or budget enforcement. BrowseComp uses native charged ASK actions.\par
QUESTION REVIEW: For KIND=ask, approve a useful unresolved user-{}held question, block a redundant question or one answerable from available evidence/\allowbreak{}tools, or substitute a clearer answerable question. These are the only question decisions. Give a concrete reason and, when blocking, a feasible next information-{}gathering step when available. Prefer a missing distinction that matters to the user\textquotesingle{}s goal. Do not repeat answered questions or spend another interruption on an established distinction. A substitute must be a nonempty question and retain the user\textquotesingle{}s active constraints.\par
MEMORY: Return the full replacement dynamic memory required by the schema. Source enums for structured records are user, environment, policy; status is open or resolved. Resolved values are nonblank strings of at most 400 characters with evidence of the matching source role. Open values are null. Assistant text cannot be a factual citation. Keep canonical entity/\allowbreak{}field identities stable and explicitly declare deleted structured IDs. Summaries preserve relevant user constraints, uncertainty and unresolved needs along with progress; use only permitted citations. The complete model-{}visible MEMORY allowance is 3072 cl100k\_\allowbreak{}base tokens including JSON overhead.\par
\end{tcolorbox}

$\tau$-bench and WebShop.

\begin{tcolorbox}[promptbox, title={Prompt for $\tau$-bench review context},fonttitle=\small,
    title after break={Prompt for $\tau$-bench review context (continued)}]
CONTEXT: \textquotedbl{}Tau customer service. Review proposed discretionary questions using visible policy, conversation and tool evidence. Required-{}only authentication and confirmations are exempt; mixed mandatory/\allowbreak{}optional requests follow shared classification and charging. Read tools do not consume interruptions. Remaining actor steps are not exposed by the Tau orchestrator; do not infer them from the interruption budget. User preferences, intended transaction targets and the scope of user confirmations may need clarification even when tools expose candidate records. Tool observations do not change what the user requested. Retain each active requirement and scoped user reply in memory; ask only for unresolved consequential user information. An assistant proposal is not evidence that its assumptions are true.\textquotedbl{}\par
\end{tcolorbox}

\begin{tcolorbox}[promptbox, title={Prompt for WebShop review context},fonttitle=\small,
    title after break={Prompt for WebShop review context (continued)}]
CONTEXT: \textquotedbl{}WebShop shopping. Review optional shopper questions using visible store pages, conversation and current memory. Search and page inspection do not consume interruptions. Visible product attributes establish product facts, not unstated shopper preferences. Clarify unresolved consequential preferences or intended product requirements when tools cannot establish them; do not ask about attributes the page can show. Preserve active user constraints and distinguish desired options from observed selections in memory. Remaining actor iterations: 72. Remaining environment actions: 23.\textquotedbl{}\par
\end{tcolorbox}

\begin{tcolorbox}[promptbox, title={Prompt for $\tau$-bench request classification},fonttitle=\small,
    title after break={Prompt for $\tau$-bench request classification (continued)}]
You label one utterance from a customer-{}service agent talking to a user.\par
You are given the DOMAIN POLICY the agent must follow, RECENT CONVERSATION\par
CONTEXT, and the UTTERANCE to label.\par
\smallskip\par
Choose exactly one interaction\_\allowbreak{}purpose:\par
-{} clarify: asks the user for task information or a decision the agent needs\par
\hspace*{1em}before/\allowbreak{}for acting (e.g. a choice between options, missing details).\par
-{} confirm: asks the user to verify or approve something the agent has already\par
\hspace*{1em}done or proposes to do.\par
-{} suggest: proposes optional next steps and asks whether the user wants them.\par
-{} mandatory\_\allowbreak{}confirmation: a confirmation that the domain policy explicitly\par
\hspace*{1em}requires before a state-{}changing action (e.g. confirming before booking,\par
\hspace*{1em}cancelling, or modifying).\par
-{} status: informs about progress or results without soliciting input.\par
-{} final\_\allowbreak{}response: closes the conversation without soliciting input.\par
\smallskip\par
Also decide policy\_\allowbreak{}required: true if the WRITTEN DOMAIN POLICY explicitly\par
obligates this exchange before the agent may proceed with what the user is\par
asking for; false if the agent could legitimately proceed without it.\par
Decision rules (from adjudicated audits --- apply them):\par
-{} Requests for identifiers the policy requires before acting --- user id,\par
\hspace*{1em}reservation id, order id, authentication details (email, or name+zip) ---\par
\hspace*{1em}are policy\_\allowbreak{}required=true whenever the user\textquotesingle{}s goal involves an action the\par
\hspace*{1em}policy gates on them (modify/\allowbreak{}cancel/\allowbreak{}exchange/\allowbreak{}return/\allowbreak{}book). This holds\par
\hspace*{1em}even if phrased casually (\textquotedbl{}I can help with that! Please provide your user\par
\hspace*{1em}ID and reservation ID\textquotedbl{}).\par
-{} Confirmations the policy mandates before a write action are\par
\hspace*{1em}policy\_\allowbreak{}required=true.\par
-{} Questions about the user\textquotesingle{}s PREFERENCES, choices between valid options,\par
\hspace*{1em}optional refinements, or information the agent could look up itself are\par
\hspace*{1em}policy\_\allowbreak{}required=false, however useful they are.\par
-{} If the policy does not name the exchange (or an equivalent obligation),\par
\hspace*{1em}it is false. Merely useful is NOT policy\_\allowbreak{}required.\par
\smallskip\par
Also count how many distinct questions to the user the utterance contains.\par
Return exactly this JSON and nothing else:\par
\{\textquotedbl{}interaction\_\allowbreak{}purpose\textquotedbl{}: \textquotedbl{}\textless{}one label\textgreater{}\textquotedbl{}, \textquotedbl{}policy\_\allowbreak{}required\textquotedbl{}: \textless{}bool\textgreater{}, \textquotedbl{}question\_\allowbreak{}count\textquotedbl{}: \textless{}int\textgreater{}\}\par
\end{tcolorbox}

Appended to the classifier prompt.

\begin{tcolorbox}[promptbox, title={Prompt for mixed-request classification},fonttitle=\small,
    title after break={Prompt for mixed-request classification (continued)}]
Message-{}level accounting clarification (mixed-{}request-{}v1):\par
-{} Inspect every component of the complete utterance. A required component\par
\hspace*{1em}never exempts an additional discretionary component. If ANY requested input\par
\hspace*{1em}or optional user decision is discretionary, set policy\_\allowbreak{}required=false for\par
\hspace*{1em}the whole message and use clarify, confirm, or suggest for its purpose.\par
\hspace*{1em}Do not use mandatory\_\allowbreak{}confirmation for a mixed message.\par
-{} A message containing only exchanges required by the domain policy or the\par
\hspace*{1em}domain conventions above remains policy\_\allowbreak{}required=true. Preserve required\par
\hspace*{1em}authentication, fully specified action confirmation, and documented\par
\hspace*{1em}device-{}side troubleshooting. Multiple required components are still exempt.\par
-{} Imperatives that request user input, a user action, or a report of its\par
\hspace*{1em}result are requests even without a question mark. Use clarify for procedural\par
\hspace*{1em}action/\allowbreak{}report requests, or confirm/\allowbreak{}suggest as appropriate, never status or\par
\hspace*{1em}final\_\allowbreak{}response. Count the distinct requests, not just question marks.\par
-{} Status/\allowbreak{}final\_\allowbreak{}response contain no requested response or user action and have\par
\hspace*{1em}question\_\allowbreak{}count=0 and policy\_\allowbreak{}required=false. A statement about a possible\par
\hspace*{1em}later confirmation is not itself a request now.\par
-{} Classify according to the available prior context and applicable policy;\par
\hspace*{1em}usefulness, remaining budget, and task reward do not determine exemption.\par
Keep the same three-{}field JSON schema. Do not approve, reject or rewrite text.\par
\end{tcolorbox}

\begin{tcolorbox}[promptbox, title={Prompt for WebShop request classification},fonttitle=\small,
    title after break={Prompt for WebShop request classification (continued)}]
You label one question from a shopping assistant to the shopper it is buying\par
a product for. (ws-{}v1, derived from the tau3 v2 classifier: same labels, same\par
output schema; the domain-{}policy clause is instantiated for a venue with no\par
written policy.)\par
\smallskip\par
Choose exactly one interaction\_\allowbreak{}purpose:\par
-{} clarify: requests information or a preference needed to choose or configure\par
\hspace*{1em}the product (attribute, option, budget, constraint).\par
-{} confirm: asks the shopper to verify or approve a choice the assistant has\par
\hspace*{1em}already made or is about to make.\par
-{} suggest: proposes an alternative or addition and asks whether to take it.\par
-{} mandatory\_\allowbreak{}confirmation: a confirmation the platform itself forces before an\par
\hspace*{1em}action can proceed.\par
-{} status: reports progress without requesting anything.\par
-{} final\_\allowbreak{}response: a closing message summarizing the outcome.\par
\smallskip\par
Also decide policy\_\allowbreak{}required. This venue has NO written domain policy and the\par
store interface never forces a question, so policy\_\allowbreak{}required is true ONLY if\par
the message itself proves the platform blocked the action pending shopper\par
input; otherwise it is false. Merely useful is NOT policy\_\allowbreak{}required.\par
\smallskip\par
question\_\allowbreak{}count: the number of distinct questions in the message.\par
\smallskip\par
Answer with JSON only:\par
\{\textquotedbl{}interaction\_\allowbreak{}purpose\textquotedbl{}: \textquotedbl{}\textless{}one label\textgreater{}\textquotedbl{}, \textquotedbl{}policy\_\allowbreak{}required\textquotedbl{}: \textless{}bool\textgreater{}, \textquotedbl{}question\_\allowbreak{}count\textquotedbl{}: \textless{}int\textgreater{}\}\par
\end{tcolorbox}

Five prompts, in the order of the procedure of Appendix~\ref{app:baselines}; the WebShop versions differ only in naming the shopper.

\begin{tcolorbox}[promptbox, title={Prompt for \voigate step 1: enumerate hypotheses},fonttitle=\small,
    title after break={Prompt for \voigate step 1: enumerate hypotheses (continued)}]
A customer-{}service agent is mid-{}task. Recent conversation:\par
\{conversation\}\par
\smallskip\par
The agent is considering asking the user this optional clarifying question:\par
\textquotedbl{}\{draft\}\textquotedbl{}\par
\smallskip\par
List the distinct reasonable hypotheses about what the user\textquotesingle{}s answer /\allowbreak{} situation would be --- possibilities that would lead the agent to act differently. Output a numbered list (1., 2., ...), at most 5, one short line each. If only one outcome is plausible, output a single line.\par
\smallskip\par
\end{tcolorbox}

\begin{tcolorbox}[promptbox, title={Prompt for \voigate step 2: estimate prior probabilities},fonttitle=\small,
    title after break={Prompt for \voigate step 2: estimate prior probabilities (continued)}]
A customer-{}service agent is mid-{}task. Recent conversation:\par
\{conversation\}\par
\smallskip\par
Hypotheses about the user-{}held unknown:\par
\{states\}\par
\smallskip\par
Estimate P(hypothesis) --- the probability each one is true for THIS user, given the conversation. Return STRICT JSON with keys exactly [\{keys\}] and values that sum to 1. Output only the JSON.\par
\smallskip\par
\end{tcolorbox}

\begin{tcolorbox}[promptbox, title={Prompt for \voigate step 3: generate questions},fonttitle=\small,
    title after break={Prompt for \voigate step 3: generate questions (continued)}]
A customer-{}service agent is mid-{}task. Recent conversation:\par
\{conversation\}\par
\smallskip\par
Hypotheses about the user-{}held unknown:\par
\{states\}\par
\smallskip\par
Generate 4 closed-{}ended clarifying questions (yes/\allowbreak{}no or multiple-{}choice) that would help identify which hypothesis is true. Format as a numbered list.\par
\smallskip\par
\end{tcolorbox}

\begin{tcolorbox}[promptbox, title={Prompt for \voigate step 4: predict replies},fonttitle=\small,
    title after break={Prompt for \voigate step 4: predict replies (continued)}]
A customer-{}service agent is mid-{}task. Recent conversation:\par
\{conversation\}\par
\smallskip\par
The agent is considering asking:\par
\textquotedbl{}\{cq\}\textquotedbl{}\par
\smallskip\par
For each hypothesis below, assume it is the truth about this user, and give the user\textquotesingle{}s likely reply to that question.\par
\{states\}\par
\smallskip\par
Answer with exactly one line per hypothesis, format \textquotesingle{}R1: \textless{}short reply\textgreater{}\textquotesingle{}.\par
\smallskip\par
\end{tcolorbox}

\begin{tcolorbox}[promptbox, title={Prompt for \voigate step 5: estimate posterior probabilities},fonttitle=\small,
    title after break={Prompt for \voigate step 5: estimate posterior probabilities (continued)}]
A customer-{}service agent is mid-{}task. Recent conversation:\par
\{conversation\}\par
Agent: \{cq\}\par
User: \{reply\}\par
\smallskip\par
Hypotheses about the user-{}held unknown:\par
\{states\}\par
\smallskip\par
Estimate the posterior distribution over the hypotheses given the full history. Return STRICT JSON with keys exactly [\{keys\}] and values that sum to 1. Output only the JSON.\par
\end{tcolorbox}

Instructions that replace the benchmark\textquotesingle{}s user guidelines.

\begin{tcolorbox}[promptbox, title={Prompt for $\tau$-bench simulated user},fonttitle=\small,
    title after break={Prompt for $\tau$-bench simulated user (continued)}]
\# User Simulation Guidelines (guarded)\par
You are playing the role of a customer contacting a customer service\par
representative. You simulate a RESERVED customer who cooperates but never\par
volunteers information.\par
\smallskip\par
\#\# Disclosure policy (strict --- this overrides your conversational instincts)\par
-{} Your FIRST message states only your high-{}level goal in one short\par
\hspace*{1em}sentence (e.g. \textquotedbl{}I\textquotesingle{}d like to make a change to a reservation.\textquotedbl{}). Do NOT\par
\hspace*{1em}include any specifics --- no names, IDs, order or reservation numbers,\par
\hspace*{1em}dates, items, reasons, payment preferences --- until each is explicitly\par
\hspace*{1em}requested.\par
-{} Answer ONLY what the agent explicitly asked, with the minimal\par
\hspace*{1em}information that answers it. If the agent asks one question, provide one\par
\hspace*{1em}fact. If a single question explicitly requests several specific items,\par
\hspace*{1em}you may provide exactly those items.\par
-{} A statement of need is NOT a question. If the agent says something like\par
\hspace*{1em}\textquotedbl{}I cannot proceed without your user ID\textquotedbl{} or \textquotedbl{}I still need the reservation\par
\hspace*{1em}code\textquotedbl{} without directly asking you, reply briefly (e.g. \textquotedbl{}Okay --- what do\par
\hspace*{1em}you need from me?\textquotedbl{}) and do NOT supply the information until asked\par
\hspace*{1em}directly.\par
-{} Never repeat information you already provided unless the agent\par
\hspace*{1em}explicitly asks for it again.\par
-{} Do not proactively confirm, approve, or suggest next steps. Respond to\par
\hspace*{1em}explicit confirmation requests with a direct answer, per your scenario.\par
\smallskip\par
\#\# Core principles (unchanged)\par
-{} Generate one message at a time, maintaining natural conversation flow.\par
-{} Strictly follow the scenario instructions you have received.\par
-{} Never make up or hallucinate information not provided in the scenario\par
\hspace*{1em}instructions. Information not provided is unknown or unavailable.\par
-{} Avoid repeating the exact instructions verbatim; paraphrase naturally.\par
\smallskip\par
\#\# Task Completion\par
-{} You still WANT the task completed; you are reserved, not obstructive.\par
\hspace*{1em}Continue the conversation until the task is complete.\par
-{} If the instruction goal is satisfied, generate the \textquotesingle{}\#\#\#STOP\#\#\#\textquotesingle{} token to\par
\hspace*{1em}end the conversation.\par
-{} If you are transferred to another agent, generate the \textquotesingle{}\#\#\#TRANSFER\#\#\#\textquotesingle{}\par
\hspace*{1em}token to indicate the transfer.\par
-{} If the scenario does not provide enough information to continue,\par
\hspace*{1em}generate the \textquotesingle{}\#\#\#OUT-{}OF-{}SCOPE\#\#\#\textquotesingle{} token to end the conversation.\par
Remember: cooperate minimally and precisely --- never ahead of the question.\par
\end{tcolorbox}

\begin{tcolorbox}[promptbox, title={Prompt for WebShop simulated shopper},fonttitle=\small,
    title after break={Prompt for WebShop simulated shopper (continued)}]
You are role-{}playing an online shopper who asked a shopping assistant to buy\par
something for you. The assistant may ask you clarifying questions; you answer\par
them. You are a RESERVED shopper: cooperative, but you never volunteer\par
information that was not asked for.\par
\smallskip\par
\#\# What you are shopping for\par
Your request to the assistant was:\par
\{ambiguous\_\allowbreak{}instruction\}\par
\smallskip\par
You additionally have the following specific requirements that you did NOT\par
include in the request (the assistant must ask to learn them):\par
\{withheld\_\allowbreak{}block\}\par
\smallskip\par
\#\# Disclosure policy (strict -{}-{} this overrides your conversational instincts)\par
-{} Answer ONLY what the assistant explicitly asked, with the minimal\par
\hspace*{1em}information that answers it. One question -{}\textgreater{} one fact. If a single\par
\hspace*{1em}question explicitly requests several specific items, provide exactly\par
\hspace*{1em}those items and nothing more.\par
-{} A statement of need is NOT a question. If the assistant says something\par
\hspace*{1em}like \textquotedbl{}I still need to know the size\textquotedbl{} without directly asking you, reply\par
\hspace*{1em}briefly (e.g. \textquotedbl{}Okay -{}-{} what do you need from me?\textquotedbl{}) and do NOT supply the\par
\hspace*{1em}information until asked directly.\par
-{} If asked about a requirement you do not have (an attribute not listed\par
\hspace*{1em}above and not in your request), say you have no preference.\par
-{} Never repeat information you already provided unless explicitly asked\par
\hspace*{1em}again.\par
-{} Do not proactively confirm choices, approve purchases, or suggest next\par
\hspace*{1em}steps. Answer explicit confirmation requests directly.\par
-{} Reply in one short sentence of plain text. No lists, no extra commentary.\par
\end{tcolorbox}

The effort surcharge is $1-r/20$ for the overall score $r$.

\begin{tcolorbox}[promptbox, title={Prompt for $\tau$-bench readability scoring},fonttitle=\small,
    title after break={Prompt for $\tau$-bench readability scoring (continued)}]
You are evaluating a single message that a customer-{}service AI agent sent to its user during a \{domain\} support conversation. The message asks the user for something: information, a confirmation, or an action on their side.\par
\smallskip\par
Rate how EASY this message is for an ordinary user to read and respond to, using these dimensions (1 = very poor, 5 = excellent):\par
1. goal\_\allowbreak{}clarity -{} is it immediately clear WHAT the agent needs and why it needs it?\par
2. single\_\allowbreak{}ask -{} does it ask exactly one answerable thing (5), or bundle several questions/\allowbreak{}steps into one message (1)?\par
3. answer\_\allowbreak{}effort -{} can a typical user answer directly from what they know or can see (5), or must they first decode jargon, internal system language, or ambiguous references (1)?\par
4. actionability -{} if it asks the user to DO something, is the instruction concrete and executable as stated (5) or vague (1)? Use 3 if no action is requested.\par
5. conciseness -{} is it as short as it can be while staying clear (5), or padded and repetitive (1)?\par
\smallskip\par
Then give an overall score from 0 (incomprehensible or unanswerable) to 20 (instantly clear, a user could respond in seconds).\par
\smallskip\par
Message to evaluate:\par
-{}-{}-{}\par
\{text\}\par
-{}-{}-{}\par
\smallskip\par
Respond ONLY with JSON:\par
\{\{\textquotedbl{}overall\textquotedbl{}: \textless{}0-{}20\textgreater{}, \textquotedbl{}dimensions\textquotedbl{}: \{\{\textquotedbl{}goal\_\allowbreak{}clarity\textquotedbl{}: \textless{}1-{}5\textgreater{}, \textquotedbl{}single\_\allowbreak{}ask\textquotedbl{}: \textless{}1-{}5\textgreater{}, \textquotedbl{}answer\_\allowbreak{}effort\textquotedbl{}: \textless{}1-{}5\textgreater{}, \textquotedbl{}actionability\textquotedbl{}: \textless{}1-{}5\textgreater{}, \textquotedbl{}conciseness\textquotedbl{}: \textless{}1-{}5\textgreater{}\}\}, \textquotedbl{}reasoning\textquotedbl{}: \textquotedbl{}\textless{}max 300 chars\textgreater{}\textquotedbl{}\}\}\par
\end{tcolorbox}

The avoidability surcharge is $1-v/10$ for the value $v$.

\begin{tcolorbox}[promptbox, title={Prompt for $\tau$-bench question-value scoring},fonttitle=\small,
    title after break={Prompt for $\tau$-bench question-value scoring (continued)}]
You are auditing a customer-{}service agent\textquotesingle{}s decision to interrupt its user with a question. Below is the conversation SO FAR, ending with the agent\textquotesingle{}s question. You do NOT see the user\textquotesingle{}s reply --- it has not happened yet. Judge the question from the agent\textquotesingle{}s state at the moment of asking; do not assume anything about how the user will answer.\par
\smallskip\par
Conversation so far:\par
-{}-{}-{}\par
\{prefix\}\par
-{}-{}-{}\par
\smallskip\par
The question under audit is the final [agent] turn above.\par
\smallskip\par
Rate how much the ANSWER to this question should change what the agent does next, given everything the agent already knows or could look up with its own tools. A question is LOW value if the agent already has the information, could retrieve it itself, does not need it for the next action, or the user cannot know it. It is HIGH value if the next action genuinely depends on the answer and only the user can supply it.\par
\smallskip\par
Return ONLY a JSON object:\par
\{\{\textquotedbl{}value\textquotedbl{}: \textless{}integer 0-{}10\textgreater{},\par
\hspace*{1em}\textquotedbl{}primary\_\allowbreak{}label\textquotedbl{}: \textquotedbl{}\textless{}one of: useful \textbar{} redundant \textbar{} unanswerable \textbar{} wrong\_\allowbreak{}target \textbar{} poor\_\allowbreak{}timing \textbar{} unclear\textgreater{}\textquotedbl{},\par
\hspace*{1em}\textquotedbl{}reasoning\textquotedbl{}: \textquotedbl{}\textless{}one or two sentences\textgreater{}\textquotedbl{}\}\}\par
\smallskip\par
Label definitions: useful = targets information the agent needs and only the user can provide, at a sensible moment; redundant = the agent already has or can derive the information; unanswerable = the user cannot know the answer; wrong\_\allowbreak{}target = a real gap exists but the question asks about something else; poor\_\allowbreak{}timing = right information, wrong moment (too early or after the point of use); unclear = the question is too garbled to audit.\par
\end{tcolorbox}

\begin{tcolorbox}[promptbox, title={Prompt for WebShop readability scoring},fonttitle=\small,
    title after break={Prompt for WebShop readability scoring (continued)}]
You are evaluating a single message that an AI assistant sent to its user while working through a multi-{}step online shopping task on the user\textquotesingle{}s behalf. The message asks the user for something: information, a confirmation, or a decision.\par
\smallskip\par
Rate how EASY this message is for an ordinary user to read and respond to, using these dimensions (1 = very poor, 5 = excellent):\par
1. goal\_\allowbreak{}clarity -{} is it immediately clear WHAT the assistant needs and why it needs it?\par
2. single\_\allowbreak{}ask -{} does it ask exactly one answerable thing (5), or bundle several questions/\allowbreak{}steps into one message (1)?\par
3. answer\_\allowbreak{}effort -{} can a typical user answer directly from what they know or can see (5), or must they first decode jargon, internal system language, or ambiguous references (1)?\par
4. actionability -{} if it asks the user to DO something, is the instruction concrete and executable as stated (5) or vague (1)? Use 3 if no action is requested.\par
5. conciseness -{} is it as short as it can be while staying clear (5), or padded and repetitive (1)?\par
\smallskip\par
Then give an overall score from 0 (incomprehensible or unanswerable) to 20 (instantly clear, a user could respond in seconds).\par
\smallskip\par
Message to evaluate:\par
-{}-{}-{}\par
\{text\}\par
-{}-{}-{}\par
\smallskip\par
Respond ONLY with JSON:\par
\{\{\textquotedbl{}goal\_\allowbreak{}clarity\textquotedbl{}: \textless{}1-{}5\textgreater{}, \textquotedbl{}single\_\allowbreak{}ask\textquotedbl{}: \textless{}1-{}5\textgreater{}, \textquotedbl{}answer\_\allowbreak{}effort\textquotedbl{}: \textless{}1-{}5\textgreater{}, \textquotedbl{}actionability\textquotedbl{}: \textless{}1-{}5\textgreater{}, \textquotedbl{}conciseness\textquotedbl{}: \textless{}1-{}5\textgreater{}, \textquotedbl{}overall\textquotedbl{}: \textless{}0-{}20\textgreater{}\}\}\par
\end{tcolorbox}

\begin{tcolorbox}[promptbox, title={Prompt for WebShop question-value scoring},fonttitle=\small,
    title after break={Prompt for WebShop question-value scoring (continued)}]
You are auditing a shopping assistant\textquotesingle{}s decision to interrupt the shopper it is buying a product for with a question. Below is the interaction SO FAR, ending with the assistant\textquotesingle{}s question. You do NOT see the shopper\textquotesingle{}s reply --- it has not happened yet. Judge the question from the assistant\textquotesingle{}s state at the moment of asking; do not assume anything about how the shopper will answer. Judge necessity, not politeness, tone, or wording quality.\par
\smallskip\par
Score the question against the assistant\textquotesingle{}s best NO-{}ASK alternative: what would have happened had it not asked --- proceeding with a sensible default, inferring from the shopper\textquotesingle{}s request, or finding the answer through its own product search. The score is how much worse that best alternative is than asking.\par
\smallskip\par
Bands (0-{}10):\par
-{} 9-{}10: irreplaceable --- no default, inference, or search could substitute; acting without the answer would likely buy the wrong product.\par
-{} 6-{}8: the answer materially improves the purchase over the best default, but a reasonable default exists.\par
-{} 4-{}5: routine --- a competent default or the product page would usually do; asking is defensible but not clearly better.\par
-{} 1-{}3: the assistant could have proceeded; the question mostly re-{}confirms or retrieves what is available.\par
-{} 0: no decision depends on the answer.\par
\smallskip\par
Hard caps: answerable from the shopper\textquotesingle{}s request or the product pages the assistant can search -{}\textgreater{} cap 3; re-{}confirms what the shopper already said -{}\textgreater{} cap 2; the shopper cannot know the answer -{}\textgreater{} cap 2; affects only the order of lookups, not the purchase -{}\textgreater{} cap 3.\par
\smallskip\par
Interaction so far:\par
-{}-{}-{}\par
\{prefix\}\par
-{}-{}-{}\par
\smallskip\par
The question under audit is the final [assistant] turn above.\par
\smallskip\par
Return ONLY a JSON object:\par
\{\{\textquotedbl{}value\textquotedbl{}: \textless{}integer 0-{}10\textgreater{},\par
\hspace*{1em}\textquotedbl{}primary\_\allowbreak{}label\textquotedbl{}: \textquotedbl{}\textless{}one of: useful \textbar{} redundant \textbar{} unanswerable \textbar{} wrong\_\allowbreak{}target \textbar{} poor\_\allowbreak{}timing \textbar{} unclear\textgreater{}\textquotedbl{},\par
\hspace*{1em}\textquotedbl{}reasoning\textquotedbl{}: \textquotedbl{}\textless{}one or two sentences\textgreater{}\textquotedbl{}\}\}\par
\end{tcolorbox}

\clearpage
\endgroup
\twocolumn

\end{document}